\pdfoutput=1
\documentclass[journal=jacsat,manuscript=article]{achemso}
\setkeys{acs}{articletitle = true}
\usepackage[utf8]{inputenc}
\usepackage{pdfpages}
\usepackage[version=3]{mhchem} 
\usepackage{subcaption}
\usepackage[font=scriptsize]{caption}
\usepackage{hyperref}

\title{Unassisted Noise Reduction of Chemical Reaction Data Sets}

\usepackage{titlesec}
\titleformat{\section}[block]
  {\fontsize{14}{17}\bfseries\sffamily}
  {\thesection}
  {1em}
  {}
\titleformat{\subsection}[hang]
  {\fontsize{12}{15}\bfseries\sffamily}
  {\thesubsection}
  {1em}
  {}

\newcommand{\black}{\textcolor[rgb]{0.00,0.00,0.00}}

\author{Alessandra Toniato}
\affiliation{IBM Research Europe -- Zurich,  R\"{u}schlikon, Switzerland}
\email{ato@zurich.ibm.com}
\author{Philippe Schwaller}
\affiliation{IBM Research Europe -- Zurich,  R\"{u}schlikon, Switzerland}
\alsoaffiliation{Department of Chemistry, University of Bern, Bern, Switzerland}
\author{Antonio Cardinale}
\affiliation{IBM Research Europe -- Zurich,  R\"{u}schlikon, Switzerland}
\alsoaffiliation{Department of Chemistry, University of Pisa, Pisa, Italy}
\author{Joppe Geluykens}
\affiliation{IBM Research Europe -- Zurich,  R\"{u}schlikon, Switzerland}
\author{Teodoro Laino}
\affiliation{IBM Research Europe -- Zurich,  R\"{u}schlikon, Switzerland}

\title[\texttt{achemso} demonstration]
{Unassisted Noise Reduction of Chemical Reaction Data Sets}

\begin{document}

\begin{abstract}
Existing deep learning models applied to reaction prediction in organic chemistry can reach high levels of accuracy ($>$ 90\% for Natural Language Processing-based ones). With no chemical knowledge embedded than the information learnt from reaction data, the quality of the data sets plays a crucial role in the performance of the prediction models. While human curation is prohibitively expensive, the need for unaided approaches to remove chemically incorrect entries from existing data sets  is essential to improve  artificial intelligence models' performance in synthetic chemistry tasks. Here we propose a machine learning-based, unassisted approach to remove chemically wrong entries from chemical reaction collections. We applied this method to the collection of chemical reactions Pistachio and to an open data set, both extracted from USPTO (United States Patent Office) patents.
\black{Our results show an improved prediction quality for models trained on} the cleaned and balanced data sets. For the retrosynthetic models, the round-trip accuracy metric grows by 13 percentage points and the value of the cumulative Jensen Shannon divergence decreases by 30\% \black{compared to its} original record. The \black{coverage remains high with 97\%}, and the value of the class-diversity is not affected by the cleaning.
The proposed strategy is the first unassisted rule-free technique \black{to address} automatic noise reduction in chemical \black{data sets}.
\end{abstract}

\section{Introduction}
\begin{figure}
\centerline{
\includegraphics[width=.9\linewidth]{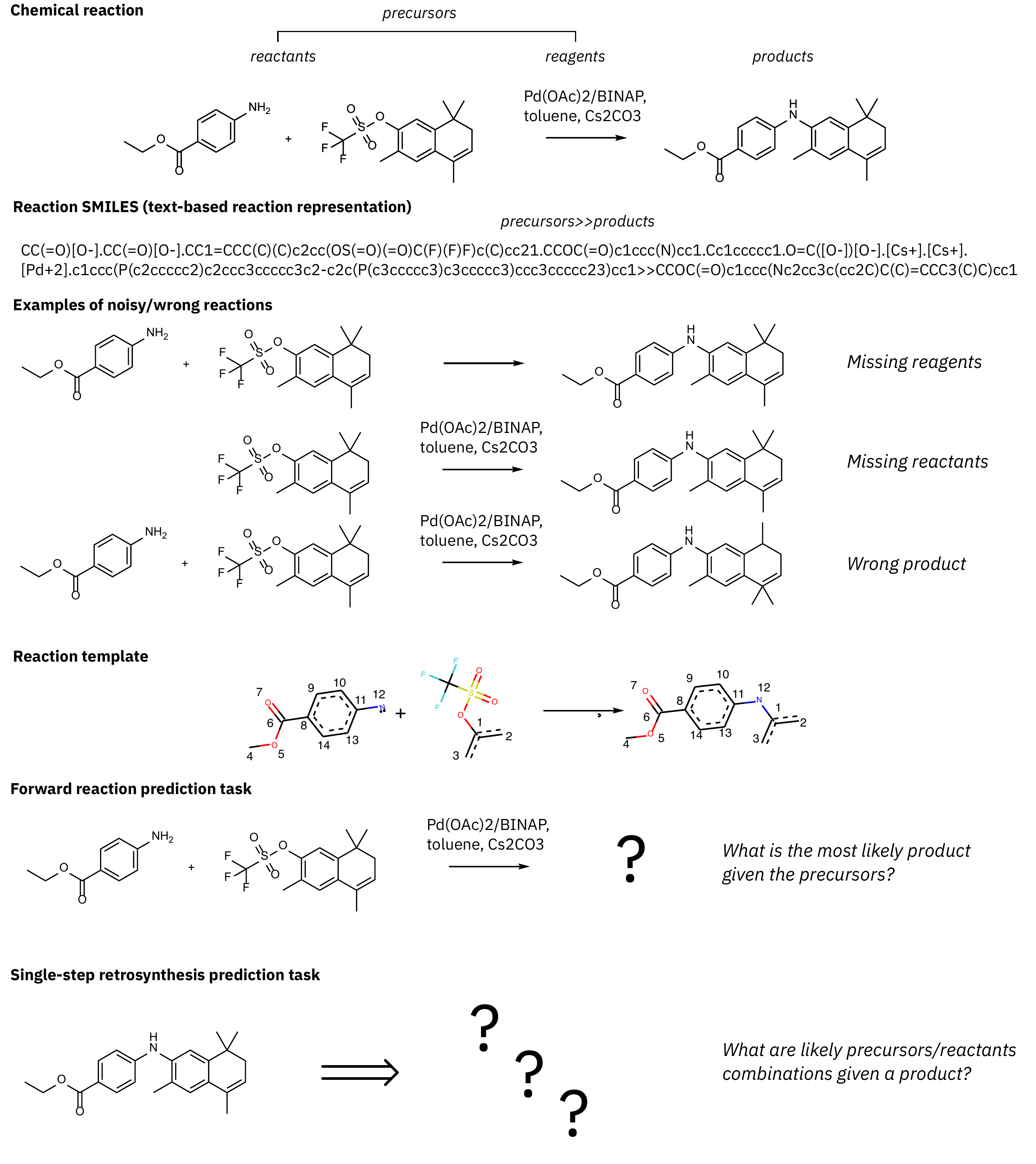}}
\captionsetup{width=.9\linewidth,font=footnotesize,labelfont=bf}
  \caption{\textbf{Overview.} \black{Starting from the top, an example of a chemical reaction and its SMILES representation, examples of potential forms of noise present in a chemical reaction, reaction templates, the tasks of forward and retrosynthesis prediction respectively. In this context, noise is defined as any incomplete reaction entry missing any reactants, reagents or incorrect product recorded.}}
  \label{fig:intro}
\end{figure}

The last decade witnessed a \black{extensive development and the} application of several data-driven approaches to synthetic organic chemistry, mainly thanks to the availability of chemical reaction data sets\cite{lowe2012extraction, Lowe2017}. The publicly available United States Patent Office (USPTO) \cite{Lowe2017} chemical reaction collection, along with proprietary Pistachio \cite{Pistachio} and Reaxys \cite{Reaxys} fueled the development of several deep learning models and architectures to assist organic chemists in chemical synthesis planning \cite{Segler2018,Coleyeaax1566, schwaller2019book, SchwallerFWD,schwaller2020predicting, ozturk2020exploring}. \black{The aim of chemical synthesis planning is to find pathways from commercially available molecules to a target molecule. The individual steps in a chemical synthesis are represented by chemical reactions. Chemical reactions describe how precursor molecules react and form product molecules. Figure \ref{fig:intro} shows an example of a chemical reaction and a graphical overview of different chemical terms. The product of a reaction in a pathway is either the target molecule or used as a precursor for the subsequent reaction. Computationally solving synthesis planning typically requires a combination of a retrosynthesis prediction model, predicting precursors given a product, and reaction or pathway scoring models \cite{satoh1995sophia, Segler2018, schwaller2020predicting}.}
Despite efforts \black{to build} models that effectively learn chemistry from data, the data sets' quality remains the primary limitation on performance improvements \cite{SchwallerFWD, schwaller2020predicting}. 
The \black{influence of data set size} and variability on the performance of computer-assisted synthesis planning tools \black{has recently been} investigated by Thakkar et al. \cite{Amol}.  Nevertheless, the influence of chemically wrong examples \black{present} in training data sets remains a niche topic regardless of its relevance and impact on all data-driven chemical applications. 

Some deep learning architectures may represent data more effectively than others, but the presence of chemically wrong entries in training data sets has a detrimental effect on all of them.  Trying to learn from a large portion of chemically wrong examples affects the way models represent latent chemical rules, biasing the predictions towards unreasonable connections and disconnections. \black{For our purpose, we define a chemically valid reaction as an entry that has all reactants, reagents, and correct products. A reaction entry missing any portion of these three components or has incorrect products, we define as chemically invalid}. From a machine learning and data science point of view these errors are different expressions of the data set noise \cite{noise-class-attributes}.
The currently available strategies to remove noise from data sets rely only on applying a few specific rules designed by domain experts. For instance, identifying "wrong chemistry" based on the unsuccessful matching of predefined reaction templates provides a simple approach to remove incorrect chemistry. However, the failure to map chemical reactions with existing reaction templates may lead to a loss of crucial chemical knowledge for unmatched and potentially relevant examples.
On the other hand, humanly curating large data sets composed of millions of entries to create ground truth sets is prohibitively expensive and hinders the development of supervised approaches to identify chemically wrong examples.  The need for unaided, automatic and reliable protocols to minimize the loss of meaningful chemical knowledge, while removing noise is of critical importance to create trust in data-driven chemical synthesis models and encourage its widely adoption in industrial setup.

Here, we present an efficient technique to reduce noise in chemical reaction data sets to improve the performance of existing predictive models. In particular,
inspired by an approach applied to classification tasks\cite{2018forgetting}, \black{we designed a new unassisted balancing strategy for data noise reduction} based on machine learning (ML). The main idea behind the proposed protocol relies on the \textit{catastrophic forgetting}, the tendency of AI models to forget previously learnt events when trained on new tasks. In the context of AI-driven chemical synthesis, this behaviour can be explained \black{by a limited overlap between the distributions of the chemical reaction features of different training batches.}
\black{Similarly to language models, where the data points more difficult to learn are likely examples of wrong grammar, the most difficult examples to learn while training reaction prediction models are likely examples of wrong chemistry when compared to the chemical grammar described by the majority of the data set.} Starting from this hypothesis, we inspect each entry of the data set for the number of times it is forgotten during training. \black{A certain percentage of the} most forgotten examples are then removed and a new model is trained with the ``cleaned" data sets, \black{providing} a better representation of the latent chemical grammar. We show that this strategy leads to a \black{statistically significant} noise reduction in chemical data sets. The disclosed protocol can be used to remove chemically wrong examples from large collections of public and proprietary data sets\black{. Cleaner data is expected to improve the performance and reliability of existing forward prediction and retrosynthetic architectures \cite{Retrosym, NeuralSym,  SchwallerFWD, schwaller2020predicting, somnath2020learning, GLN, SCROP, sacha2020molecule, tetko2020augmented}.}

\section{Results and discussion}

\subsection{The forgotten events strategy} \label{the-forgotten-events-strategy}
Learning algorithms \black{have} achieved state-of-the-art performance in many application domains, including chemical reaction prediction and synthesis design. When training a neural network on new data, there is a tendency to forget the information previously learnt. This usually means new data will likely override the weights learnt in the past and degrade the model performance for past tasks. This behaviour is known as catastrophic forgetting or catastrophic interference \cite{catforgetting} and is one of the main impediments to transfer learning.

The application of the principles of catastrophic forgetting within different epochs of the same training session leads to the definition of ``forgotten" data points learnt in previous epochs\cite{2018forgetting}. This behavior may be a symptom of under-representing the forgotten entries in the data set, \black{which are} outliers of the underlying feature distribution. The outliers may be carriers of some critical feature rarely seen relative to its importance, but quite often they are \black{merely} semantically wrong data points. Assuming \black{that} the underlying feature distribution of the entire data set is a statistically correct representation of the carried knowledge, one can improve its significance by removing a certain fraction of the more frequently forgotten events. The use of domain specific statistical metrics helps \black{to determine} the maximum threshold to remove \black{the outliers}.

\subsection{Forward prediction model noise reduction} \label{forward- synthesis-prediction}
\begin{figure}[h]
\centerline{
\includegraphics[width=190mm]{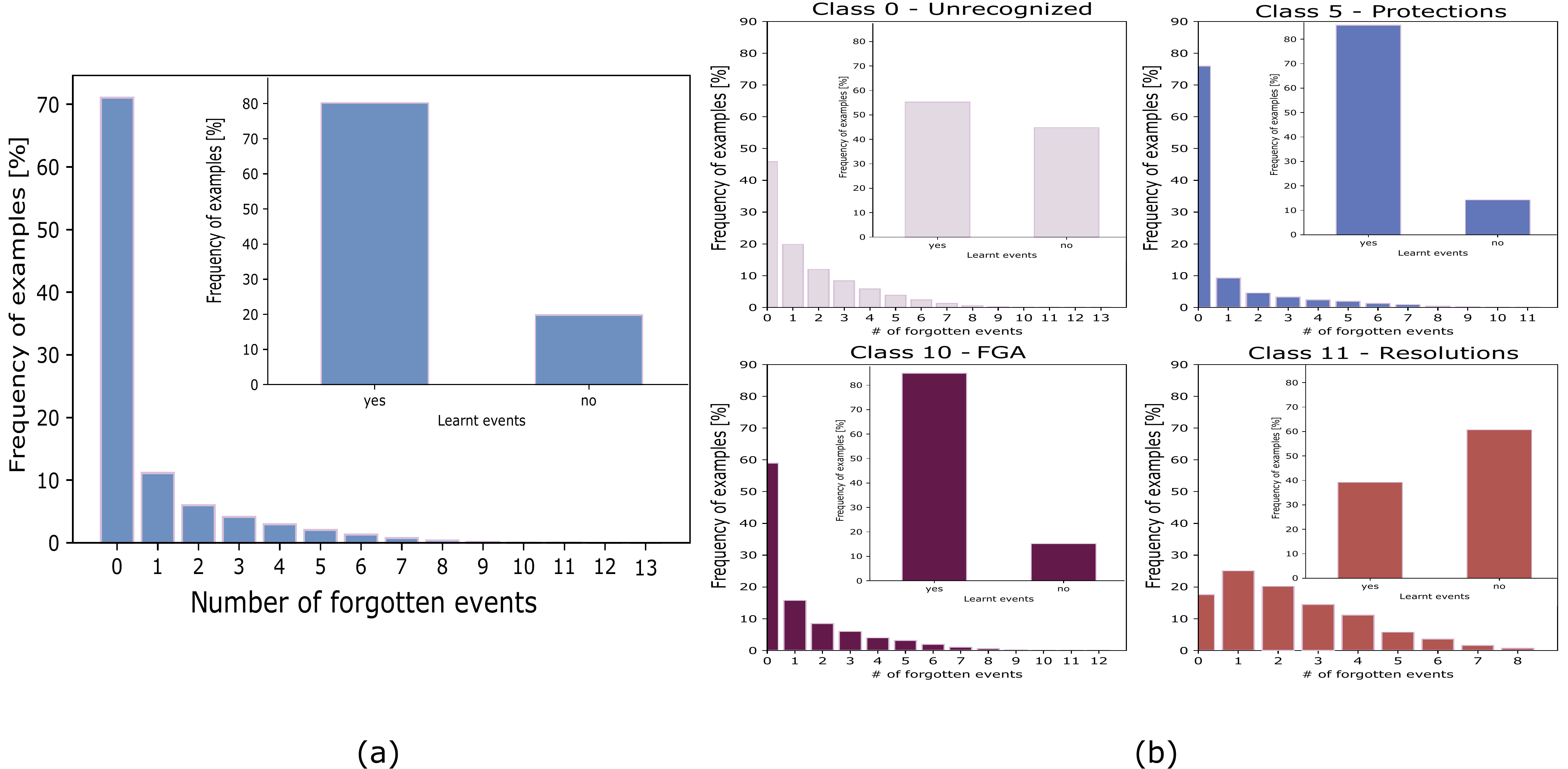}}

\captionsetup{width=.9\linewidth,font=footnotesize,labelfont=bf}
  \caption{\textbf{Results of the forgetting forward experiment.} (a) Percentage of forgotten examples as a function of the frequency of the forgotten events. Note that the percentage refers to the number of examples that have been learnt at least once. The label "0" denotes examples that were learnt once and never forgotten at a later epoch. (b) Percentage of forgotten examples as a function of the frequency of the forgotten events for specific superclasses: Unrecognized, Protections, Functional Group Addition, Resolutions. The insets show the percentage of events learnt at least once (yes) or never learnt (no).}
  \label{fig:forgettingfwd}
\end{figure}

We applied this strategy to the transformer-based \cite{vaswani2017attention} forward model by Schwaller et al. \cite{SchwallerFWD} using a proprietary data set, Pistachio, and an open data set, both containing chemical reactions extracted from Patents. Here, we report the results on the commercial Pistachio data set, properly pre-filtered (see section \ref{the-data} for pre-filtering strategies). \black{The results} for the open data set are discussed in Section \ref{open-data-set}.  The molecular transformer is a sequence-based approach, which \black{treats} the reaction prediction as a translation problem. \black{For the predictions during the training after each epoch, we considered only the top-1 prediction out of five beams.} Like previous transformer studies, we  trained long enough to reach convergence (approx. 48 hours on a 1-GPU cluster node or \black{260k} training steps. The batch size was set to 6144). \black{The convergence point was identified by monitoring the accuracy of the validation set and selecting the training step after which this accuracy stabilized.} During \black{the training}, we analyzed the forgetting events across 34 epochs.
Figure \ref{fig:forgettingfwd} shows the results for the forgetting experiment.

Approximately 80\% of the entire data set \black{was} learnt at least once during training. Out of \black{these} 80\%, 70\% of the examples \black{were} never forgotten by the transformer across epochs, once they \black{were} learnt. As expected, the distribution of forgotten events is quite \black{inhomogeneous across different superclasses: in Functional Group Additions (class 10) and Protections (class 5) the number of examples never forgotten is higher in percentage than in} Unrecognized (class 0) and Resolutions (class 11) (see Figure \ref{fig:forgettingfwd}.b). For \black{some} classes, such as Unrecognized (class 0) and Resolutions (class 11), the distribution of the number of forgotten events shows a less pronounced peak around zero and a significant population of entries that are learnt and forgotten with larger frequencies. \black{The two aforementioned superclasses contain largest amount of the wrong chemistry in the Pistachio data set, and are therefore where the model struggles to learn.} The large population of chemically wrong examples in these classes originates from the difficulties of consistently text-mining stereochemical information (for Resolutions) or matching a text-mined reaction with existing chemical templates (for Unrecognized). The results for the other classes can be found in the SI (\black{Supplementary Figure 2 and 3}).
 
The cohort of never learnt examples likely includes chemically wrong data and chemically correct reactions with features (i.e. reaction templates) rarely seen across the entire data set. The removal of a large section of such elements would cause the loss of important information and, consequently \black{lead to} a reduced performance of the model. While there is no possibility to tag each forgotten event as either rare or wrong chemistry, we need to apply strategies to minimize the loss of rare but important information while removing noise. For this purpose, the set of training samples was first \black{sorted} based on the registered number of forgetting events. Starting \black{from entries that} were never learnt and proceeding from the high-end tail of the distribution of forgotten events towards the never forgotten ones, we removed increasingly \black{larger} portions of the data set up to a maximum of 40\%. Each reduced set was used to train a new forward \black{prediction} model. In Table \ref{table:percentages} we report the new models' top-1 and top-2 results \black{achieved} on a common test set, in comparison with the baseline model (see section \ref{the-baseline-model} for baseline and test set details).

\begin{table}[]
\makebox[\textwidth][c]{
\begin{tabular}{@{}llllllll@{}}
\textbf{Model}  & \textbf{Test set}  & \textbf{\% Forgotten}  & \textbf{Model name}  & \textbf{\# of} & \textbf{Top-1} & \textbf{Top-2} & \textbf{$\sqrt{\text{CJSD}}$} \\
\textbf{type}   &                 &   \textbf{events}     &               &      \textbf{samples}      &               &       \\
forward  & Pist2019  & 0.1\%  & forget01perc  & 2 376 480  &  68.8 & 74.6  & 0.052\\
forward  & Pist2019  & 1\%   & forget1perc   & 2 355 070  & 68.9  & 74.8 & 0.051\\
forward  & Pist2019  & 5\%   & forget5perc  & 2 259 916  & 68.9  & 74.5 & 0.048\\
forward  & Pist2019  & 10\%   & forget10perc & 2 140 973  & 69.0  & 74.5 & 0.044\\
forward  & Pist2019  & 15\%   & forget15perc & 2 022 030 & 69.0 & 74.0 & 0.040\\
forward  & Pist2019  & 20\%   & forget20perc & 1 903 087 & 69.1 & 74.1 & 0.033\\
forward  & Pist2019  & 25\%   & forget25perc & 1 784 144 & 69.2 & 74.0 & 0.031\\
forward  & Pist2019  & 30\%   & forget30perc & 1 665 201 & 68.1 & 73.0 & 0.028\\
forward  & Pist2019  & 40\%   & forget40perc & 1 427 315 & 66.3 & 71.0 &  0.020\\
forward  & Pist2019  & 0\%  & allmixed (baseline) & 2 378 859 & 68.5 & 74.2 & 0.051
\end{tabular}
}
\captionsetup{width=.9\linewidth,font=footnotesize, labelfont=bf}
\caption{\textbf{Results on the cleaned forward models.} The top-1 and top-2 accuracies of the models trained with increasing percentages of data set entries removed by the forward forgetting experiment. For the model "forget01perc", 0.1\% of the data set was removed, for "forget1perc" the removed percentage was 1\%, and so on. All models are compared to the results of the baseline model (full data set) on the same test set. The values of the CJSD are also reported. These are calculated using the cumulative probability distributions with 300 bins, excluding the one for class 11 - Resolutions. Data set name and version have been abbreviated: e.g. "Pist2019" reads "Pistachio version 2019".}
\label{table:percentages}
\end{table}

We notice that the top-1 accuracy is \black{only} weakly affected by removing portions of the data set \black{up to} approximately 25\% of the data set is deleted: \black{for reductions of 30\% and 40\%}, the performances start to decrease as a consequence of the loss of meaningful chemical knowledge. 

We \black{have} recently introduced the square root of the cumulative Jensen Shannon divergence (CJSD) to quantify bias in retrosynthetic disconnection strategies \cite{schwaller2020predicting}. Here, we \black{have} revised its definition using a non-parametric approach, free of kernel density estimation (see sections \ref{the-metrics-for-performance-evaluation} and \ref{cumulative-jensen-shannon-divergence}).
The revised CJSD improves with the removal of \black{parts} of the data set (see Table \ref{table:percentages} and SI, Figure \black{S4}), as a consequence of an increased similarity between the prediction confidence distributions (lower $\sqrt{\text{CJSD}}$).
In fact, populations across classes in the training set become more balanced as many unrecognized reactions \black{were} removed, thus leading the trained models towards a similar confidence performance across all classes (see section \ref{the-data} and \black{Supplementary Figure 1} for class population changes \black{in} noise-reduced data sets). Moreover, removing noisy entries \black{improved} the model's confidence in establishing the correctness of a prediction, resulting in all distributions peaking more towards a confidence of 1.0.

When increasing the number of removed entries, the CJSD monotonically decreases, exhibiting a convergence towards unbiased confidence distributions across the different classes. \black{However}, the model with the best CJSD (40\% of the data set removed) \black{is} also the one with the lowest top-1 accuracy. The removal of 25\% of the data set \black{allows} the model to retain the high top-n accuracy while improving the CJSD. We implemented consistency checks for the entire removal strategy \black{by} using different training random seeds to guarantee that the number of forgetting events experienced by each sample was not the result of a random evaluation (see section 1.6 of the SI for details).

While removing forgotten events \black{increased} the significance of the underlying statistical distribution of the chemical reaction features, \black{there was still the} need to demonstrate that the entire process effectively \black{removed} chemically wrong data.
To assess this, we designed two different experiments where, after introducing artificial noise into the data, we analyzed the forgetting frequency of those entries. In the first experiment, all products of the reactions present in the validation set were shuffled and randomly reassigned to a new group of precursors, similarly to the work by Segler et al. \cite{Segler2018}. This was considered the "easy to identify" type of noise. As expected, the noise removal protocol led to the identification of 99\% of the chemically wrong examples. 

\begin{figure}
\centerline{
\includegraphics[width=190mm]{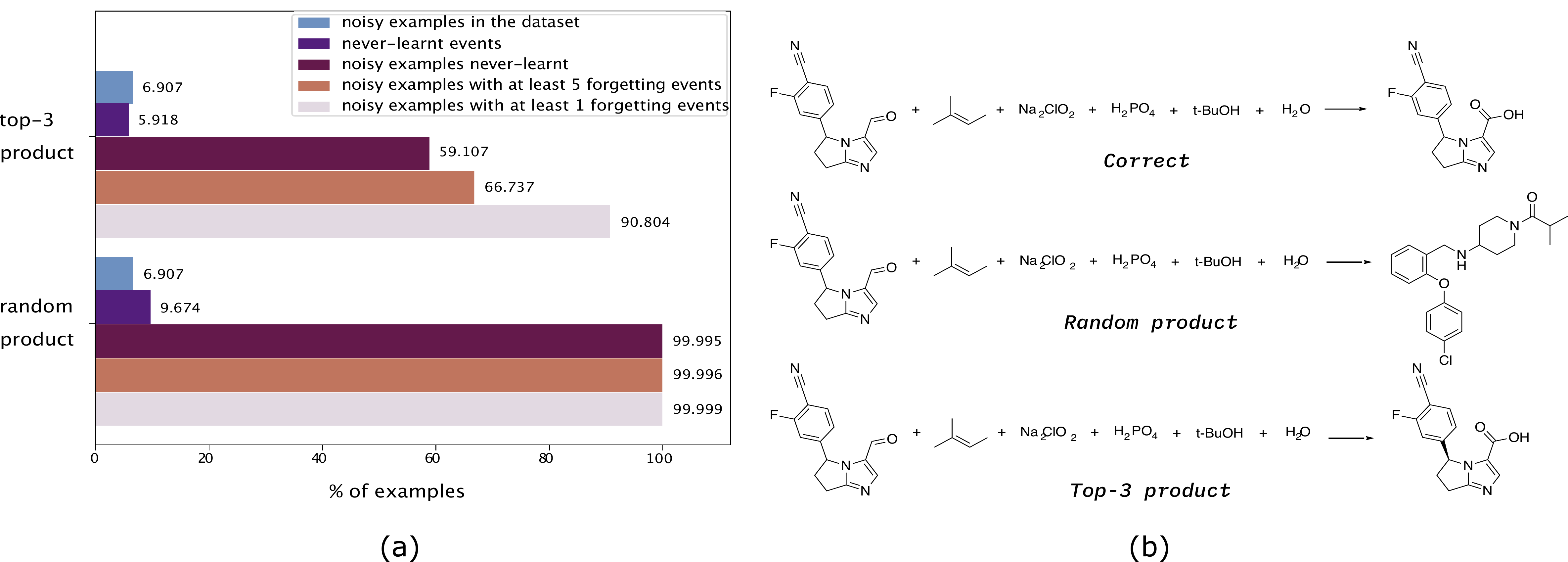}}
\captionsetup{width=.9\linewidth,font=footnotesize,labelfont=bf}
  \caption{\textbf{Artificial noise addition.} (a) Comparison of the two experiments with introduced noise. Shown in blue is the percentage of noise introduced into the data set, in purple the percentage of events which were never learnt. The bars 3-5 depict the percentage of noisy examples found in the never learnt events,  with at least 5 forgetting events and with at least 1 forgetting event respectively. "random product" is the noise that is easy to detect: the examples were generated by shuffling the products of the reactions in the validation set \cite{Segler2018}. "top-3 product" was the experiment where the noise introduction was more subtle: the target product in the validation set was substituted with the top-3 prediction of the same set. (b) Examples of the two kinds of noise introduced, shown for the same reaction SMILES. On top the correct reaction, in the middle the random product assigned. At the bottom the top-3 prediction (which shows a specific stereocenter that the reaction cannot generate).}
  \label{fig:noise}
\end{figure}

This is shown \black{in the bottom row of Figure \ref{fig:noise}.a, where all of the noise introduced falls} inside the 10\% of the whole training data set never learnt. 
For the second experiment, the introduced noise was slightly more subtle.
All original target products of the validation set, computed with the "forget-25perc" model, were substituted by the third entry of the top-3 predictions, which  is usually wrong. While being  more challenging, the forgotten event strategy correctly identified 60\% of the chemically wrong entries as never-learnt reactions. 90\% of the introduced noisy data experienced at least one forgotten event (Figure \ref{fig:noise}.a, top row). An example of this type of noisy reaction is given in Figure \ref{fig:noise}.b. In addition, we went through a random sample of 250 reactions removed by the forget-25\% model. The reactions can be found in the additional file to the Supporting Information, named \textcolor{blue}{human\_expert\_checked\_250sample\_of\_removed\_reactions.pdf}. 77.2\% of the 250 reactions were correctly identified as wrong by the model.

\black{A final comparison  of the newly trained forward model on the 25\% cleaned data set against the baseline was performed with the addition of some regularizing strategies. The complete plots are reported in Supporting Information Figure S5. }

The similarity metric (CJSD) for cumulative density functions is better for the group of models trained with the clean data set. In contrast, the top-1 accuracy remains high in both the baseline and the new forward model. It is reported for the original test set (where the predictions were hashed to identify tautomers and redox pairs) and for the test set \black{with all one-precursor reactions removed} (with one reagent/reactant only). These are mostly incorreclty text-mined reactions  and should consequently be considered irrelevant for the current evaluation. All reported metrics show that the cleaned forward model introduces a consistent improvement compared to the baseline.

\subsection{Comparison to random sampling}
\begin{figure}
\centerline{\includegraphics[width=150mm]{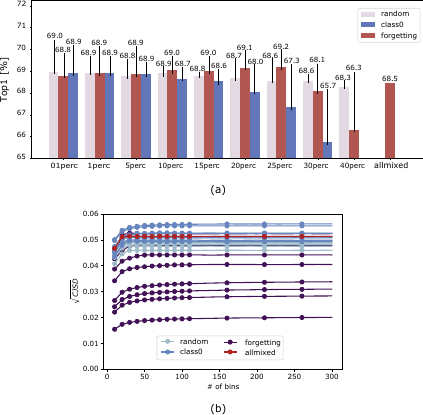}}
\captionsetup{width=.9\linewidth,font=footnotesize, labelfont=bf}
  \caption{\textbf{Comparison with null-models.} (a) top-1 accuracy on a common test set for the forgetting experiment and two different null models: random sampling and random sampling exclusively from class zero, \black{in the legend 'random' and 'class0' respectively}. (b) CJSD for the same models as in (a). The baseline model ('allmixed') is also reported. Note that the CJSD was computed without considering Resolutions (class 11).}
  \label{fig:null-model}
\end{figure}

In addition to the original baseline model, we performed an analysis of two different kinds of null models.
\black{For the first null model we randomly removed reactions}, chosen to mimic those of the forgotten events strategy regardless of the reaction class identity. \black{For the second model we removed random examples from the unrecognized class only}.
Figure \ref{fig:null-model}.a and \ref{fig:null-model}.b show the results of the random removal. 

\black{We observe a similar trend} between the three strategies until 15\% of the data set are removed, as shown in  Figure \ref{fig:null-model}.a.
Beyond the 15\% limit, the random sampling from \black{Unrecognized} only (labelled "class0") shows a drastic decrease in the test set top-1 results. This observation suggests that the removal of chemistry from the class of Unrecognized reactions can be detrimental \black{to the model and lead to a loss of meaningful chemical information. Therefore, there is a statistically significant difference applying the forgotten events strategy and randomly removing samples from regions of the data that are considered to be noisy. On the other hand, the plain random sampling experiment (labelled "random") does not lead to significant changes in the accuracy up to 40\% removal of the data set.}
\black{In contrast, the forgetting experiment gains almost 1\% in top-1 accuracy by removing 25\% of the entries identified as noisy. In the supporting information we also provide a table (\black{Supplementary Table 1}) containing the Wilson score intervals \cite{wilson-score} for the experiments in Figure \ref{fig:null-model}, which supports the results.}
The computation of the CJSD removes any doubt about the efficacy of the forgotten event strategy. \black{The two random sampling methods lead to a  negligible decrease in the CJSD} when compared to the forgotten events strategy, where the CJSD reaches a value of 0.02 (see Table \ref{table:percentages}).

\subsection{Application to an open data set}
\label{open-data-set}
We applied the same noise-reduction procedure to an open source data set\cite{Lowe2017} for reproducibility reasons. The results show the same behaviour and trends as the ones reported for the Pistachio \cite{Pistachio} data set. The top-1 accuracy improves slightly while the Cumulative Jensen Shannon divergence decreases significantly. Because of the reduced size of the open-source data set (approximately 1 million entries), we \black{already observed} the loss of  crucial chemical information when removing 10\% of the original data set, a smaller value than the one characterized for the Pistachio data set. We \black{report} the results for the open data set in the Supporting Information (Supplementary Figures 8-11).

\subsection{Retrosynthetic model noise reduction} \label{retro-synthesis-prediction}
Recently\cite{schwaller2020predicting}, we introduced  a novel statistical retrosynthetic strategy with new ranking metrics based on the use of the forward prediction model. In this scheme, the corresponding forward reaction prediction confidence plays a crucial role in determining which disconnections are the most effective among the entire set of single-step retrosynthetic predictions.

\begin{figure}
\centering
\includegraphics[width=.6\linewidth]{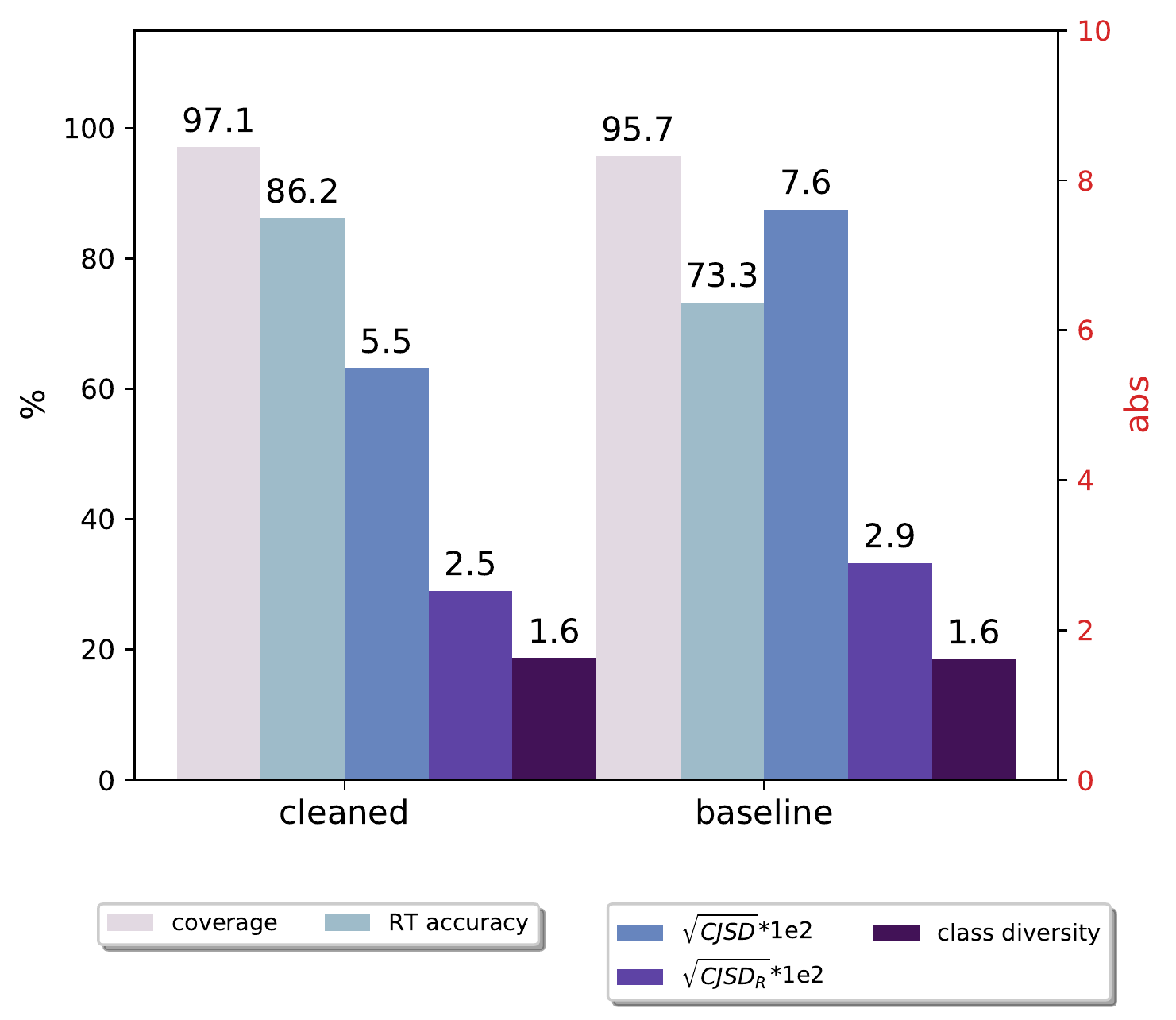}
\captionsetup{width=.9\linewidth,font=footnotesize, labelfont=bf}
  \caption{\textbf{Retrosynthesis models results.} Results for the baseline and new retrosynthesis model, trained on the Pistachio data set cleaned by forgetting forward prediction model. Four metrics are shown: \black{coverage (left y-axis), round-trip (RT) accuracy (left y-axis), $\sqrt{CJSD}$ (right y-axis) and class diversity (right y-axis)}. $\sqrt{CJSD_{R}}$ excludes the Resolution class}
  \label{fig:retrocomparison}
\end{figure}

In this context, the application of the noise-reduction schema to the retrosynthetic model\cite{schwaller2020predicting} \black{did not} lead to any significant noise reduction in the considered data set. In fact, the single-step retrosynthetic model operates only as a prompter of possible disconnections, ranked subsequently by the forward prediction model. Therefore, the noise-reduction strategy \black{was} effective only when used in combination with the forward prediction model. The corresponding noise-reduced data set \black{was} subsequently used to train the single-step retrosynthetic model.
Figure \ref{fig:retrocomparison} shows the performance of the new retro model compared to the baseline (trained with the data set non cleaned) \black{with regard to} to coverage,  class diversity,  round-trip accuracy \cite{schwaller2020predicting} and the square root of the cumulative Jensen-Shannon divergence, CJSD (see sections \ref{the-metrics-for-performance-evaluation} and \ref{cumulative-jensen-shannon-divergence} for details).

The coverage is high in both experiments, ensuring the existence of at least one valid disconnection. Class diversity does not show any degradation, indicating that the noise-reduction strategy does not affect the predictions' diversity. The round-trip accuracy\cite{schwaller2020predicting}, on the other hand, \black{improves} by almost 15\%: removing noisy entries allows both forward and retrosynthetic models to learn \black{more successfully} the difference between correct and wrong chemistry. In Figure \ref{fig:retrocomparison} we also report the CJSD, the measure of the bias towards specific classes. The CJSD decreases substantially, indicating a reduced bias in the confidence distributions across superclasses compared to the baseline.

\subsection{Noise-reduced model assessment}
We assessed the quality of the improved forward and single-step retrosynthetic models using the same set of chemical reactions used in Schwaller et al. \cite{schwaller2020predicting,SchwallerFWD}. The new forward prediction model achieves the same performance as the original model \cite{SchwallerFWD} (see SI, \black{Supplementary Figures 12 and 13}). 
\black{In the Supplementary Information (section 1.8), we share a critical discussion on the evaluation of the retrosynthetic model with the noise removed,} the compounds used for the evaluation,  the parameters settings and the predicted multistep routes. In addition, we considered some more retrosynthetic examples particularly challenging for the baseline model \cite{IBMRXN} that are discussed in the following below.

\begin{figure}
    \centerline{
    \includegraphics[width=170mm]{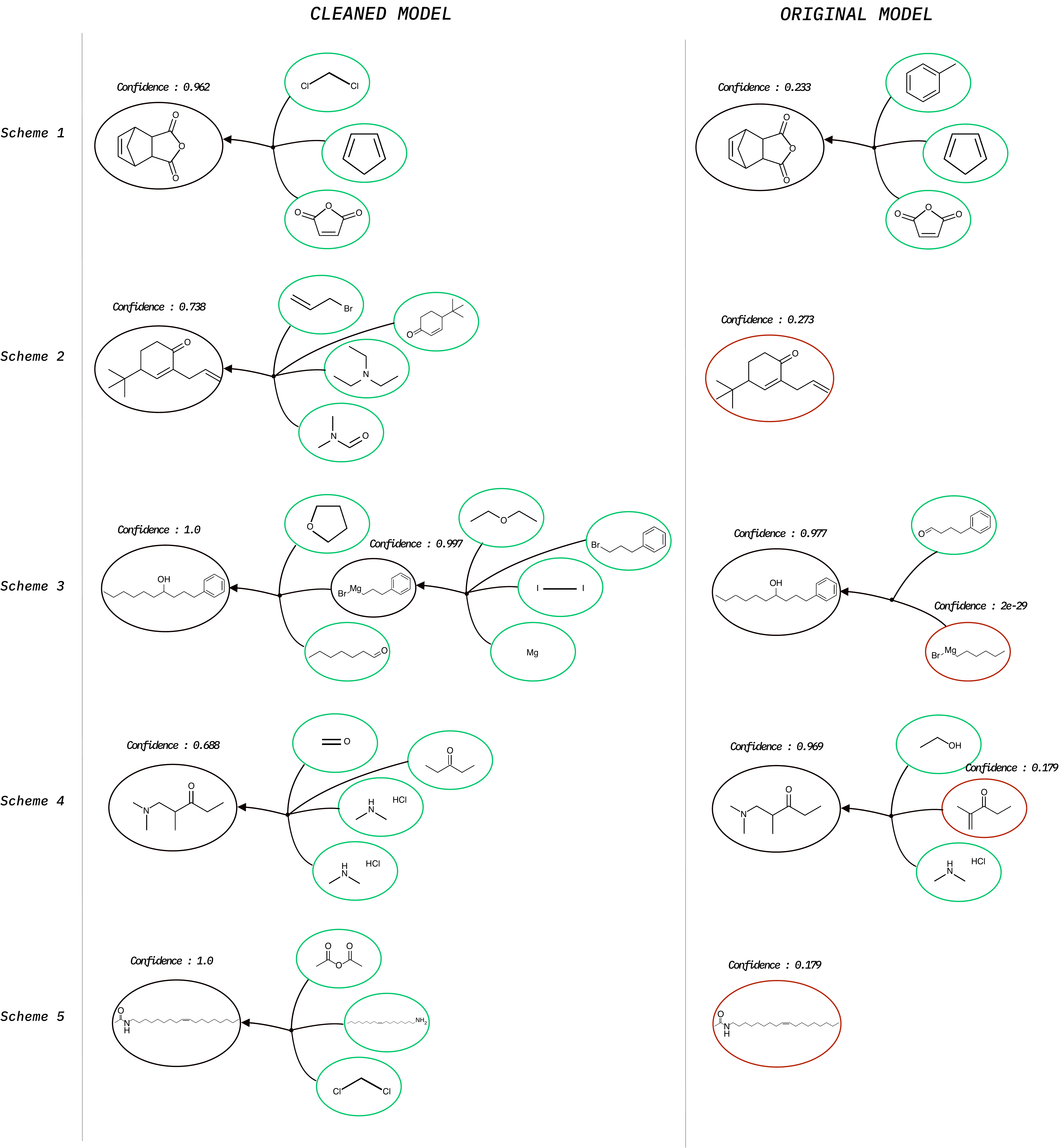}}
    \captionsetup{width=.9\linewidth,font=footnotesize, labelfont=bf}
    \caption{\textbf{Multistep retrosynthesis examples} Examples of retrosynthesis challenging for the original model, like the formation of an enolate or the correct synthesis of a reactive organometallic species. The middle column reports retrosynthesis predicted by the cleaned model. A tree structure is used to emphasize the algorithm structure. On the right, the results from the original model on the same compounds are reported. The precursors highlighted in green are successfully found as commercially available, while the red ones are not. Reported above each non-commercial molecule is the confidence that the model assigns to the set of precursors predicted: if the confidence falls below a certain threshold the algorithm does not carry on to the next step (this happens for two of the molecules with the original model). More details on the retrosynthetic algorithm can be found in Schwaller et al. \cite{schwaller2020predicting}}
    \label{fig:antonio_rxns}
\end{figure}

Figure \ref{fig:antonio_rxns} reports the analyzed compounds, together with the results for the baseline and improved models. \textbf{Scheme 1} shows the impact of the improved confidence distributions of the new model: the proposed Diels-Alder top-1 prediction has a much higher confidence for the cleaned model (0.962) than for the original  one (0.233). \textbf{Scheme 2} reports a reaction which involves the formation of an enolate. The cleaned model \black{proposes} a correct retrosynthetic strategy, whereas the original model \black{fails} due to the low level of confidence in all the proposed sets of precursors. Another interesting case regards a multistep synthesis that involves the preparation of reactive organometallic species, in particular non commercial Grignard (\textbf{Scheme 3}): the new model is clearly more confident about the predictions and this allows to complete the retrosynthesis of the compound. The same does not happen with the original model. It does not provide any suggestion after the synthesis of the metal-carbon bond. Regarding \textbf{Scheme 4}, the new model successfully completes the retrosynthesis for a Mannich reaction, proposing the correct set of reagents with a reasonable value of confidence. The original model \black{instead suggests} a first addition step with a secondary ammine ($\alpha$,$\beta$-unsaturated ketone), which \black{is most likely} not be the best choice even if chemically valid. In addition, for the original model the algorithm \black{stops} at the second step (the preparation of the $\alpha$,$\beta$-unsaturated ketone) due to low confidence values in the proposed disconnections. We also noted that the original model \black{assigns} rather low values of confidence (0.179) to reactions like \black{the one in} \textbf{Scheme 5}, where a primary ammine and acetyl anhydride are used. \black{The cleaned model, on the other hand, successfully synthesizes} the compound with high confidence.

\section{Conclusions}
In this work, we present the first unassisted, machine-learning based technique to automatically remove noisy chemical reactions from data sets. This methodology provides a statistical alternative to more tedious human or rule-based curation schemes. 
We applied the noise-reduction strategy to two  USPTO-based data sets, one open and one proprietary. We used the cleaned data sets to retrain forward, retrosynthetic, and classification models. \black{Statistical analysis of the results on the test set revealed an improvement in the values for all meaningful metrics (round-trip accuracy + 13\%, coverage + 1.4\%, CJSD down to 60\% of its original value for the forward model and 70\% for the retro model). We tested the strategy on previously addressed} chemical problems\cite{schwaller2020predicting,SchwallerFWD}, finding that the new model is able to either reproduce the baseline results or improve them by proposing valid alternatives. In most cases, the confidence of the new model on the retrosynthetic predictions increased (see Figure \ref{fig:antonio_rxns}). The results show that it is not the quantity but the quality of \black{the} knowledge embedded in training data sets \black{that} leads to more reliable models.  We hope the development of an unassisted noise-reduction protocol for improving the quality of existing data sets will have a significant impact on the further development of data-driven  models for chemistry.

\section{Methods}

\subsection{Transformer model}
\label{the-transformer-model}
As in previous work\cite{schwaller2020predicting,SchwallerFWD}, we used machine translation architectures to map chemical syntheses onto the world of machine learning models. The reactants, reagents and products were codified as simplified molecular-input line-entry system (SMILES) \cite{weininger1988smiles, weininger1989smiles} strings, tokenized, and fed to the Molecular Transformer\cite{SchwallerFWD}, the architecture based on the well known sequence-2-sequence transformer by Vaswani et al. \cite{vaswani2017attention}. The hyperparameters of the model were kept fixed throughout all simulations.
The transformer is made up of a set of encoder layers and a set of decoder layers. The tokens of the input SMILES string are encoded into (learnt) hidden vectors by the encoder. Those vectors are then fed to the decoder to predict the output sequence, one token at a time. Based on the work by Schwaller et al. \cite{schwaller2020predicting}, we decided to set the number of layers in both the encoder and decoder to 4 (size 384). The main characteristic of the transformer is the presence of multi-head attention and the number of these heads was set to 8.  Dropout was also included in the model at a rate of 0.1. An adam optimizer was used for loss minimization and the starting learning rate was set to 2. We used the OpenNMT framework \cite{opennmt} and PyTorch \cite{NEURIPS2019_9015} to build the models. 

\subsection{Data} \label{the-data}
The reaction data set used to experiment on the forgetting cleaning strategy was the \black{proprietary} Pistachio (release of Nov. 18th, 2019), derived \black{by} text-mining chemical reactions in US patents. Molecules and reactions were represented as SMILES strings. A first coarse filtering strategy was applied to the raw data set, which \black{contained} approximately 9.3 million reactions.
All reactions were parsed by RDKIT and canonicalized, then duplicates were removed.
Equal molecules in the same precursor set were made unique and the set was then alphabetically sorted and any duplicates removed. For consistency with the sequential design of retrosynthetic routes, multi product reactions were eliminated and for single product reactions only the largest fragment (as defined by RDKit \cite{greg_landrum_2019_3366468}) was kept.
Purification reactions were removed entirely. 

The pre-filtered data set was then randomly split into training, test and validation sets. \black{This procedure resulted in} 2`378`860 entries for training, 132`376 for validation and 131`547 for testing (90\%/5\%/5\%).
To enable a more exhaustive model evaluation the splitting was performed on unique products selections, to avoid similar examples being present in the training and test or validation. 
In the Supporting Information, \black{Supplementary Figure 1}, we report some statistics on the train, test and validation sets. 
Note that the test and validation sets were not subject to cleaning.

As in a previous work \cite{SchwallerFWD}, we choose not to distinguish between reactants and reagents because, even chemically, this splitting is not always well defined and can change with the tool used to make the separation. Each char in the SMILES string was codified as a single token and a new token, the '$\sim$', was introduced to model fragment bonds.

\subsection{Definition of forgotten events}
In the context of chemical reactions and retrosynthesis, the definition of forgotten events is of crucial importance. For forward prediction, given a data set of chemical examples $A = (x_{i}, y_{i})$, we denote as ${y}_{i}$ the predicted product given reactants/reagents ${x}_{i}$. The $acc_{i}^{t}$ is a binary variable denoting whether example $i$ is correctly predicted at time step $t$. \black{We define the end of the epochs as our time steps.} A $forward$ forgetting event occurs when a product molecule, previously classified as correct at $t-1$, is then incorrectly predicted at time step $t$. \black{On the other hand, a learning event occurs when a previously wrong prediction becomes correct in the next epoch}:
\begin{eqnarray*}
 \text{forgetting event} & & acc_{i}^{t} < acc_{i}^{t-1}\\
  \text{learning event}  & & acc_{i}^{t} > acc_{i}^{t-1}
\end{eqnarray*}

It can also occur that reactions are \textit{never learnt} or \textit{never forgotten}. The former are labeled as having an infinite number of forgetting events, the latter as having none. \black{While the definition of what is ``forgotten" is in line with reference \cite{2018forgetting} for the forward predictions, some modifications had to be made for the retrosynthesis definition.}

In fact, for retrosynthetic predictions, the top-N accuracy of the single-step retrosynthetic model is only a measure of how efficient the model is in memorizing data rather than extracting contextual knowledge. The goal is not to propose the most commonly reported sets of reactants and reagents that result in a certain target molecule, but to generate many chemically correct sets, ideally with a high diversity. In this context, the diversity is \black{regarded} as the number of diverse reaction superclasses predicted by the retrosynthetic model. A more appropriate metric would be the round-trip accuracy, obtained by reapplying back a forward prediction to the predicted precursor set and comparing the result with the original product molecule. In this sense, a learnt retro-event occurs when we can recover the product molecule. Following the formalism used for the forward model, we define a data set of chemical examples $D = (x_{i}, y_{i})$ where $x_{i}$ is a molecule to be synthesized and $y_{i}$ the target set of precursors. Again $\overline{y}_{i}$ denotes the model prediction (this time the precursor set). A $retro$ forgetting event occurs when a set of precursors, which led back to the product molecule at $t-1$, fails \black{to recover} this at time step $t$:
\begin{eqnarray*}
 \text{forgetting event} & & RTacc_{i}^{t} < RTacc_{i}^{t-1}\\
  \text{learning event}  & & RTacc_{i}^{t} > RTacc_{i}^{t-1}
\end{eqnarray*}
$RTacc_{i}^{t}$ denotes the round-trip accuracy, in binary form, of example $i$ at time step $t$.

\subsection{Metrics for performance evaluation} \label{the-metrics-for-performance-evaluation}
In order to evaluate a single-step forward and retrosynthetic prediction, appropriate metrics needed to be designed which do not rely on tedious, manual analysis by a human chemist. Here, we used: top-n accuracy, round-trip accuracy, class diversity and coverage, as reported in a previous work\cite{schwaller2020predicting}. In Schwaller et al.\cite{schwaller2020predicting}, the authors introduced the Jensen-Shannon divergence, while here we turned to its cumulative version, explained in detail in the next section.

\subsection{Cumulative Jensen-Shannon divergence} \label{cumulative-jensen-shannon-divergence}
\black{The} Jensen-Shannon divergence is a measure of how similar two or more discrete probability distributions are.
\begin{equation}
    JSD(P_{0},P_{1},...P_{N}) = H\left(\sum_{i=0}^{N}\frac{1}{N} P_{i}\right) - \frac{1}{12}\sum_{i=0}^{N}H\left(P_{i}\right)
\end{equation}
Where $P_{i}$ denotes a probability distribution and $H(P_{i})$ the Shannon Entropy for the distribution $P_{i}$. However, when \black{dealing} with continuous probability density functions the generalization of this formula is not straightforward.
The \black{difficulty lies in the definition of the entropy. The first expression proposed by Shannon \cite{Shannon48} to deal with a continuous case was the one of differential entropy:}
\begin{equation}
    H\left(P\right) = - \int p(x)\log p(x) dx
\end{equation}
the capital letter $P$ denotes the probability distribution and $p$ the probability density. This definition raises many concerns, as highlighted in detail by Murali et al. \cite{murali2004entropy}\black{. The} most important one is the fact that it is ``inconsistent", in the sense that a probability density function, differently from a probability function, can take values which are greater than 1. As a consequence, \black{entropies with a value of less} than 1 will contribute positively to the entropy, while those greater than one with a negative sign and the points equal to 1 will not contribute at all. This brings to an ambiguous definition of the information content carried by the differential entropy expression.
A way to overcome this issue is to define a new version of entropy which defines both continuous and discrete distributions. This quantity is known as Generalized Cumulative Residual Entropy \cite{murali2004entropy,Nguyen2015cjsd}:
\begin{equation}
    GCRE(P) = - \int_{-\infty}^{\infty}F(X>x)\log F(X>x) dx
\end{equation}
\begin{equation}
    F(X>x) = 1-P(X<x)
\end{equation}
The key idea is that $P(X<x)$ denotes the cumulative distribution and \black{not} the probability density function \black{any longer}.
\black{The advantages of using this definition are twofold}. First, there is no more inconsistency: the principle that the logarithm over a distribution is a measure of its information content is preserved. Second, we do not have to rely on probability density functions, which need to be estimated from the data with parametric methods like kernel density. This way, we can easily calculate, directly from the data, the cumulative distribution function, which is the "real" one representing the data (\black{given} that we have enough observations). If not enough observations are provided, the error in the calculation of the entropy will only \black{be} related to missing information and will not be dependent on the parameter used to estimate \black{the} probability density function.
As a consequence, a \textbf{cumulative Jensen- Shannon divergence} \cite{Nguyen2015cjsd} can be defined to compare cumulative distributions:
\begin{equation}
    CJSD(P_{0},P_{1},...P_{N}) = GCRE\left(\sum_{i=0}^{N}\frac{1}{N} P_{i}\right) - \frac{1}{12}\sum_{i=0}^{N}GCRE\left(P_{i}\right)
\end{equation}
With this new metric we were able to compare \black{the} information content of likelihood distributions extracted from different reaction \black{superclasses}. As extensively explained in a previous work \cite{schwaller2020predicting}, having a model with dissimilar likelihood distributions is equivalent to \black{having a} bias towards specific reaction \black{superclasses} over others. The model with the lowest CJSD value will be the one \black{that} has the most uniform likelihoods distributions. \black{Supplementary Figure 4} shows the cumulative distribution functions for the baseline and the cleaned forward model.

\subsection{The baseline model}\label{the-baseline-model}
For the baseline model of both forward and retro prediction, we trained the Molecular Transformer \cite{SchwallerFWD} directly on the 2.4 million reactions extracted from the coarse filtering of Pistachio.
The forward model used the tokenized \black{SMILES} of the precursors (reactants and reagents with no distinction), while having the SMILES string of the product molecule as a target. For the retro model the two were swapped.

First, we tried to establish a metric of comparison with our original models, which is the one currently running on the IBM RXN open source online platform for chemical synthesis prediction\cite{IBMRXN, SchwallerFWD}. These original models (both forward and retro) were tested both on the original and new test set. In Table \ref{table:oldnewmodel} the results can be compared to the performances of the new baseline models.

\begin{table}[]
\makebox[\textwidth][c]{
\begin{tabular}{@{}llllllllll@{}}
\textbf{Model} & \textbf{Train set} & \textbf{Test set}  & \textbf{Top-1}  & \textbf{Top-10} & \textbf{RT}    & \textbf{$\sqrt{\text{CJSD}}$} & \textbf{$\sqrt{\text{CJSD}_{R}}$} & \textbf{CD} & \textbf{COV} \\
\textbf{type} &   &   &  &      &     &      &  &    &        \\
forward & Pist2018 & Pist2018  & 71.86 & N/A     & N/A     & 0.066      & 0.048                     & N/A                & N/A         \\
forward & Pist2018 & Pist2019  & 68.41 & N/A     & N/A     & 0.123      & 0.047                     & N/A                & N/A         \\
forward & Pist2019 & Pist2019  & 69.32 & N/A     & N/A     & 0.095      & 0.052                     & N/A                & N/A         \\
retro & Pist2018 & Pist2018  & N/A     & 21.11  & 70.08  & 0.106       & 0.027                      & 1.5            & 89.77     \\
retro & Pist2018 & Pist2019  & N/A     & 20.33  & 71.26  & 0.106       & 0.028                     & 1.6            & 94.68     \\
retro & Pist2019 & Pist2019  & N/A     & 24.25  & 73.27 & 0.076      & 0.029                     & 1.6             & 95.71   
\end{tabular}
}
\captionsetup{width=.9\linewidth,font=footnotesize, labelfont=bf }
\caption{\textbf{Results for baseline models study.} Top-n accuracy of original models tested on the original and new test sets. For the forward models the top-1 and the $\sqrt{\text{CJSD}}$ are reported. $\sqrt{\text{CJSD}_{R}}$ is the cumulative JSD without the Resolutions. For retro models, top-10 is reported along with the four metrics presented in section \ref{the-metrics-for-performance-evaluation}: round-trip accuracy (RT), class diversity (CD), coverage (COV), and$\sqrt{\text{CJSD}}$. The $\sqrt{\text{CJSD}}$ is computed from the forward likelihood distributions generated by the cleaned forward model on the precursor set. Data set names and versions have been abbreviated: e.g. "Pist2018" stands for "Pistachio version 2018".}
\label{table:oldnewmodel}
\end{table}

This \black{comparison of} the performances was done in order to make sure that no significant difference arose from using the new release of Pistachio. \black{Indeed, we notice that the top-1 accuracy for the forward prediction models} clusters around 70\%. If the performance of the original model on the new test set \black{had been clearly better} than the one of the new baseline model, no conclusions on the comparison of the two could have been drawn, as the strategy used to extract the two sets was different. \black{An increased performance of the original model on the new test set, could have been a sign of a strong presence of examples from the new test set in its training set.}
Note that for a more appropriate comparison of the two models, the new baseline forward model was augmented with randomized reactions from the training set (in equal number to the training examples) as well as reactions extracted from textbooks. \black{This technique was also used in the previous work by Schwaller et al. \cite{SchwallerFWD}}.
For the retro model, we reported the top-10 accuracy, but more importantly the round-trip accuracy, which is already slightly better for the new baseline model (trained on Pist2019) on the new test set (Pist2019) compared to the original retro model (trained on Pist2018) on the same set. The coverage follows the same trend, while the class diversity is not affected. \black{It is important to point out that the same forward model was used for the round-trip evaluation of the retro in all three experiments (\textit{retro-Pist2018: test set Pist2018},  \textit{retro-Pist2018: test set Pist2019}, \textit{retro-2019: test set Pist2019}). It was the cleaned forward model called "forget-25perc" presented in section \ref{forward- synthesis-prediction}.}

\subsection{Retrosynthetic forgetting strategy: a random failure} \label{retro-random-failure}
To prove the random behaviour when cleaning data set using the single-step retrosynthetic model, we performed a simple statistical experiment. We used the forgetting forward strategy to label all the examples tagged for removal (20\% for the "forget-20perc", 25\% for the "forget-25perc", etc.) without actually removing them from the original data set.
Now that we had a label for the removed examples, if we were to draw some of them (random retro forgetting) from the data set without replacement we could describe the probability of drawing a "blue" sample with an hypergeometric distribution (\black{Extended Data Figure 1.a}). 

The probability of drawing all the labeled samples would be 0 for all cases and the highest for the percentages that correspond to the red points in \black{Figure Extended Data Figure 1.b}. \black{In this plot} we show the percentage of samples removed by the retro model that match those removed by the forward model. We see that the behaviour is almost random and \black{follows} the red points. Otherwise, \black{if it was not random,} we would have expected an overlap close to 100\% for all removed data sets.
\black{Additional proof for} the random behaviour of the forgetting retro can be seen in Figure \black{Extended Data Figure 1.c}. The one-precursor reactions, wrong chemical reactions with missing reagents, are detected and removed by the forgetting forward model, but not by the forgetting retro model (their percentage remains constant in the data set cleaned by the latter).

\section{Data availability}
\black{The data that supports the findings of this study is the reaction dataset Pistachio 3 (version release of Nov. 18th, 2019) from NextMoveSoftware \cite{Pistachio}. It is derived by text-mining chemical reactions in US patents. We also used two smaller open-source data sets: the dataset by Schneider et al. \cite{Schneider2016}, which consists of 50k randomly picked reactions from US patents, and the USPTO data set by Lowe \cite{Lowe2017} , an open data set with chemical reactions from US patents (1976-Sep2016). A demonstration of the code on the dataset by Schneider et al. is also available in the GitHub repository (below). Source data for the plots of the main manuscript can be found at \href{https://figshare.com/articles/journal\_contribution/Source\_Data/13674496}{\textcolor{blue}{https://figshare.com/articles/journal\_contribution/Source\_Data/13674496}}.}

\section{Code availability}
All code for data cleaning and analysis associated with the current submission is available at
\href{https://github.com/rxn4chemistry/OpenNMT-py/tree/noise\_reduction}{\textcolor{blue}{https://github.com/rxn4chemistry/OpenNMT-py/tree/noise\_reduction}} \cite{paper-doi}.

\section{Corresponding author}
For any question on the manuscript and the results email \textcolor{blue}{ato@zurich.ibm.com} (Alessandra Toniato).

\section{Acknowledgement}

We thank Matteo Manica for the help with the model deployment. We also acknowledge all the IBM RXN team for all the insightful suggestions.

\section{Authorship}
A.T. and P.S. conceived the idea and performed the experiments. A.T. verified the statistical results of the method. A.C. took care of the chemical evaluation of the results. J.G.  helped with the software implementation. T.L. supervised the project. All authors participated in discussions and contributed to the manuscript.

\section*{Competing interests}
\black{The authors declare no competing interests.}

\end{document}


\maketitle

\section{Supplementary Information}
\subsection{Data statistics}

\begin{figure}
\begin{subfigure}{.5\textwidth}
\includegraphics[width=1\textwidth]{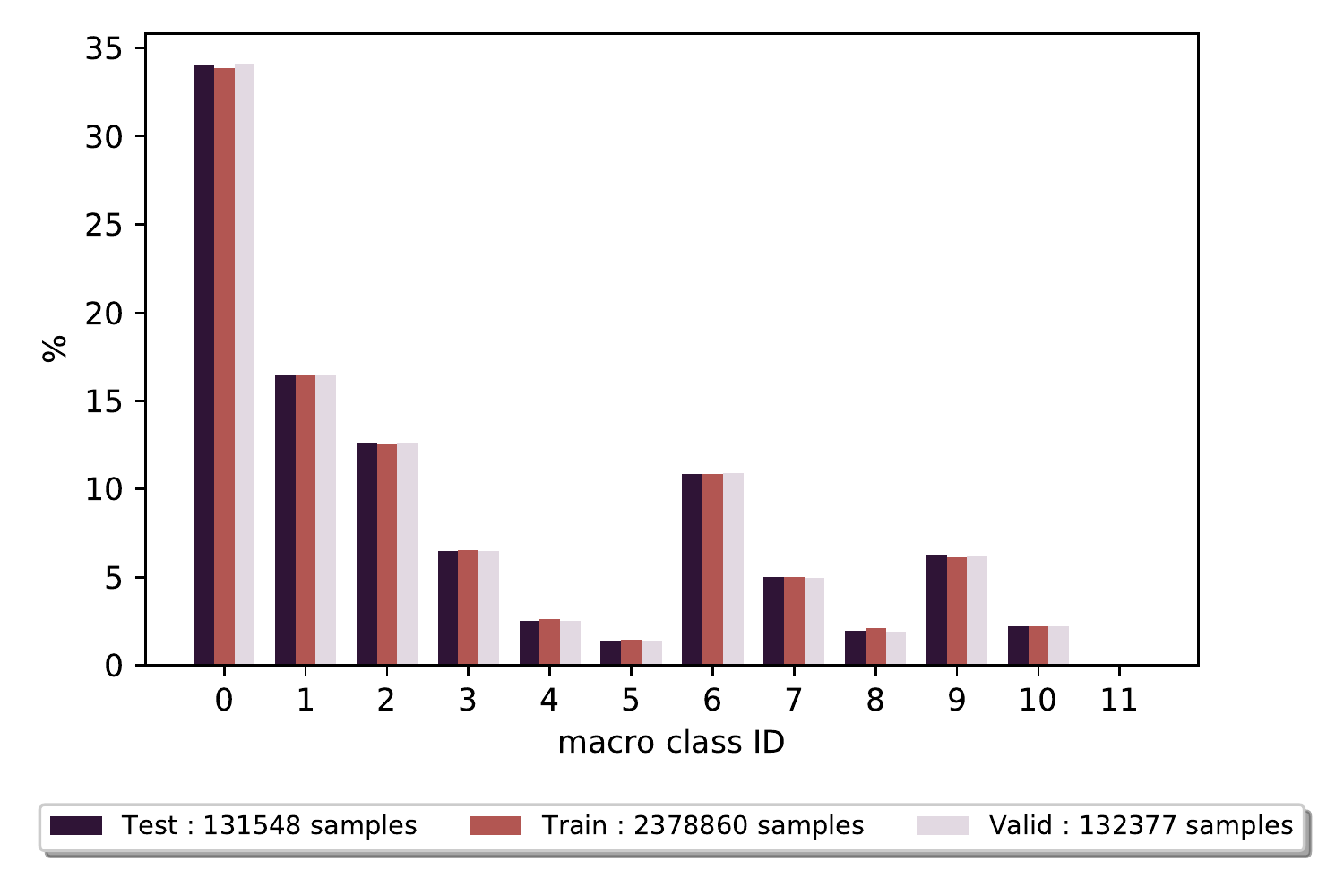}
\caption{}
\label{fig:data setsplit}
\end{subfigure}%
\begin{subfigure}{.5\textwidth}
\includegraphics[width=1\textwidth]{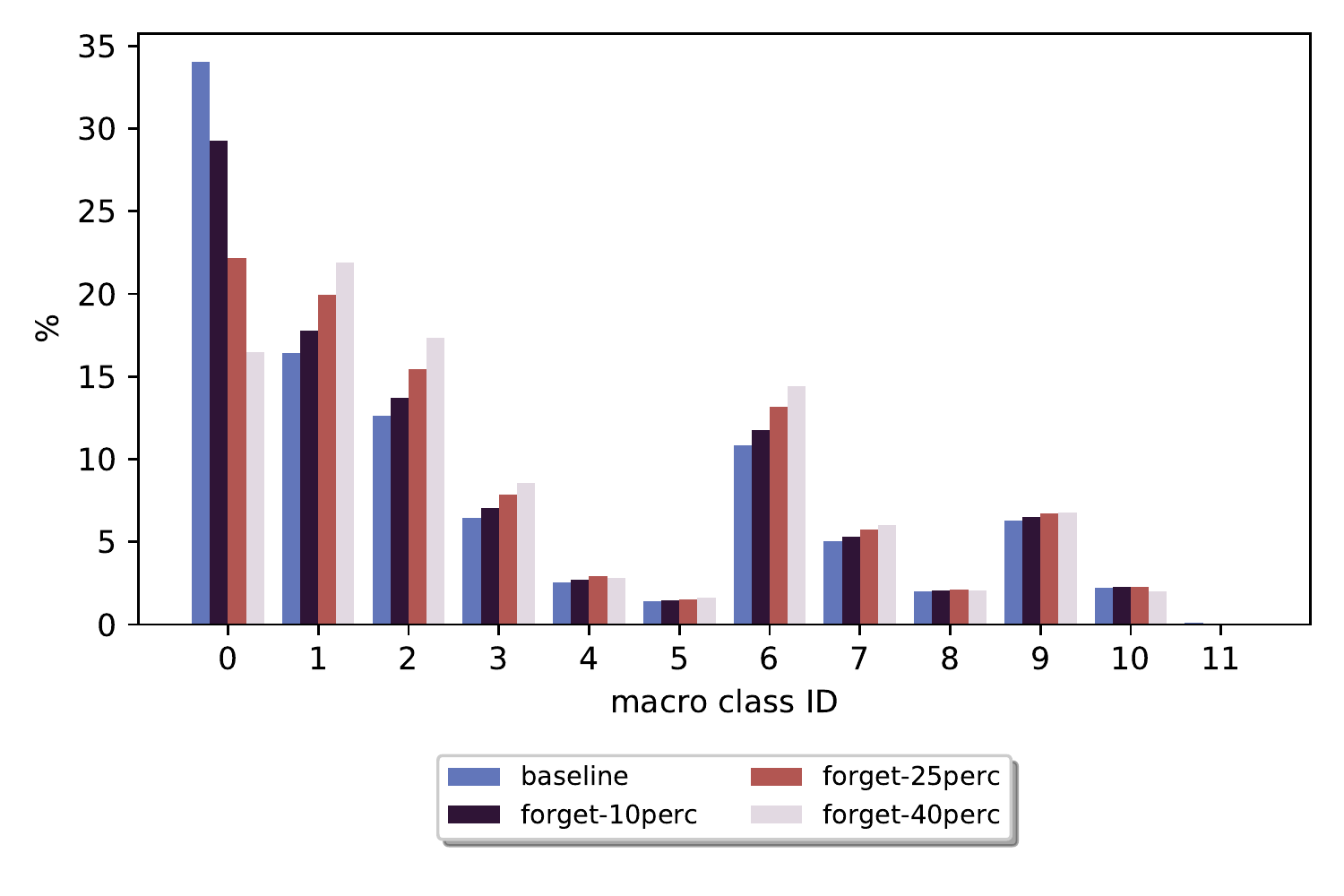}
\caption{}
\label{fig:data setsplit-forgetting}
\end{subfigure}
\captionsetup{width=.9\linewidth,font=footnotesize,labelfont=bf}
  \caption{\textbf{Data set statistics.}(a) How the different superclasses are populated in the three splits: train, validation and test set. (b) The balancing of the classes with the forgetting experiment: the baseline model is reported together with 3 of the cleaned models.}

\end{figure}
In Figure \ref{fig:data setsplit} we report the percentage of examples present in the three different splits, divided into the 12 macro classes. The same is reported for three different train data sets cleaned by the forward forgetting in Figure \ref{fig:data setsplit-forgetting} to show how the macroclasses get re-balanced.

\subsection{Analysis of the forgotten events}
\begin{figure}
\includegraphics[width=1\textwidth]{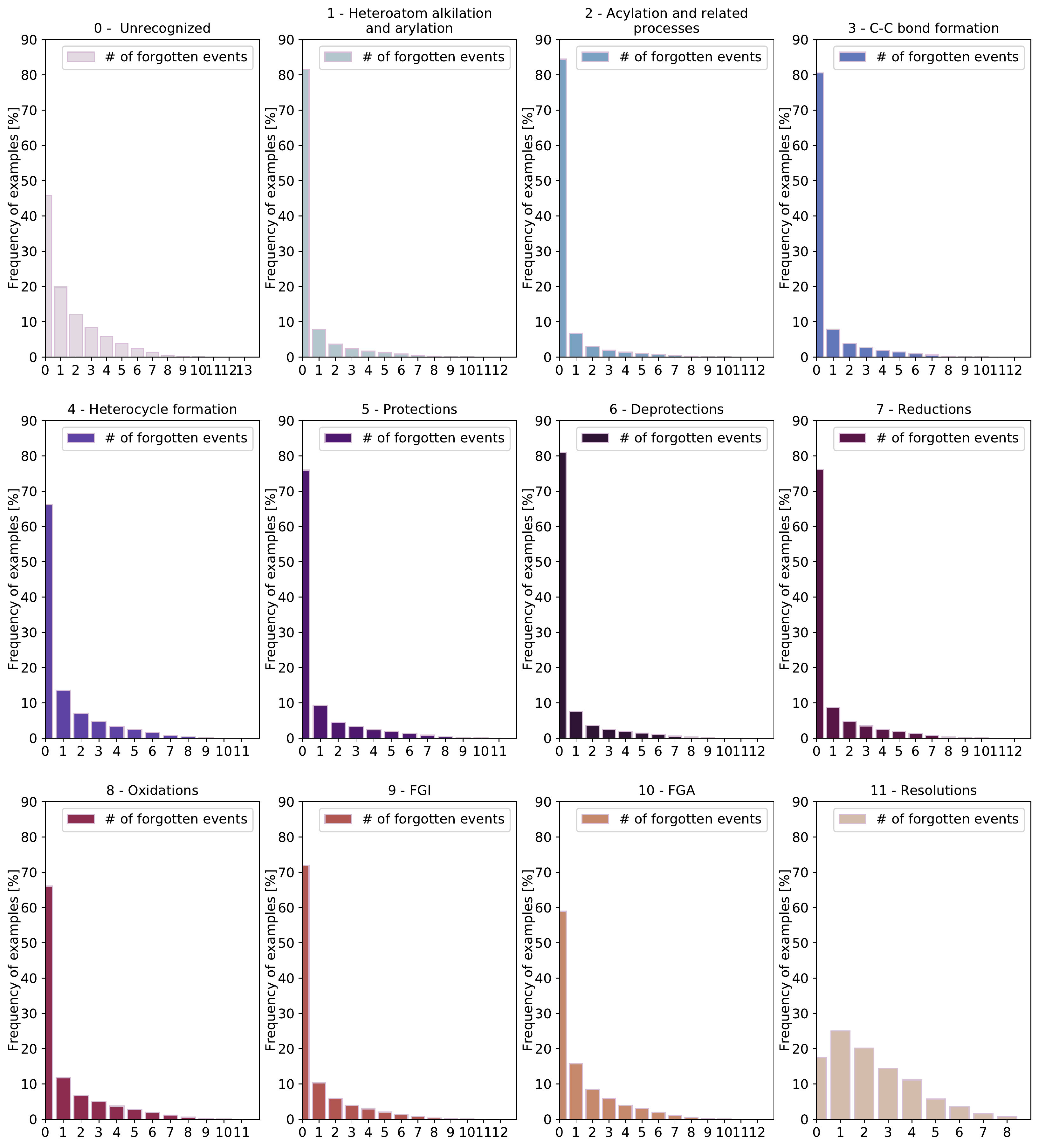}
\captionsetup{width=.9\linewidth,font=footnotesize,labelfont=bf}
  \caption{\textbf{Forgotten events distributions.}Number of forgotten events experienced by the training samples, divided by different classes.}
\label{fig:forgettingfwd-bis}
\end{figure}

\begin{figure}
\includegraphics[width=1\textwidth]{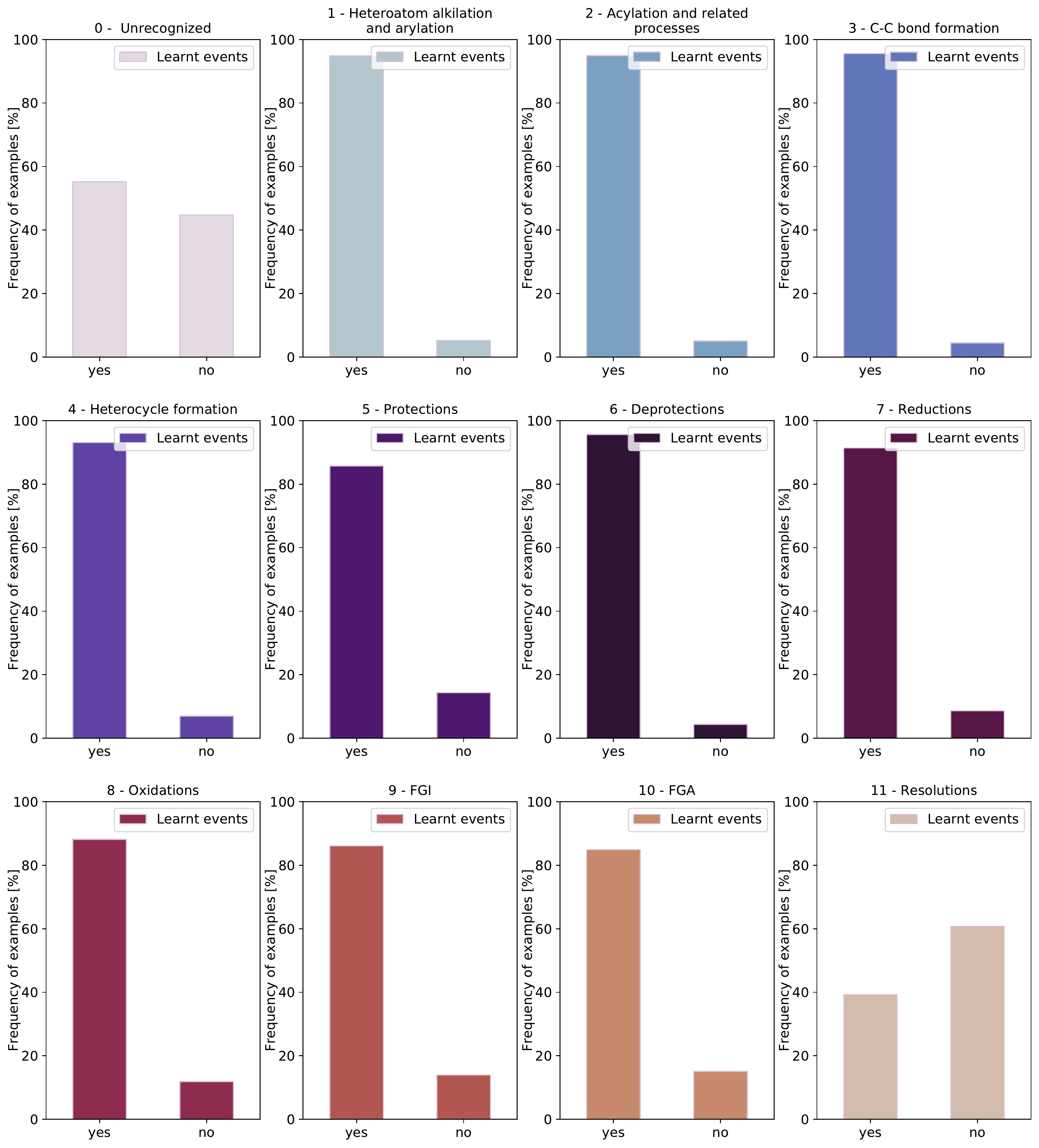}
\captionsetup{width=.9\linewidth,font=footnotesize,labelfont=bf}
  \caption{\textbf{Learnt and never-learnt events.}Percentage of events learnt and never-learnt during training, divided by different classes.}
\label{fig:forgettingfwd-bis2}
  
\end{figure}

\black{In Figure \ref{fig:forgettingfwd-bis} can be found the complete results for the number of forgotten events across superclasses. In Figure \ref{fig:forgettingfwd-bis2} are also reported the percentages of what was learnt and what was never-learnt across training.}

\clearpage

\subsection{Forward models comparison}
\begin{figure}
\begin{subfigure}{.5\textwidth}
\centering
\includegraphics[width=1\linewidth]{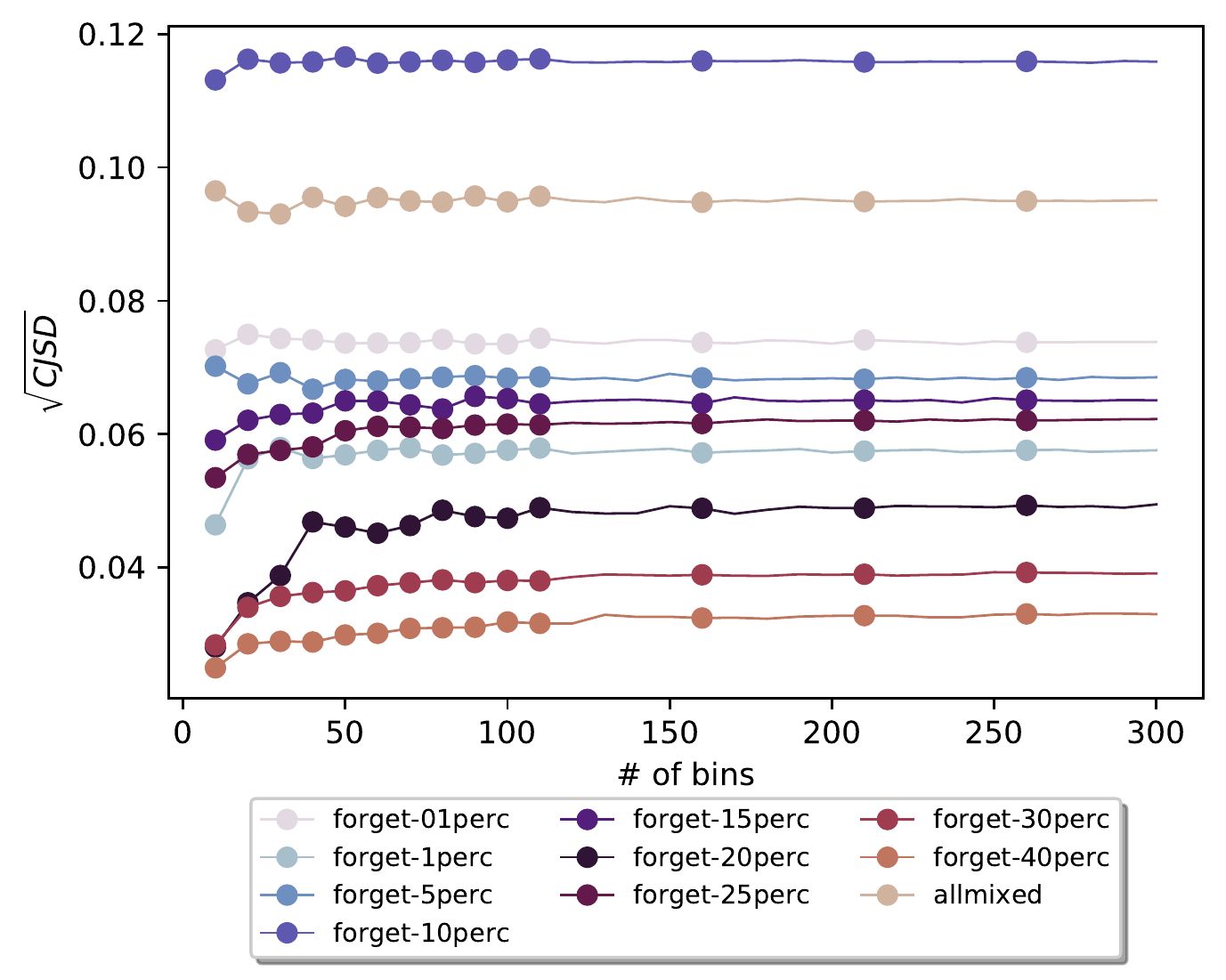}
\caption{}
\label{fig:class11}
\end{subfigure}%
\begin{subfigure}{.5\textwidth}
\centering
\includegraphics[width=1\linewidth]{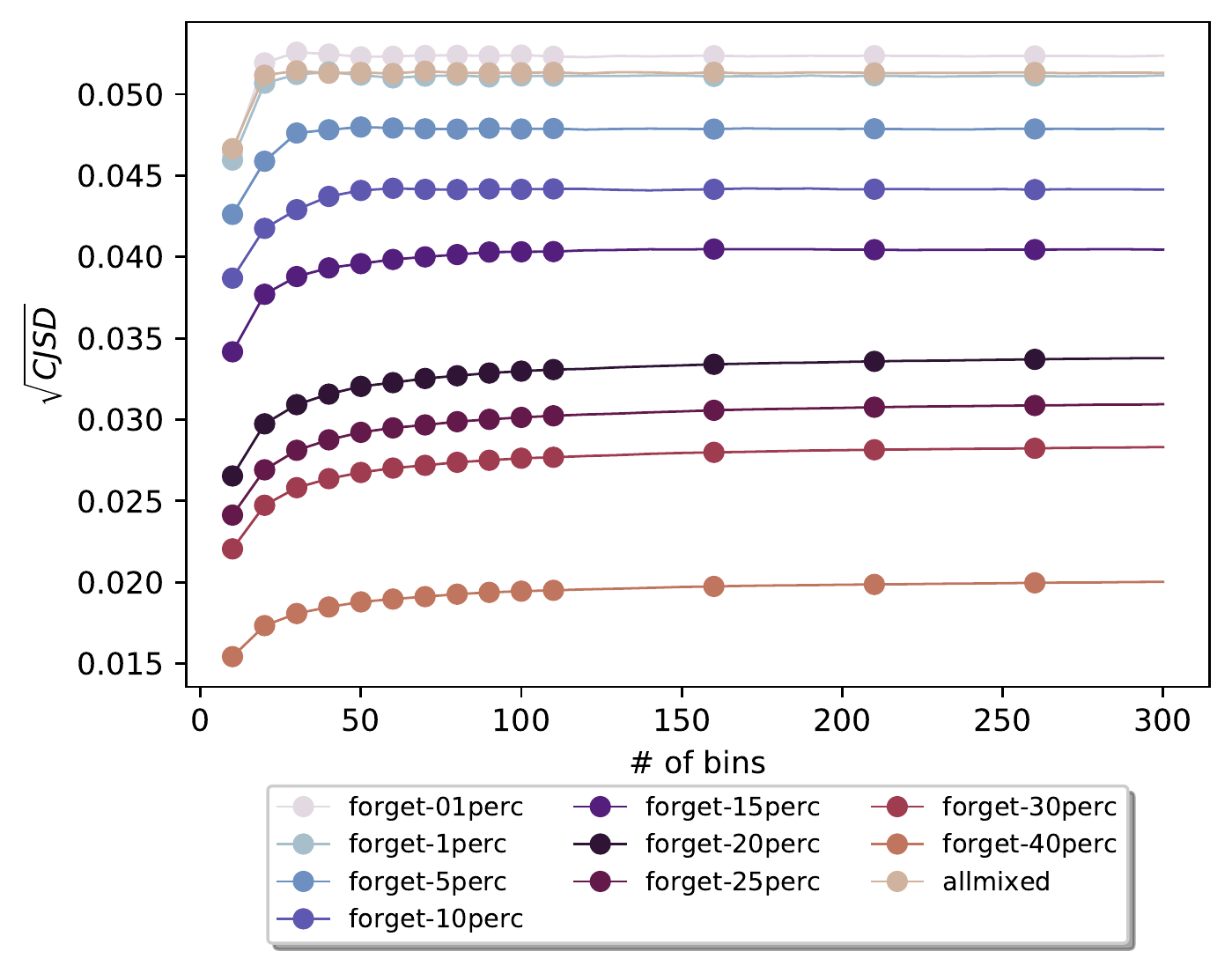}
\caption{}
\label{fig:noclass11}
\end{subfigure}
\captionsetup{width=.9\linewidth,font=footnotesize,labelfont=bf}
  \caption{\textbf{Uniformity metric for the forgetting experiment.} Square root of the Cumulative Jensen Shannon Divergence (CJSD) for the forward models trained with different portions of the data set removed. The distributions are compiled using the likelihoods of the test set entries correctly predicted. (a) CJSD as a function of the number of bins used to sample the cumulative distributions, considering the cumulative distributions of all 12 superclasses. (b) CJSD without considering the Resolution class. We report this analysis to provide an unbiased view of the true metric. The Resolution class is very weakly represented in the data set compared to all other classes, causing the calculated cumulative distribution to deviate substantially from the true distribution. This could introduce a strong bias in the calculation of CJSD. Despite this unbalance, the performance of the CJSD with and without accounting for the Resolution class shows a similar behaviour.}
  \label{fig:forgettingfwdcomparison}
\end{figure}

\begin{figure}
\begin{subfigure}{.5\textwidth}
\centering
\includegraphics[width=.8\linewidth]{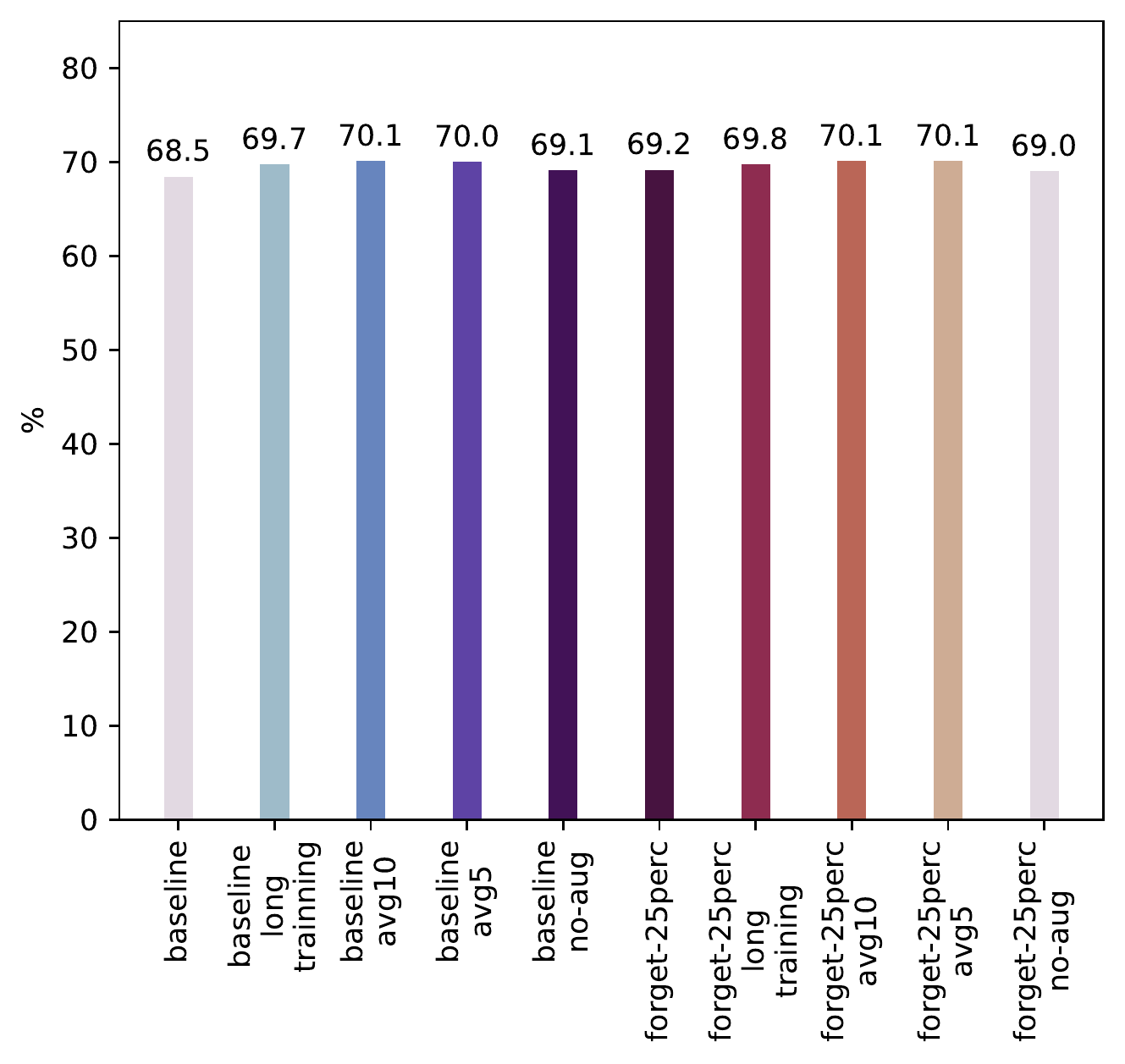}
\caption{}
\label{fig:last-forget-top1}
\end{subfigure}%
\begin{subfigure}{.5\textwidth}
\centering
\includegraphics[width=.92\linewidth]{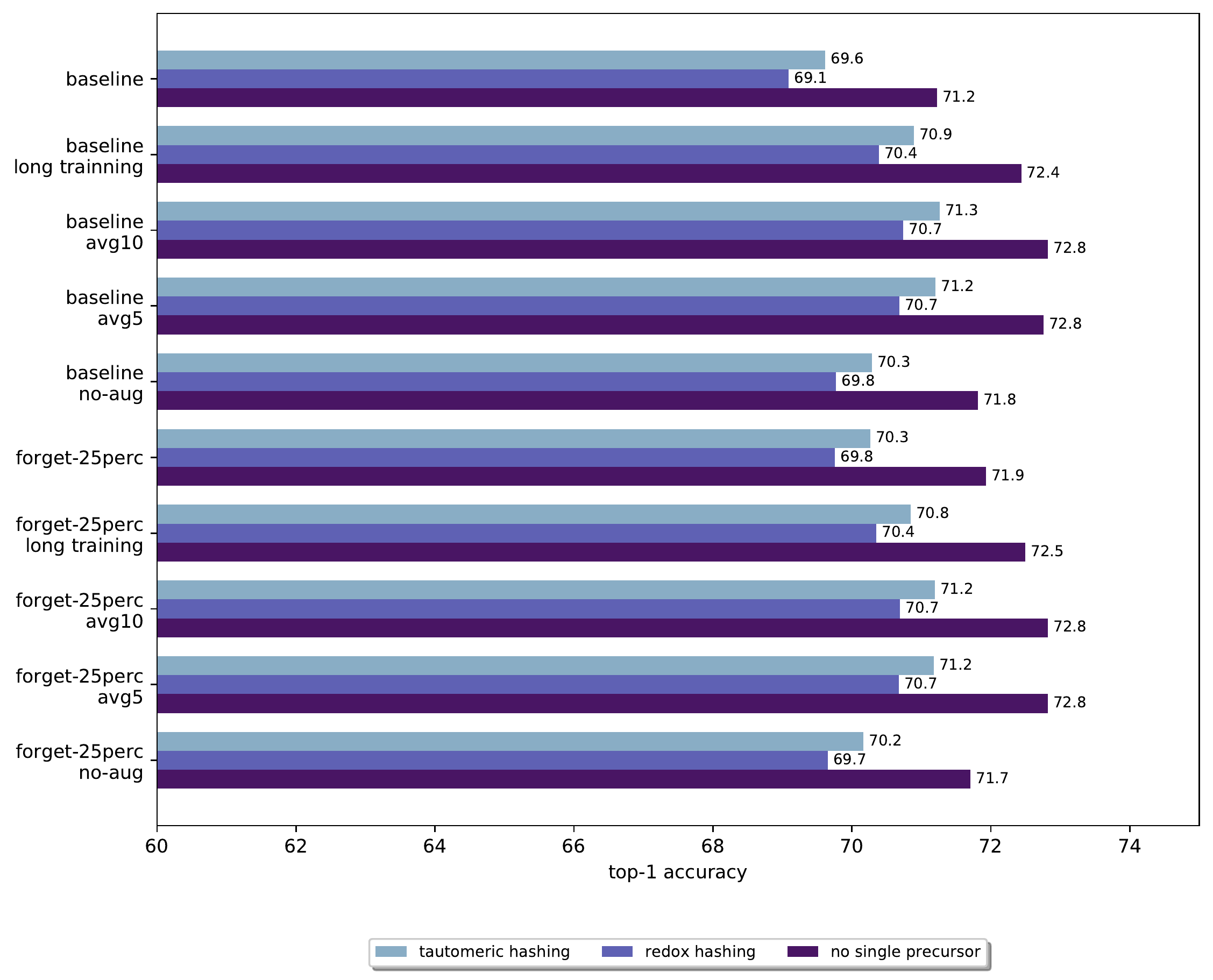}
\caption{}
\label{last-forget-tautomers}
\end{subfigure}
\begin{subfigure}{.5\textwidth}
\centering
\includegraphics[width=.9\linewidth]{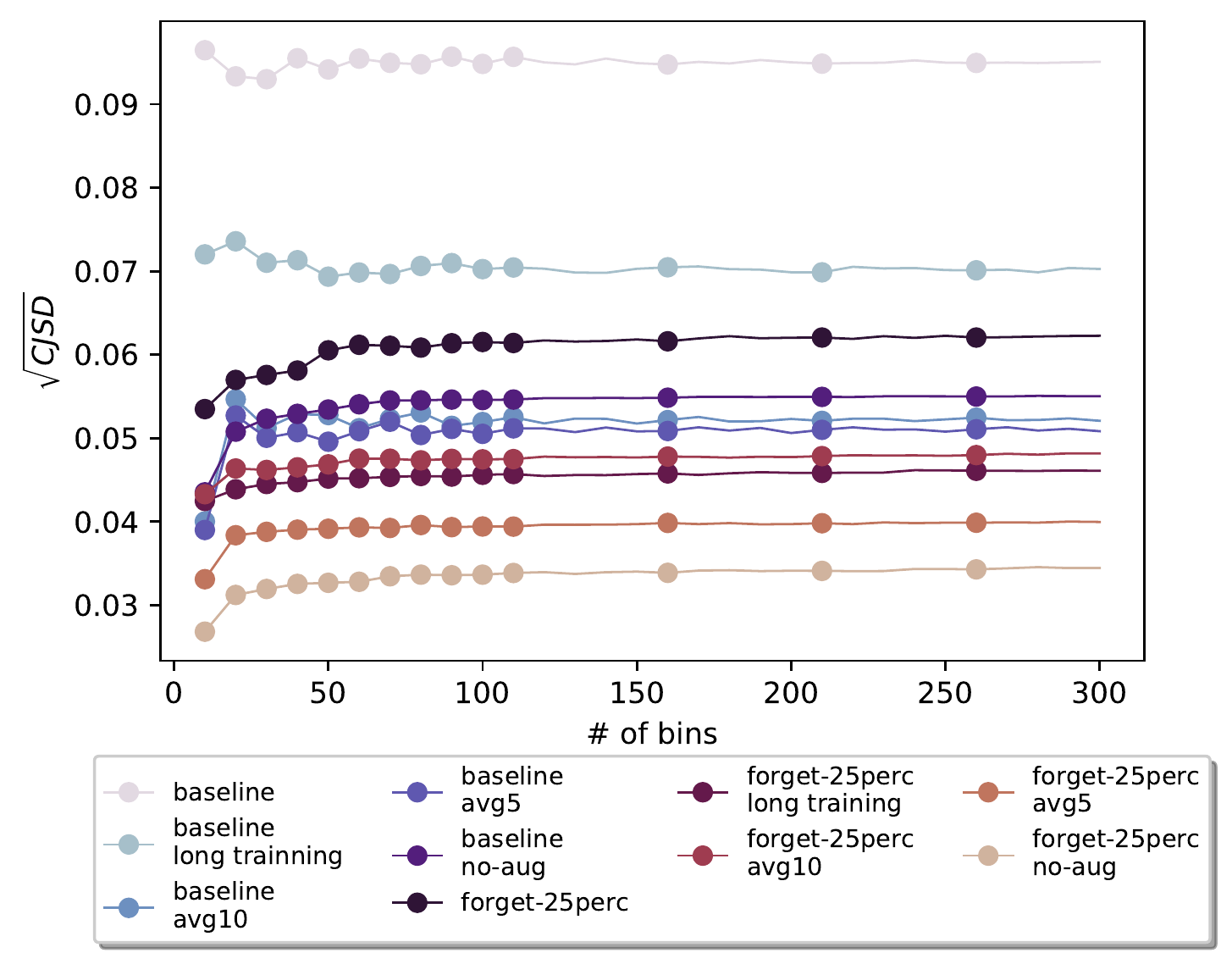}
\caption{}
\label{fig:last-forget-CJSD}
\end{subfigure}%
\begin{subfigure}{.5\textwidth}
\centering
\includegraphics[width=.9\linewidth]{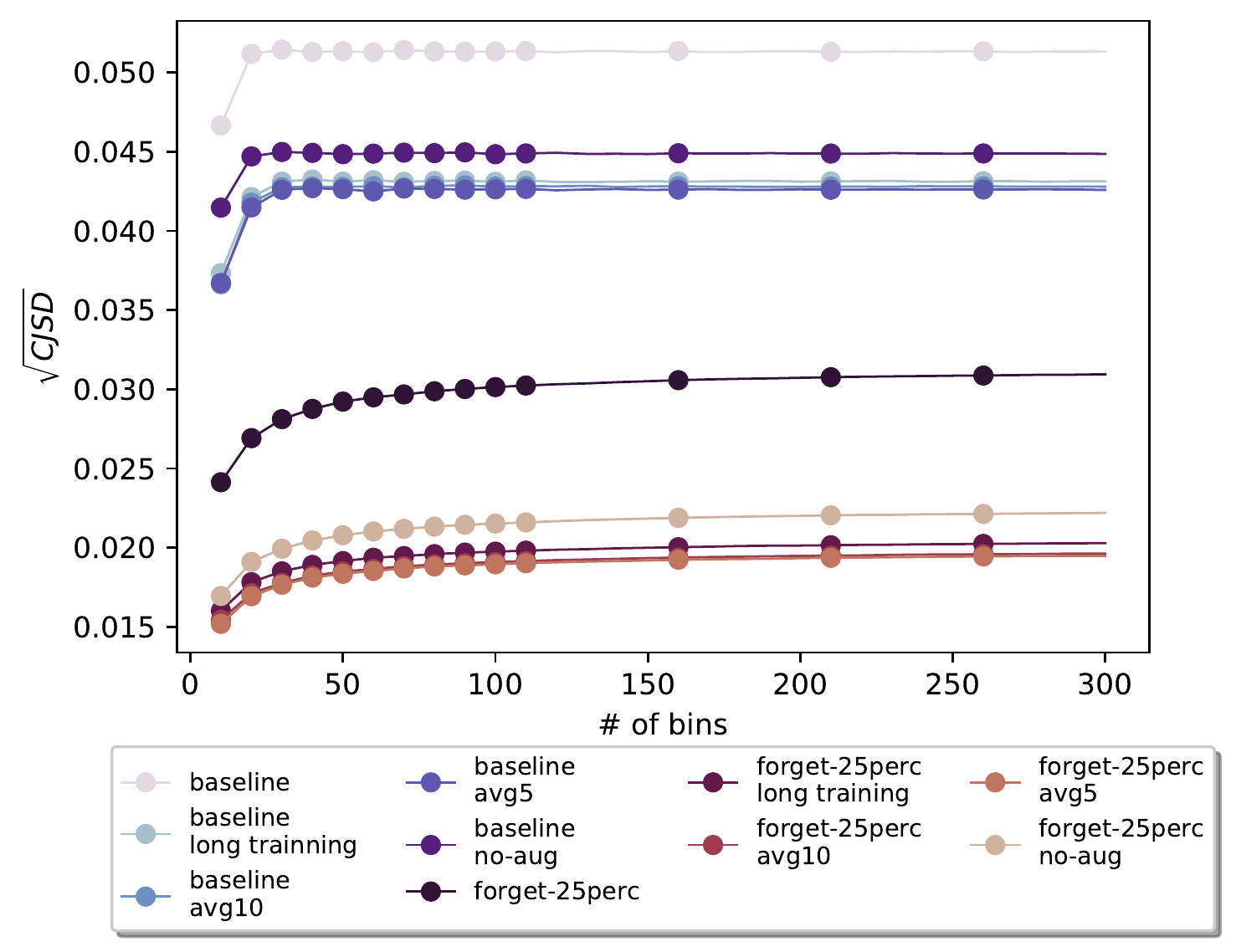}
\caption{}
\label{fig:last-forget-CJSD-cl11}
\end{subfigure}
\captionsetup{width=.9\linewidth,font=footnotesize,labelfont=bf}
  \caption{\textbf{Last models comparison.} Plots for the last comparison between models trained with the baseline data set and the cleaned one. (a) On the left some modifications of the baseline model. "baseline" is the baseline model augmented with randomized SMILEs \cite{SchwallerFWD}, "forget-25perc" is the cleaned model trained with the same augmentation strategy. The models which have "long training" appended to the label were trained for a longer period of time (96h), while the ones with "avg10" or "avg5" were obtained by averaging the last checkpoints, 10 and 5 respectively, of the models trained for longer. The same is done for the model trained with the cleaned data set ("forget-25perc"). (b) Top-1 accuracy  of the same models of where tautomers and redox hashing were taken into consideration. The RDKit utilities were used to perform the hashing. Also the top-1 where the one-precursor reactions were removed from the test set is reported . (c) Square root of the CJSD with all 12 classes. (d) Square root of the CJSD without the Resolutions.}
  \label{fig:lastforgettingfwdcomparison}
\end{figure}

 Figure \ref{fig:forgettingfwdcomparison} shows the behaviour of the square root of the cumulative Jensen Shannon Divergence (CJSD) for the reduced forward models. In Figure \ref{fig:lastforgettingfwdcomparison} is depicted instead the final comparison of the newly trained forward model on the 25\% cleaned data set with the baseline one.
 
\clearpage
\subsection{Cumulative distributions}

\begin{figure}[!]
\begin{subfigure}{.65\textwidth}
\includegraphics[width=1\textwidth]{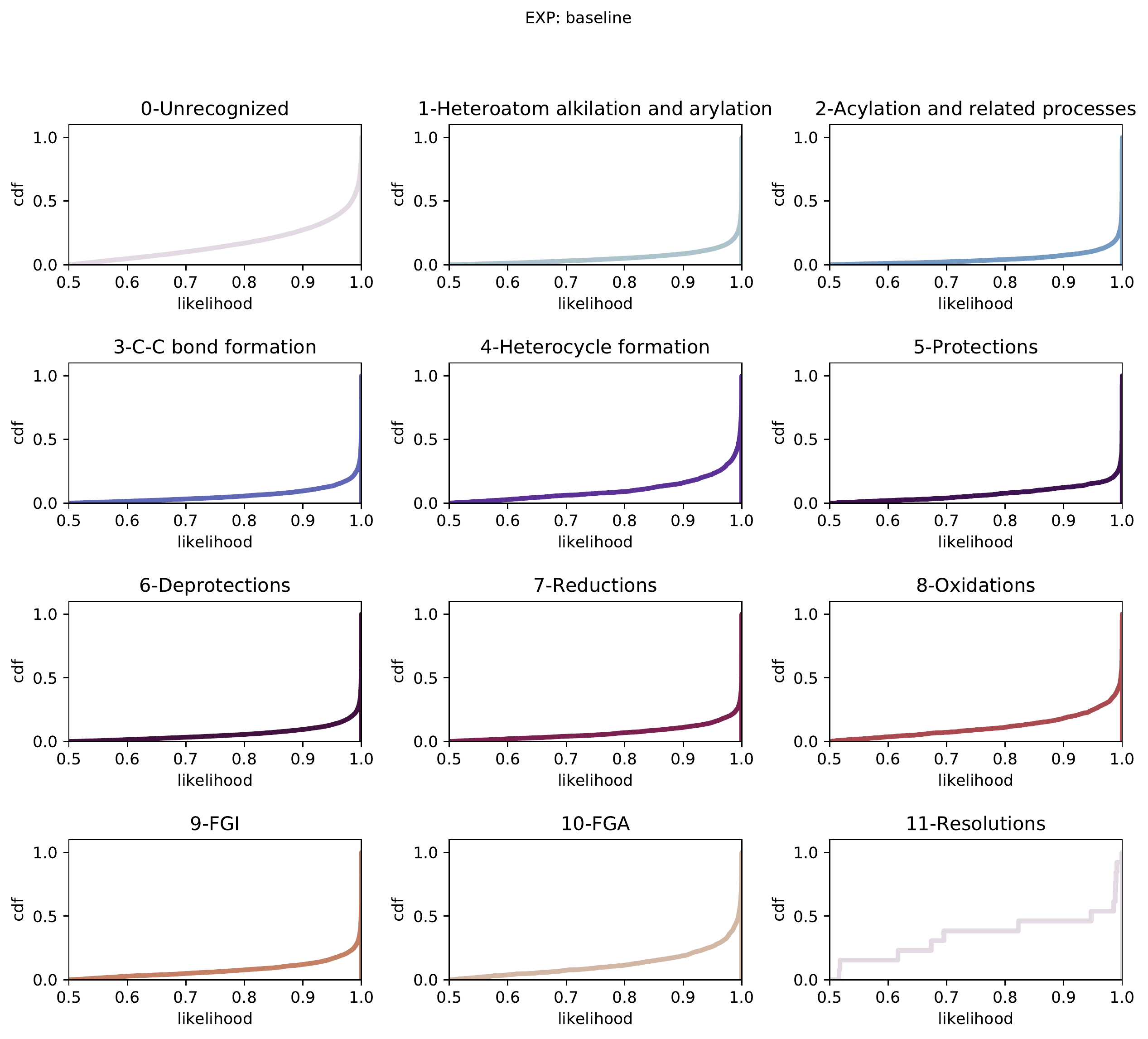}
\caption{}
\end{subfigure}
\begin{subfigure}{.65\textwidth}
\includegraphics[width=1\textwidth]{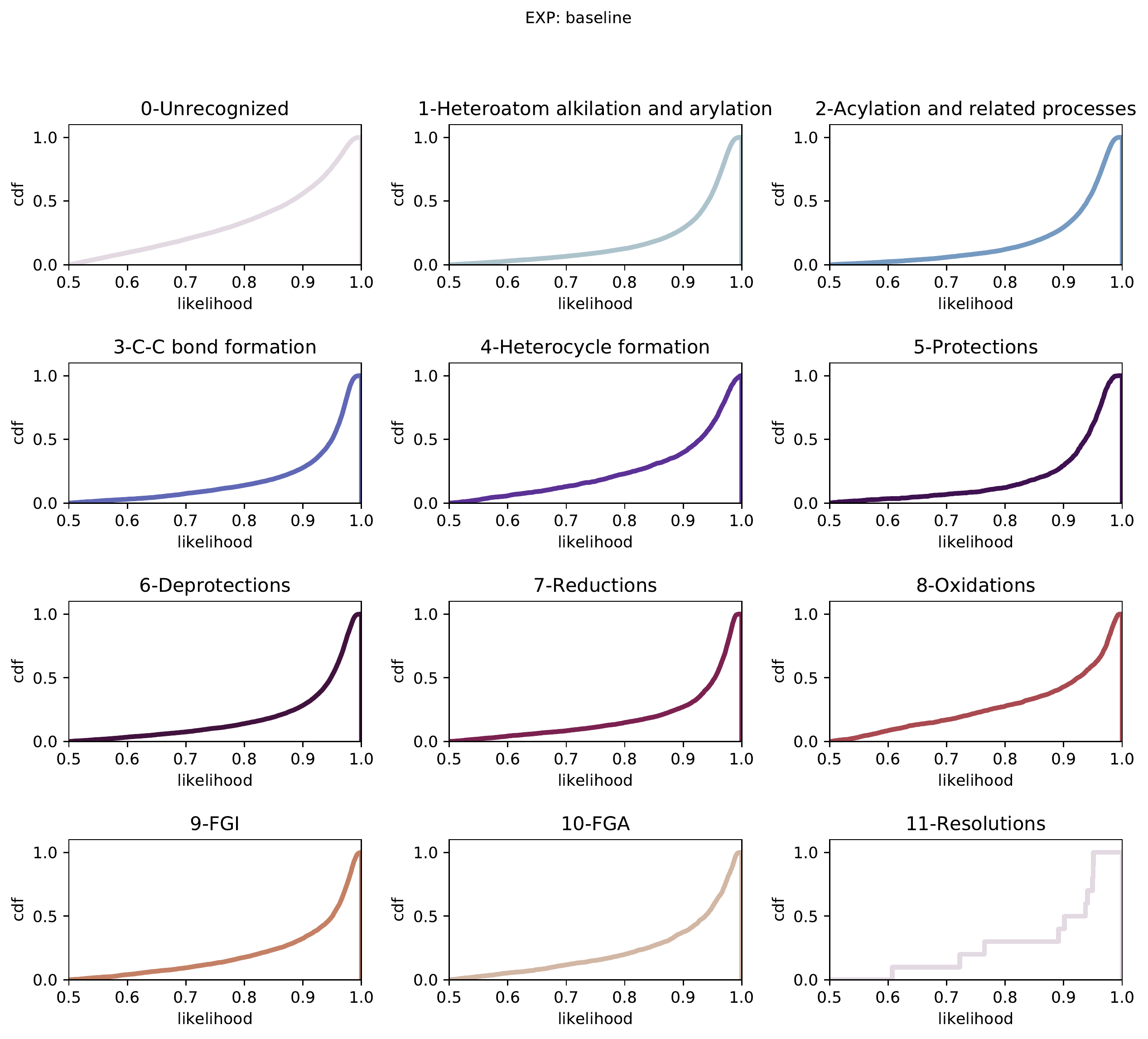}
\caption{}
\end{subfigure}
\captionsetup{width=.9\linewidth,font=footnotesize,labelfont=bf}
  \caption{\textbf{Cumulative distributions comparison.} Cumulative distribution functions divided by macro classes of the correctly predicted samples from the test set (forward model). (a) cleaned model: the distributions are more peaked towards 1.0 because removing noisy samples allows the model to be more sure of the predictions. (b) baseline model.}
\label{fig:cdfs}
\end{figure}
In Figure \ref{fig:cdfs} can be found the cumulative distribution functions for the baseline and the cleaned forward model divided by the 12 macro classes. These are the distributions of correctly predicted samples from the test set used in the evaluation of the CJSD. 
Note that the cumulative distributions for the retrosynthesis were constructed from the forward likelihoods obtained by applying back the forward model to the top-10 set of precursors.
\subsection{Wilson confidence intervals}
\black{In Table \ref{table:wilson} are reported the Wilson confidence intervals \cite{wilson-score} for the top-1 accuracies of the models in Figure 4 of the main manuscript.}

\begin{table}[!h] 
\centering 
\begin{tabular}{l l l l l}      %
\hline\\
\textbf{removal type} & \textbf{\% of removed} & \textbf{top-1 fraction} & \textbf{lower-bound} & \textbf{upper-bound} \\
\hline
forgetting & 0.1\% & 0.6879 & 0.6854 & 0.6904 \\
forgetting & 1\% & 0.6890 & 0.6864 & 0.6914 \\
forgetting & 5\% & 0.6888 & 0.6863 & 0.6913 \\
forgetting & 10\% & 0.6903 & 0.6878 & 0.6928 \\
forgetting & 15\% & 0.6903 & 0.6878 & 0.69285 \\
forgetting & 20\% & 0.6915 & 0.6890 & 0.6940 \\
forgetting & 25\% & 0.6918 & 0.6893 & 0.6943 \\
forgetting & 30\% & 0.6810 & 0.6785 & 0.6835 \\
forgetting & 40\% & 0.6631 & 0.6606 & 0.6657 \\
random & 0.1\% & 0.6896 & 0.6871 & 0.6921 \\
random & 1\% & 0.6891 & 0.6866 & 0.6916 \\
random & 5\% & 0.6878 & 0.6853 & 0.69036 \\
random & 10\% & 0.6890 & 0.6865 & 0.6915 \\
random & 15\% & 0.6879 & 0.6854 & 0.6904 \\
random & 20\% & 0.6866 & 0.6841 & 0.6891 \\
random & 25\% & 0.6856 & 0.6831 & 0.6881 \\
random & 30\% & 0.6855 & 0.6830 & 0.6880 \\
random & 40\% & 0.6826 & 0.6801 & 0.6851 \\
class0 & 0.1\% & 0.6893 & 0.6868 & 0.6918 \\
class0 & 1\% & 0.6892 & 0.6866 & 0.6917 \\
class0 & 5\% & 0.6889 & 0.6864 & 0.6914 \\
class0 & 10\% & 0.6865 & 0.6840 & 0.6890 \\
class0 & 15\% & 0.6857 & 0.6832 & 0.6882 \\
class0 & 20\% & 0.6804 & 0.6779 & 0.6830 \\
class0 & 25\% & 0.6734 & 0.6709 & 0.6759 \\
class0 & 30\% & 0.6573 & 0.6548 & 0.6599 \\
allmixed(baseline) & N/A & 0.6846 & 0.6821 & 0.6871 \\

\hline  
    \end{tabular}
    \captionsetup{width=.9\linewidth,font=footnotesize,labelfont=bf}
    \caption{\textbf{Winson confidence analysis.} Table reporting the Winson confidence intervals \cite{wilson-score} for the top-1 accuracy of the models trained with three removal techniques: forgetting, random and class0 random removal of samples. }
    \label{table:wilson}
\end{table}

\clearpage
\subsection{Statistical analysis of forgotten examples}
\label{statistical-analysis-of-forgotten-examples}

\begin{figure}

\begin{tabular}{@{}ll@{}}
\textbf{Seed}  & \textbf{Pearson}\\
42vs10  & 0.8812  \\
42vs20   & 0.8810 \\
10vs20  & 0.8831  
\end{tabular}%
\begin{subfigure}{.41\textwidth}
\centering
\includegraphics[width=.9\linewidth]{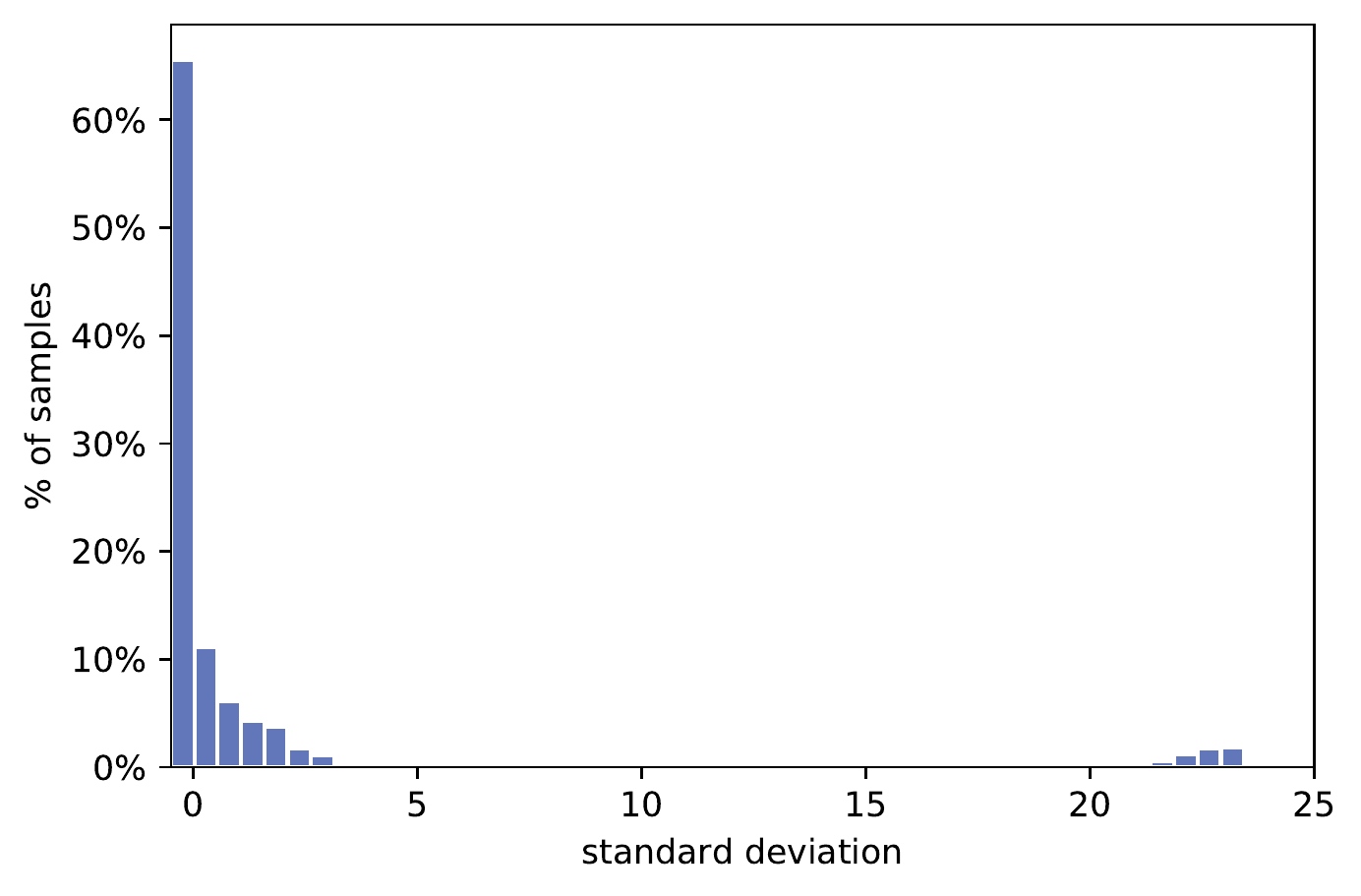}
\caption{}
\label{fig:seedstd}
\end{subfigure}%
\begin{subfigure}{.41\textwidth}
\centering
\includegraphics[width=.9\linewidth]{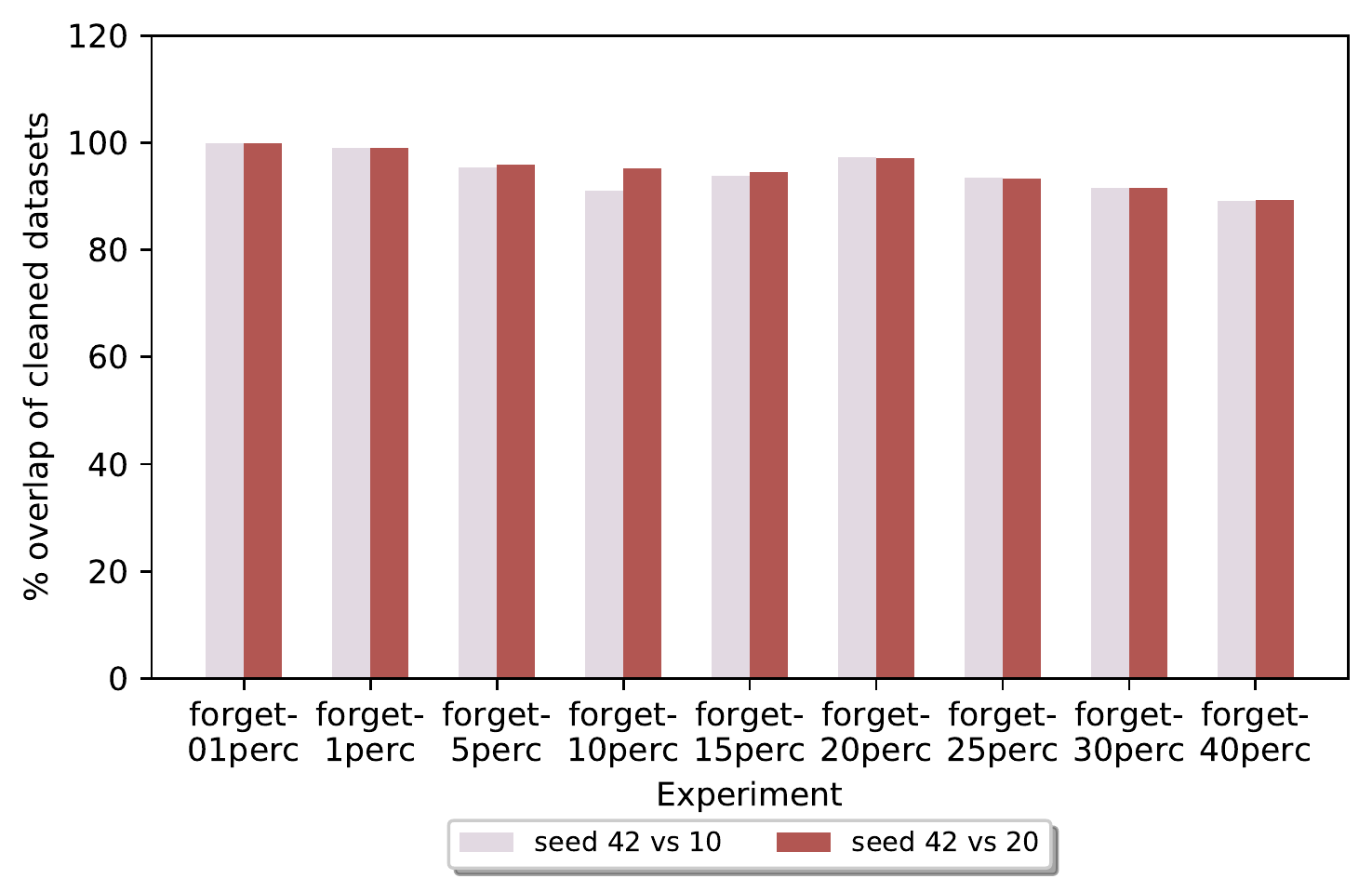}
\caption{}
\label{fig:seedoverlap}
\end{subfigure}
\captionsetup{width=.9\linewidth,font=footnotesize,labelfont=bf}
  \caption{\textbf{Statistical analysis.} On the left the Pearson coefficient calculated for couples os seeds. (a) Standard deviation of the number of forgotten events calculated for each sample across the three seeds. The concentration of outliers around 20/25 is due to the fact that never learnt events, nominally $\infty$ forgotten, were defined as forgotten 50 times in order to compute a stdv. Indeed, a minor percentage of examples in some of the seeds is never learnt and in others is learnt with a high forgetting rate. (b) Overlap between the removed data sets in different percentages across seeds. The overlap oscillates around 100\%, indicating the stability of the methods across seeds.}
  \label{fig:seedcomparison}
\end{figure}

Concerning the forward forgetting experiment, some statistical evaluations were performed in order to make sure that the computation of how many times an example was forgotten/learnt was not completely random. 
First, two models were trained with a different seed (10, 20) to the original one used (42) and again the number of forgotten events was computed. These counts were then analyzed across seeds. The table in Figure \ref{fig:seedcomparison} reports the Pearson coefficient computed for couples of seeds. By being close to one it indicates a high correlation between the numbers. Also the standard deviation though all three seeds behaves well (Figure \ref{fig:seedstd}), with most of the examples concentrating around 0 and some outliers between 20 and 25. The position of the outliers depends on the number chosen (n = 50) to substitute the "$\infty$" denoting the events never learnt (this substitution was necessary to compute the standard deviation). Also the overlap of the removed sets, in all of their percentages shows its stability in oscillating around 100\% (Figure \ref{fig:seedoverlap}). 

\subsection{Application to an open-source dataset}
\begin{figure}
\centerline{
\includegraphics[width=120mm]{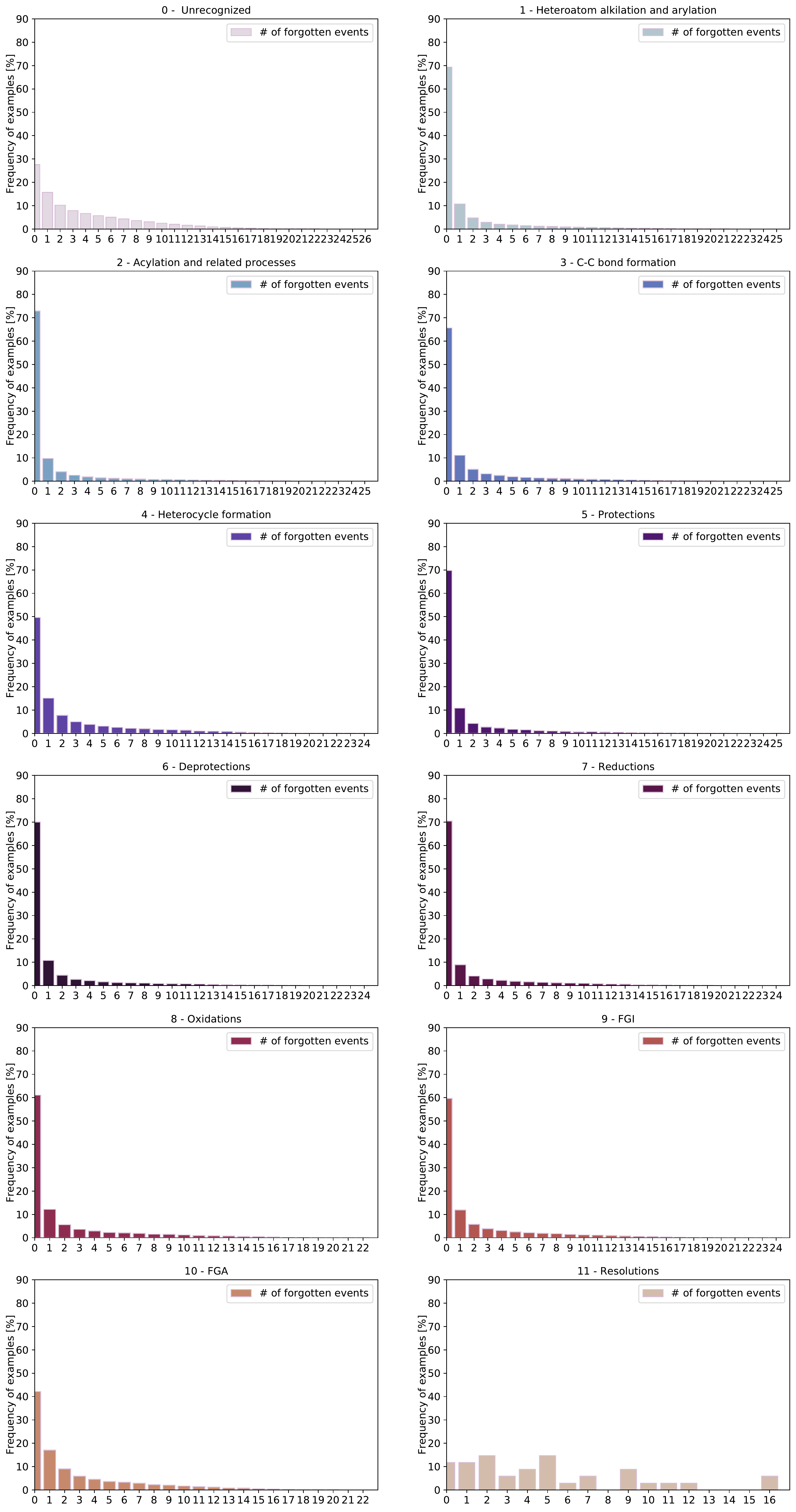}
}
\captionsetup{width=.9\linewidth,font=footnotesize,labelfont=bf}
  \caption{\textbf{Forgotten events distribution for the Lowe dataset.} Number of forgotten events divided by macroclasses for the Lowe dataset \cite{lowe2012extraction}.}
  \label{fig:stereo-forgotten-events}
\end{figure}

\begin{figure}
\centerline{
\includegraphics[width=170mm]{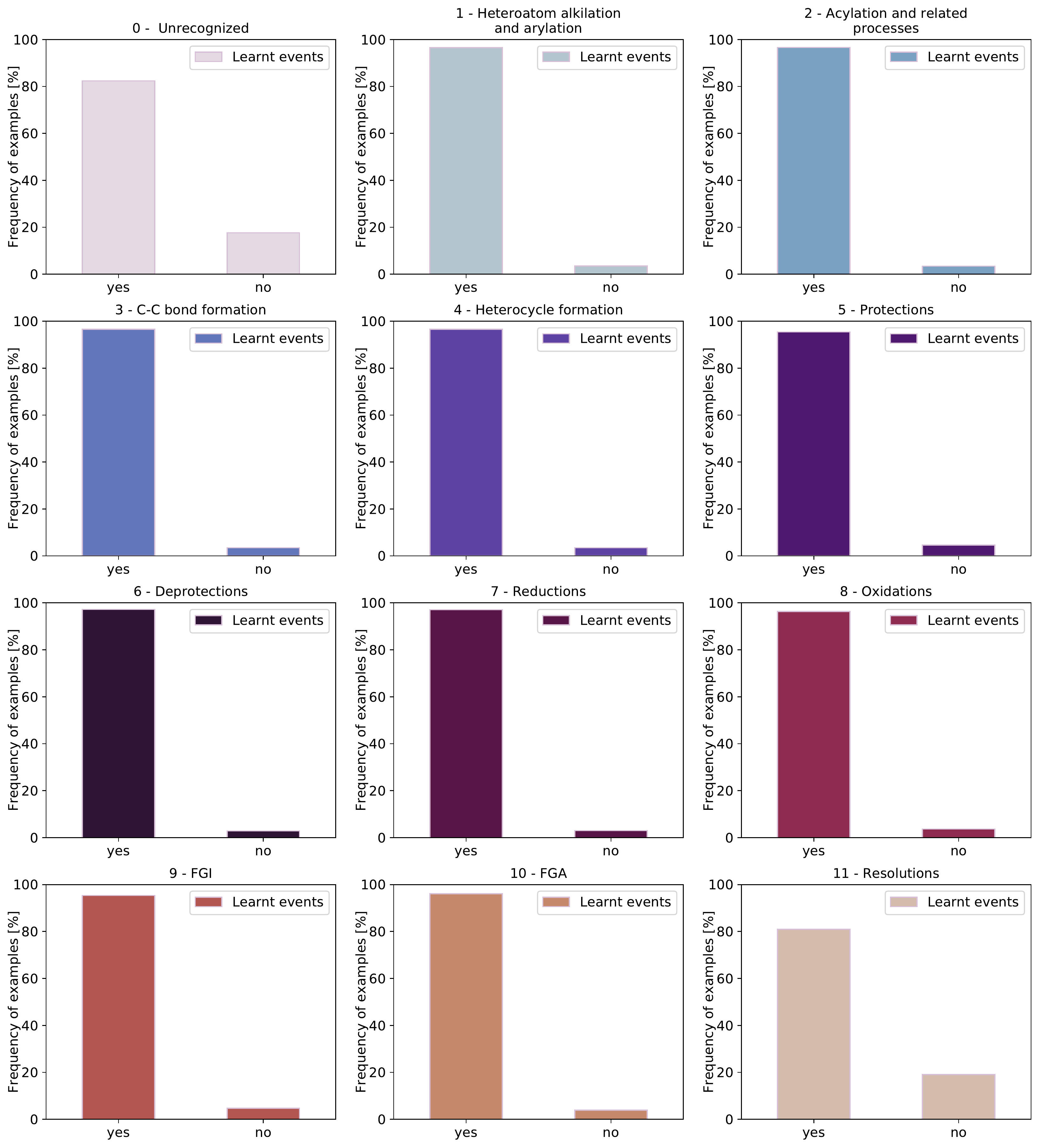}
}
\captionsetup{width=.9\linewidth,font=footnotesize,labelfont=bf}
  \caption{\textbf{Learnt and never-learnt events for the Lowe dataset.} Learnt and Never-learnt events divided by macroclasses for the Lowe dataset \cite{lowe2012extraction}.}
  \label{fig:stereo-never-learnt-events}
\end{figure}

\begin{figure}
\centerline{
\includegraphics[width=150mm]{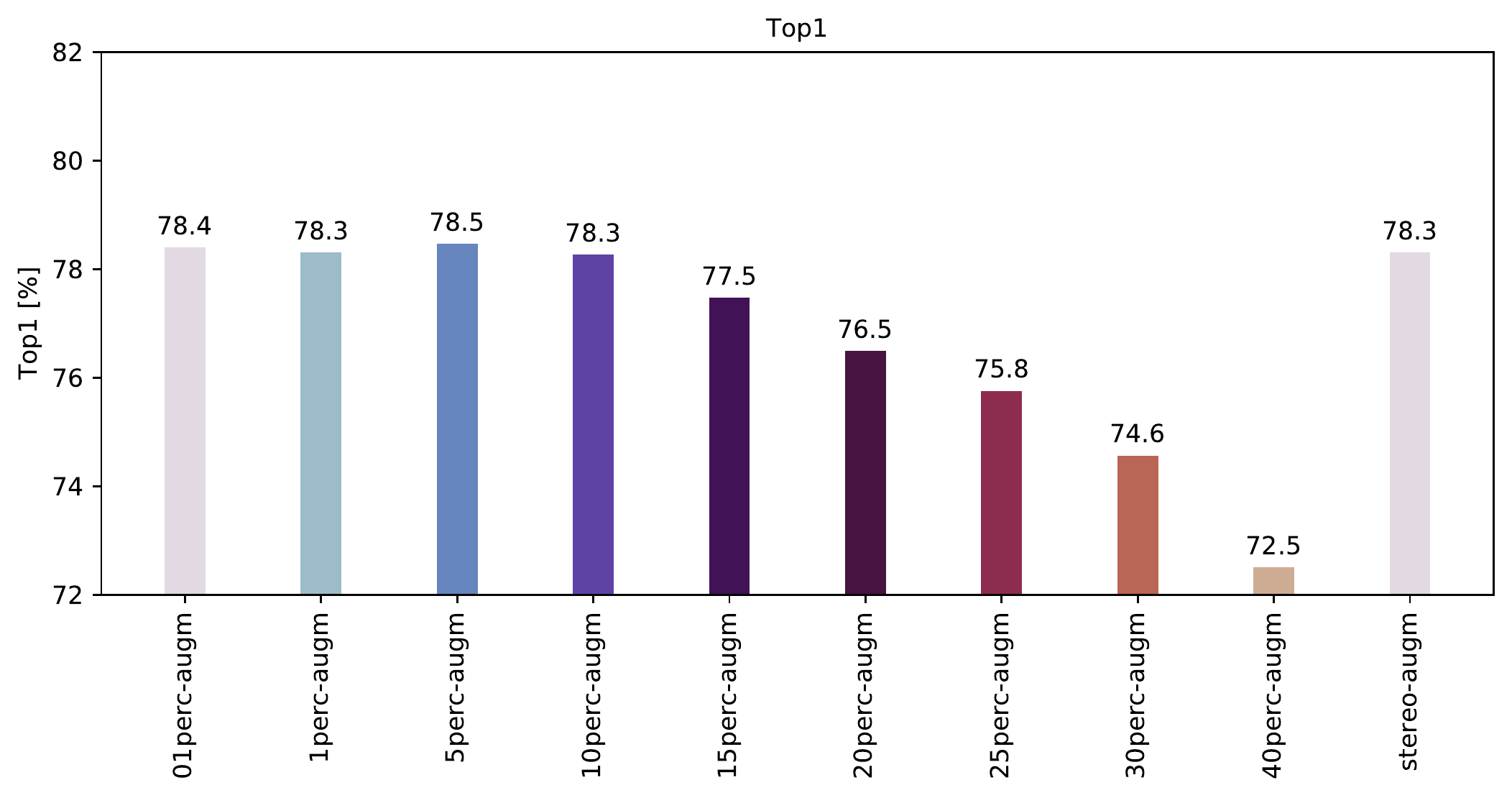}
}
\captionsetup{width=.9\linewidth,font=footnotesize,labelfont=bf}
  \caption{\textbf{Top1 accuracy for Lowe.} Top1 accuracy on a common test set for the model trained on increasing percentages of removed forgotten events. In all cases the dataset was augmented in the same manner as the experiment with Pistachio (see main manuscript). "stereo-augm" is the baseline model, where no samples were removed.}
  \label{fig:stereo-Top1-test}
\end{figure}

\begin{figure}
\centerline{
\includegraphics[width=120mm]{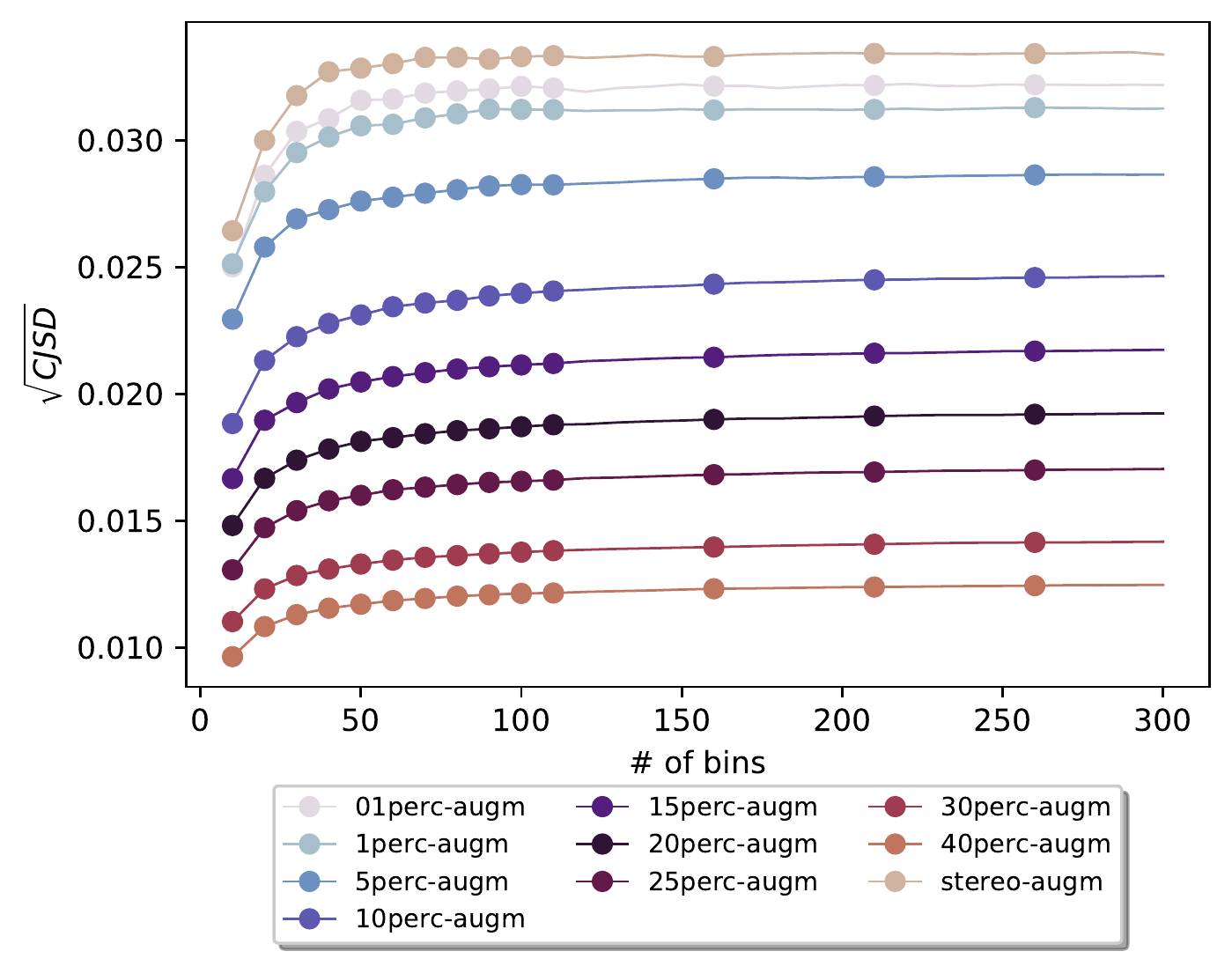}
}
\captionsetup{width=.9\linewidth,font=footnotesize,labelfont=bf}
  \caption{\textbf{CJSD for Lowe.} Cumulative Jensen Shannon divergence of the models cleaned by the forgetting strategy on the Lowe dataset \cite{lowe2012extraction}.}
  \label{fig:stereo-sqrtCJSD-11cl-test}
\end{figure}

\black{The same technique used on the Pistachio proprietary dataset \cite{Pistachio}, was applied to a open dataset: the Lowe \cite{lowe2012extraction}. Below are reported the results for the number of the forgotten events divided by macroclasses (Figure \ref{fig:stereo-forgotten-events} and Figure \ref{fig:stereo-never-learnt-events}). As for the case of Pistachio, these distributions are different within reaction classes. Again the class of Unrecognized reactions (class 0) and the one of Resolutions (class 11) are those experiencing the highest value of never-learnt events.\\
In Figure \ref{fig:stereo-Top1-test} are then reported the results for the Top1 accuracy of the trained reduced-dataset models. The trend follows the same behaviour observed in the Pistachio dataset, but the percentages of removed forgotten examples at which we have the "turning point" (important chemical examples are removed) are lower. This is expected due to the decreased size of the Lowe dataset (\~ 1 million entries). The choice of the cleaned dataset would in this case fall in between the removal of 5\% and 10\% of the forgotten examples. Finally, in Figure \ref{fig:stereo-sqrtCJSD-11cl-test} is reported also the Cumulative Jensen Shannon metric (without the Resolutions class): again the forgotten events technique allows for a better balance within different likelihood distributions.}

\clearpage
\subsection{Synthesis routes}
\begin{figure}[!h]
\centerline{
\includegraphics[width=170mm]{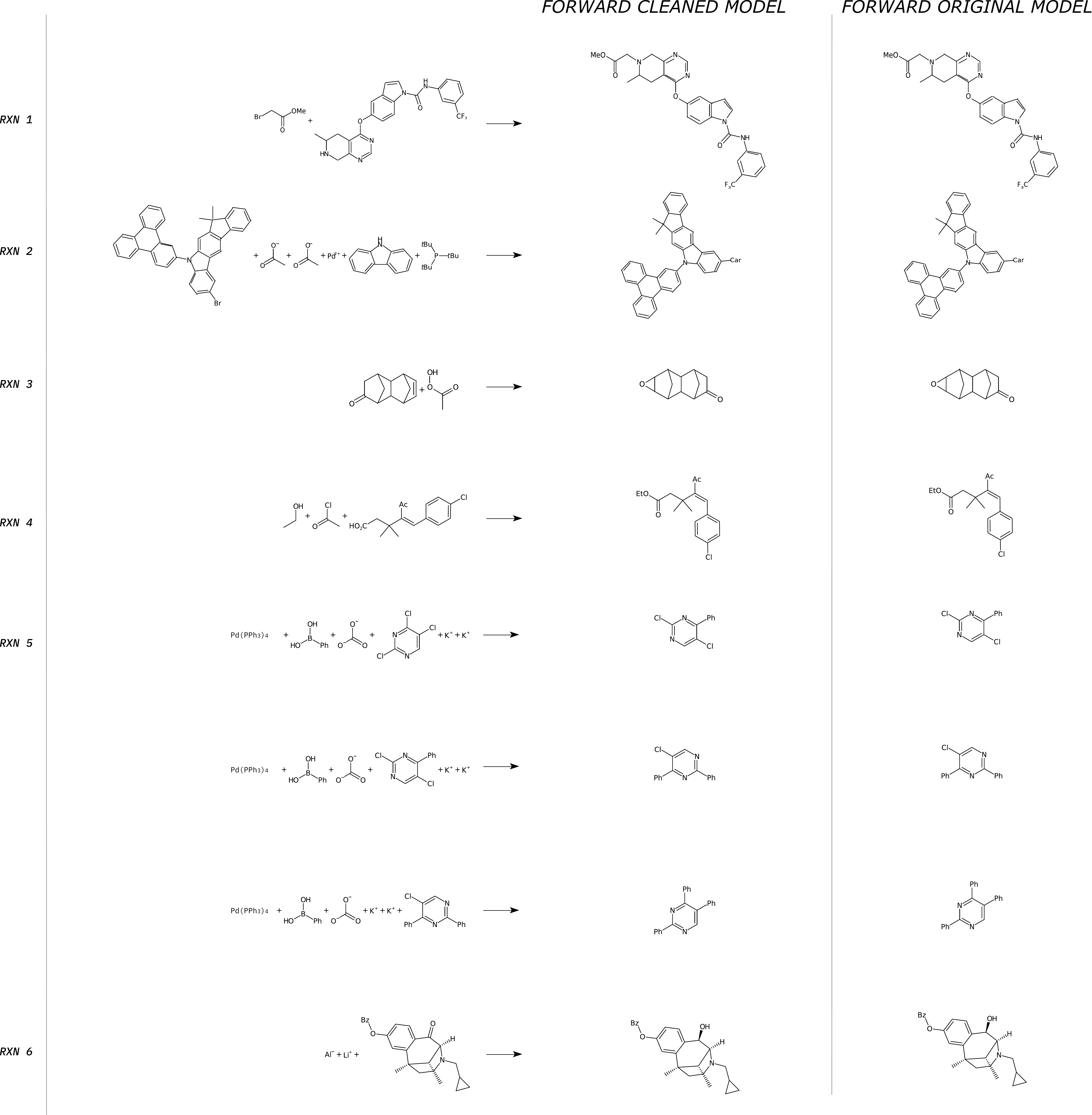}
}
\captionsetup{width=.9\linewidth,font=footnotesize,labelfont=bf}
  \caption{\textbf{Forward synthesis prediction.} Reactions tested on the new cleaned forward model. For more details we refer to Schwaller et al. \cite{SchwallerFWD}.}
  \label{fig:fwd_rxns_1}
\end{figure}

\begin{figure}[!h]
\centerline{
\includegraphics[width=170mm]{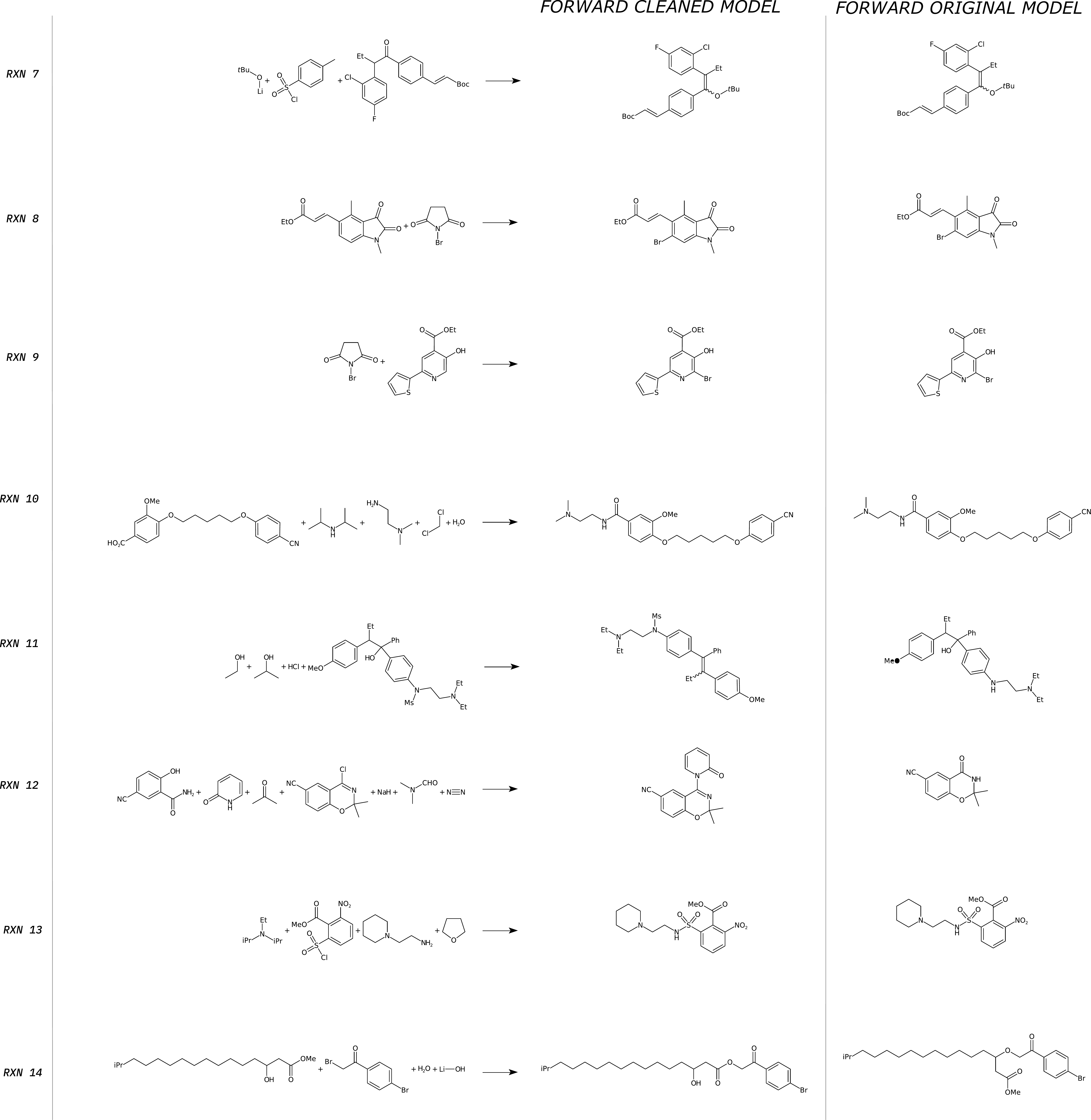}
}
\captionsetup{width=.9\linewidth,font=footnotesize,labelfont=bf}
  \caption{\textbf{Forward synthesis prediction - continuation.} Reactions tested on the new cleaned forward model, continuation. For more details we refer to Schwaller et al. \cite{SchwallerFWD}}
  \label{fig:fwd_rxns_2}
\end{figure}

In Figure \ref{fig:fwd_rxns_1} and  \ref{fig:fwd_rxns_2} are reported the compounds \cite{SchwallerFWD} tested on the original and cleaned model for what concerns the forward synthesis. The compounds for the assessment of the quality of the retrosynthesis design are reported in Table \ref{table:retro_rxns}. They are the same compounds tested in Schwaller et al. \cite{schwaller2020predicting}. Following, are the synthesis routes generated by the new model for the cited compounds.

\begin{table}[!h] 
\centering 
\begin{tabular}{c c c}      %
\hline\hline\\
\centering
\includegraphics[width=35mm]{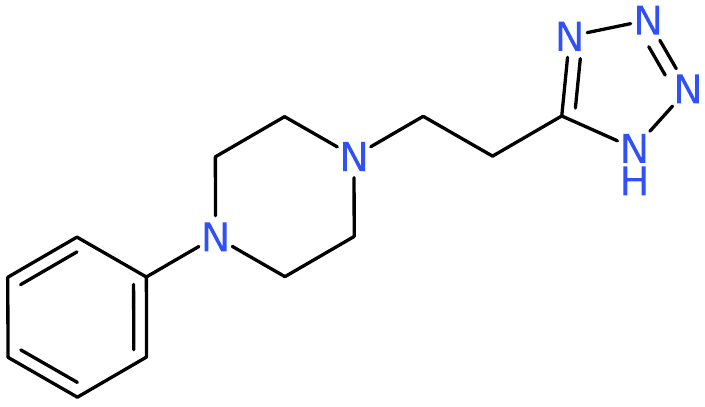} &
\includegraphics[width=35mm]{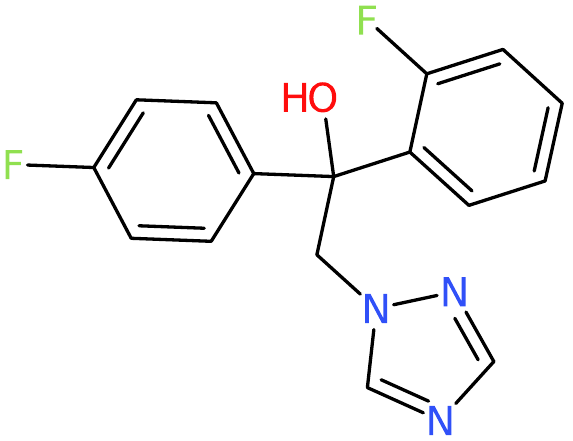} & \includegraphics[width=35mm]{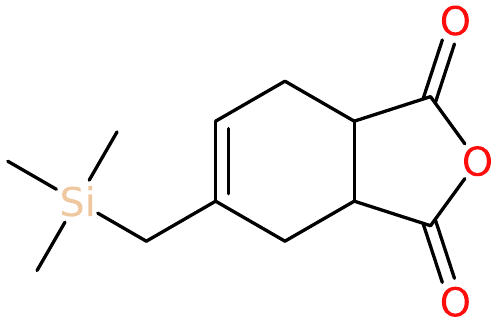} \\\\
\centering
\textbf{1} &
\textbf{2} &
\textbf{3} \\\\
\centering
\includegraphics[width=45mm]{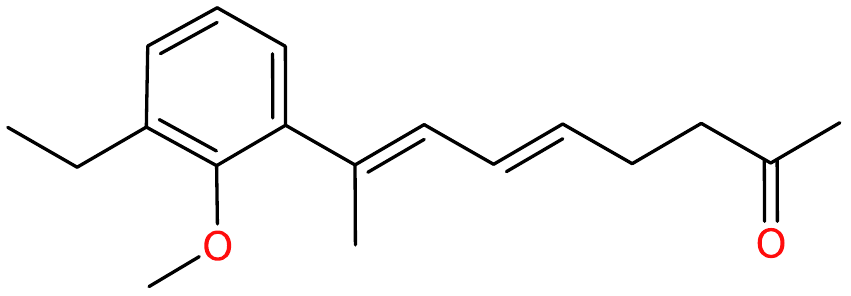} &
\includegraphics[width=45mm]{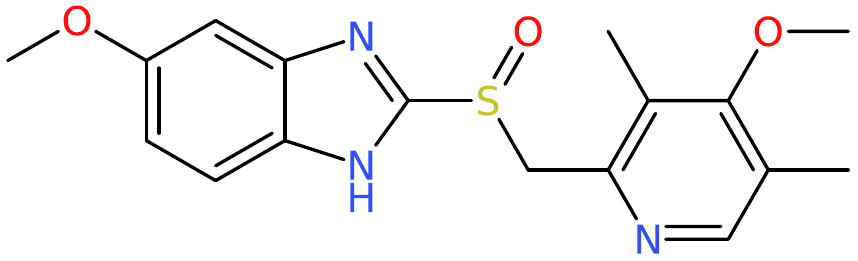} & \includegraphics[width=45mm]{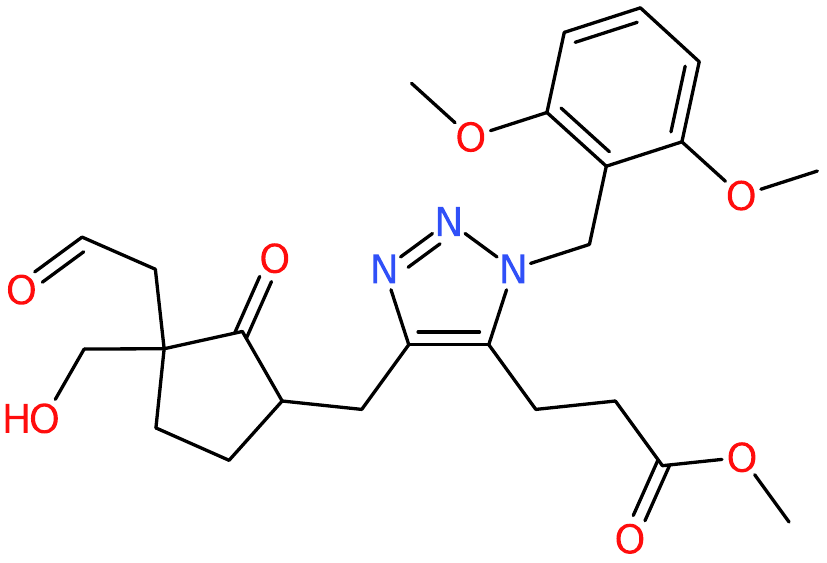} \\\\
\centering
\textbf{4} &
\textbf{5} &
\textbf{6} \\\\
\centering
\includegraphics[width=45mm]{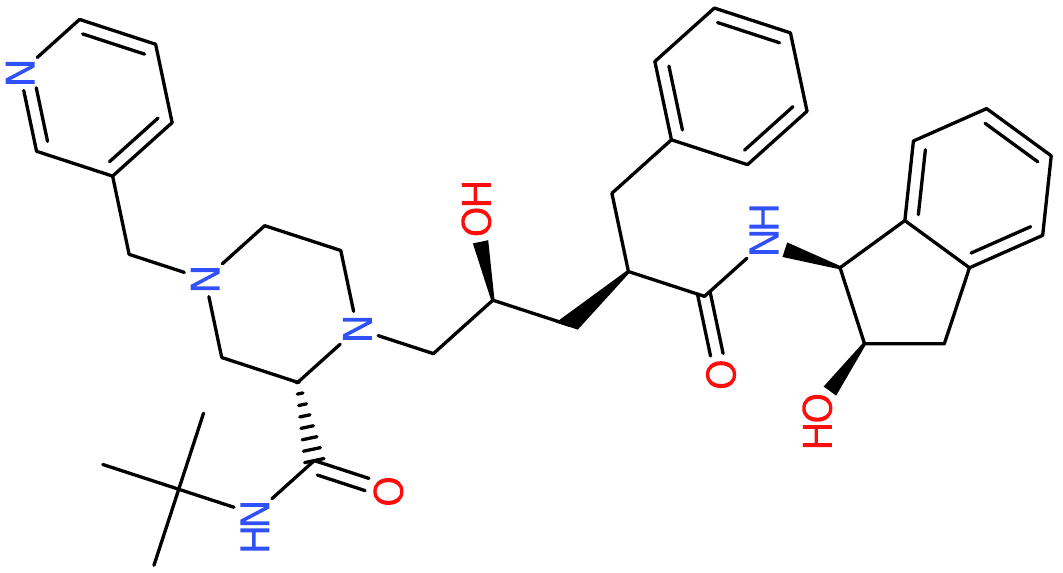} &
\includegraphics[width=25mm]{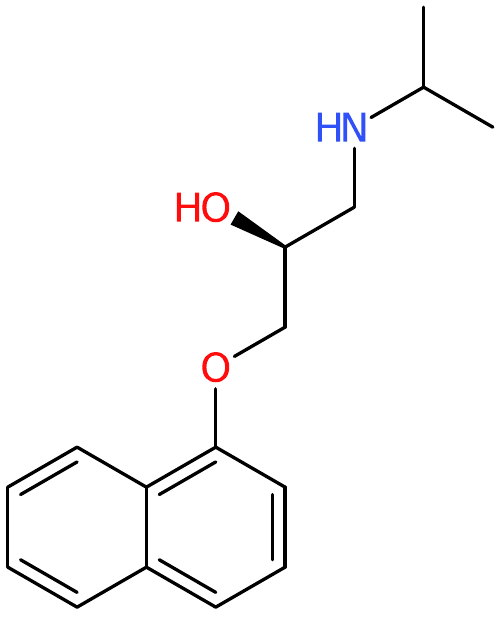} & \includegraphics[width=45mm]{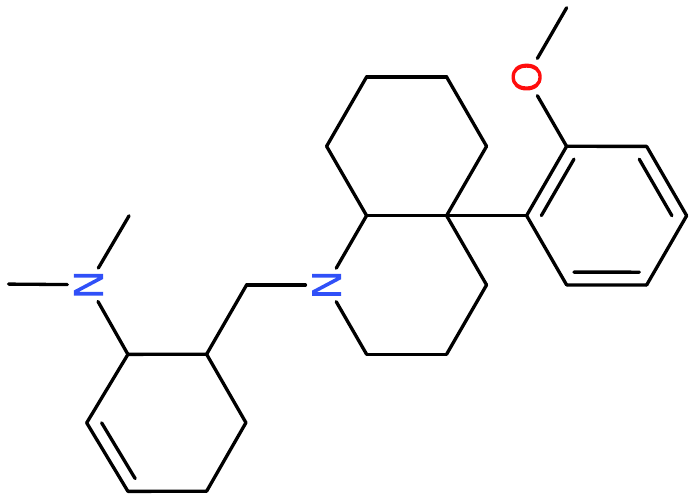} \\\\
\centering
\textbf{7} &
\textbf{8} &
\textbf{9} \\\\
\hline  
    \end{tabular}
\captionsetup{width=.9\linewidth,font=footnotesize,labelfont=bf}
    \caption{\textbf{Retrosynthesis prediction.} Nine molecules already tested for the original model \cite{schwaller2020predicting}, tested now for the cleaned retro model.}
    \label{table:retro_rxns}
\end{table}

For compound \textbf{1} the new model assigns the highest confidence to a one-step retrosynthesis, trading the formation of the tetrazole ring for a commercial precursor carrying the same substructure (Sequence 0). While this is an interesting strategy for operational reasons, it utilizes complex and more expensive starting materials. Nonetheless, the subsequent recommended path shows disconnection strategies similar to those of the literature and the original model (Sequence 3). Only the conditions for the first retrosynthesis step changes from literature: recommending trimethylsilyl azide instead of sodium azide. Both approaches are chemically valid, with the most effective being decided by costs, risks, environmental impact, yield. For compound \textbf{2}, three different synthesis are reported in literature, where the shortest one exploits the opening of the epoxide ring. The improved model recommends an alternative sequence with respect to all those known, which differs from the first one by two aspects (Sequence 0). First, two reaction steps are swapped: the attachment of the Grignard compound on the carbonyl group versus the N-alkylation. Moreover, the final ketone (alpha chloride) is not synthesized because found as commercial. Although this new route is using starting materials relatively more complex, the simplicity of the procedure may balance the increased cost of the precursors. Similar to the baseline model, among the other predicted pathways we find the optimal one reported in literature (Sequence 6). Moving to compound \textbf{3}, unlike the original model the automatic retrosynthesis using the new model did not succeed in providing any retrosynthetic path (Sequence 0). A deeper analysis shows that the first disconnection (Diels Alder cycloaddition) has a lower confidence  (0.174) compared to the original model (0.362). The removal of forgotten events may have led to the a reduction of a particular class of functional groups, which affects the assessment of the forward reaction confidence in examples containing those specific functional groups. This may be compensated by repopulating chemical reactions data with examples containing functional groups more severely affected by the noise-reduction strategy. For compound \textbf{4}, the new model suggests an alternative strategy that avoids the problem of the conjugate reduction of the aldehyde faced by the baseline model (Sequence 0). The new model improves the choice of chemoselective strategies while still showing few weakness on the stereoselectivity likely depending on the examples present in the noise-removed data sets. In fact, the choice of a less stabilized phosphonate for the Horner-Wadsworth-Emmons reaction could lead to reduced selectivity for the formation of the double bond 'E'. Similar to the radical bromination in allylic position, that may lead to selectivity issues on the primary carbon.
Compound \textbf{5} shows identical retrosynthetic pathways both with the baseline and new model (Sequence 0). Unlike for the original model, the retrosynthesis of compound \textbf{6} proposed by the new model completes with a narrower hyper- graph exploration finding a disconnection choice (alkylation of the ketone in the most substituted position) which, although not optimal in terms of regioselectivity, provides a viable path to the target. In both models, the first step is an ozonolisys, followed by the deprotection of the alcoholic group: these two steps should be inverted as the highly oxidizing conditions of ozonolysis could lead to a partial oxidation of the free hydroxyl group (Sequence 0). The retrosynthesis proposed by the new model for compound \textbf{7} is made of a single step only (Sequence 0, Confidence 0.988). If we choose to exclude the most complex molecules within the ones proposed, we can still identify a complete route (Sequence 0, Confidence 0.896). This route does not provide a central disconnection by the opening of the ossiranic ring, as reported in literature, being characterized by peripheral disconnections which lead to a sequential type of synthesis. The route is in any case valid and makes a wise use of protections and deprotections steps to handle chemoselectivity. The commercial compound found at the last step is still quite complex, but unlike the original model the handling of the many chiral centers is satisfactory. 
For compound \textbf{8} the new model proposes an additional retrosynthetic strategy, which shows the enhancement and flexibility introduced by the noise reduced data (Sequence 0). This strategy  allows for a temporary increase of the complexity of the molecule through the introduction of the carbamic ring. The retrosynthesis for compound \textbf{9} performed with the new model terminates successfully and reveals itself to be quick and simple with few obvious improvement points: at step 5 we find a saponification of an ethyl ester to acid, which in the previous steps was transformed into a methyl ester, resulting in a useless interchange of ester groups (Sequence 0).\\

\clearpage
\includepdf[pages=-]{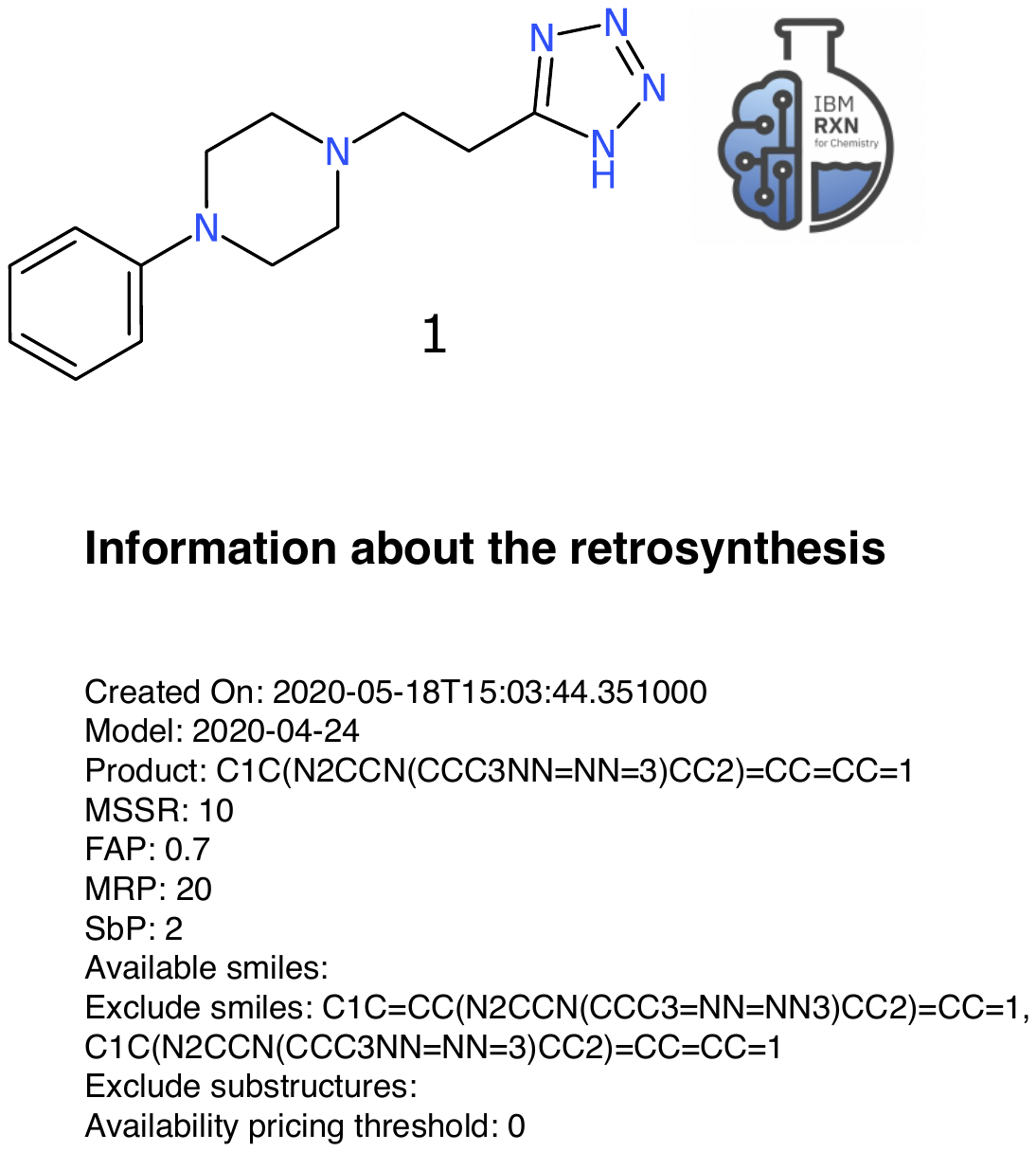}
\includepdf[pages=-]{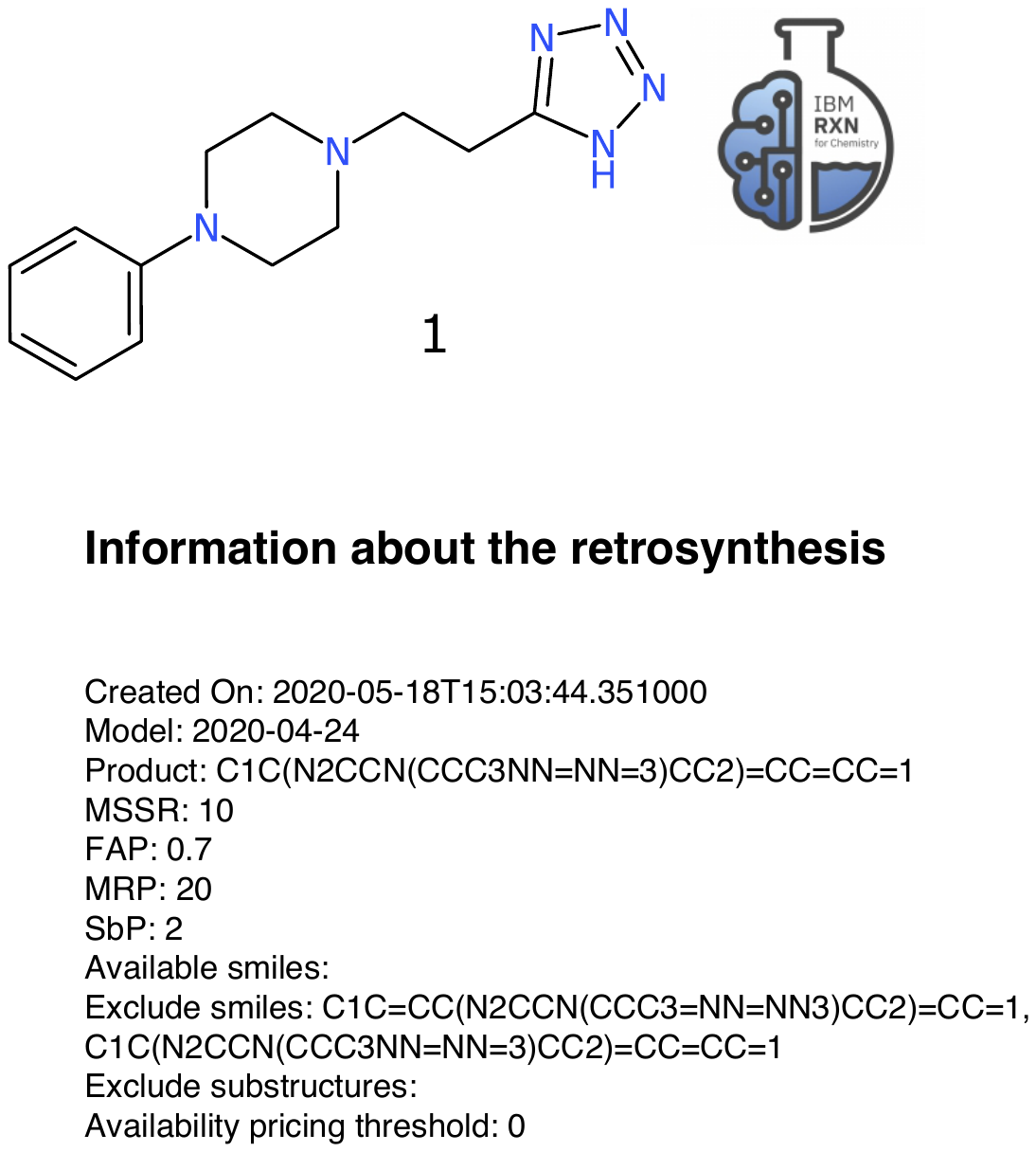}
\includepdf[pages=-]{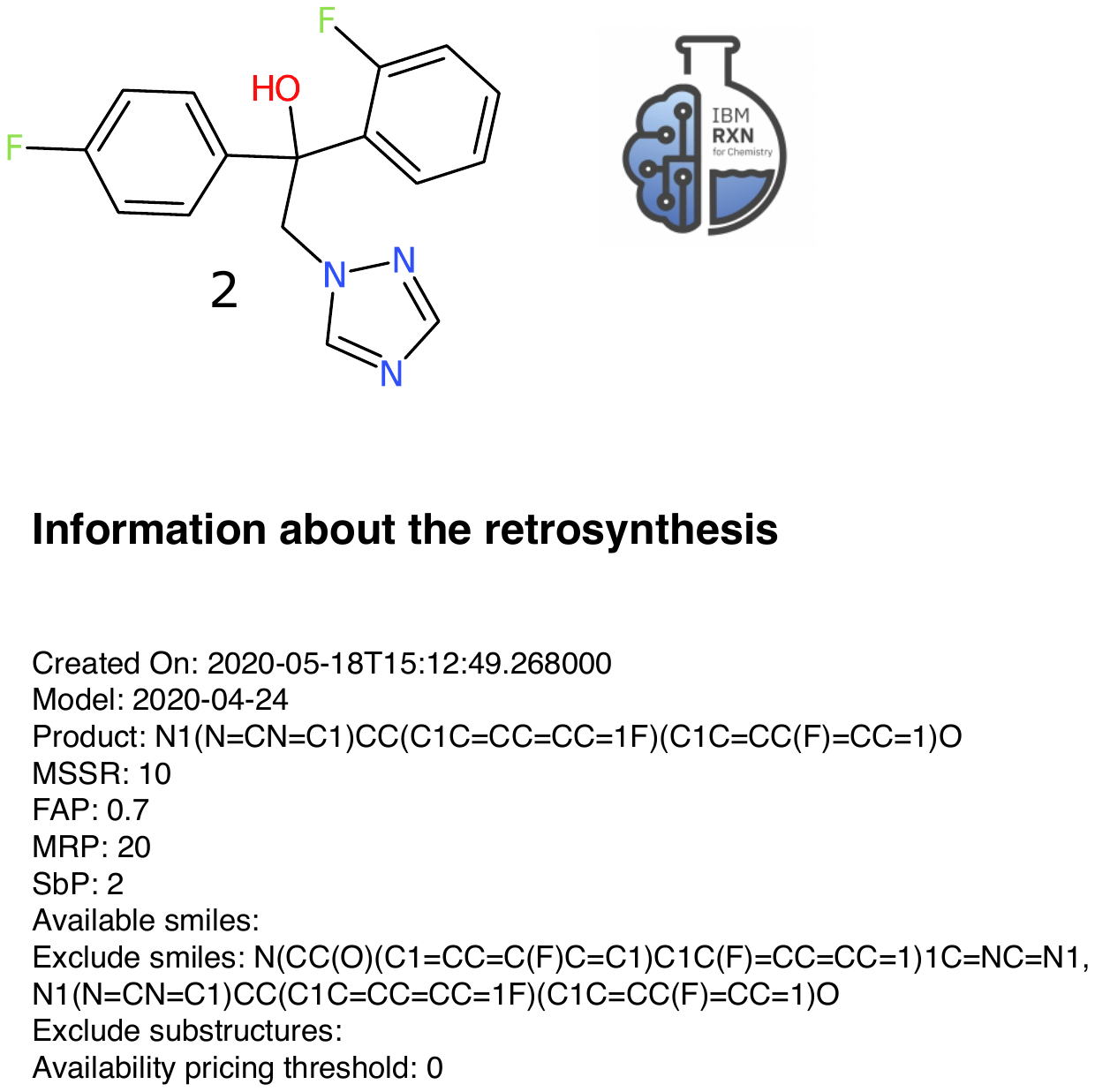}
\includepdf[pages=-]{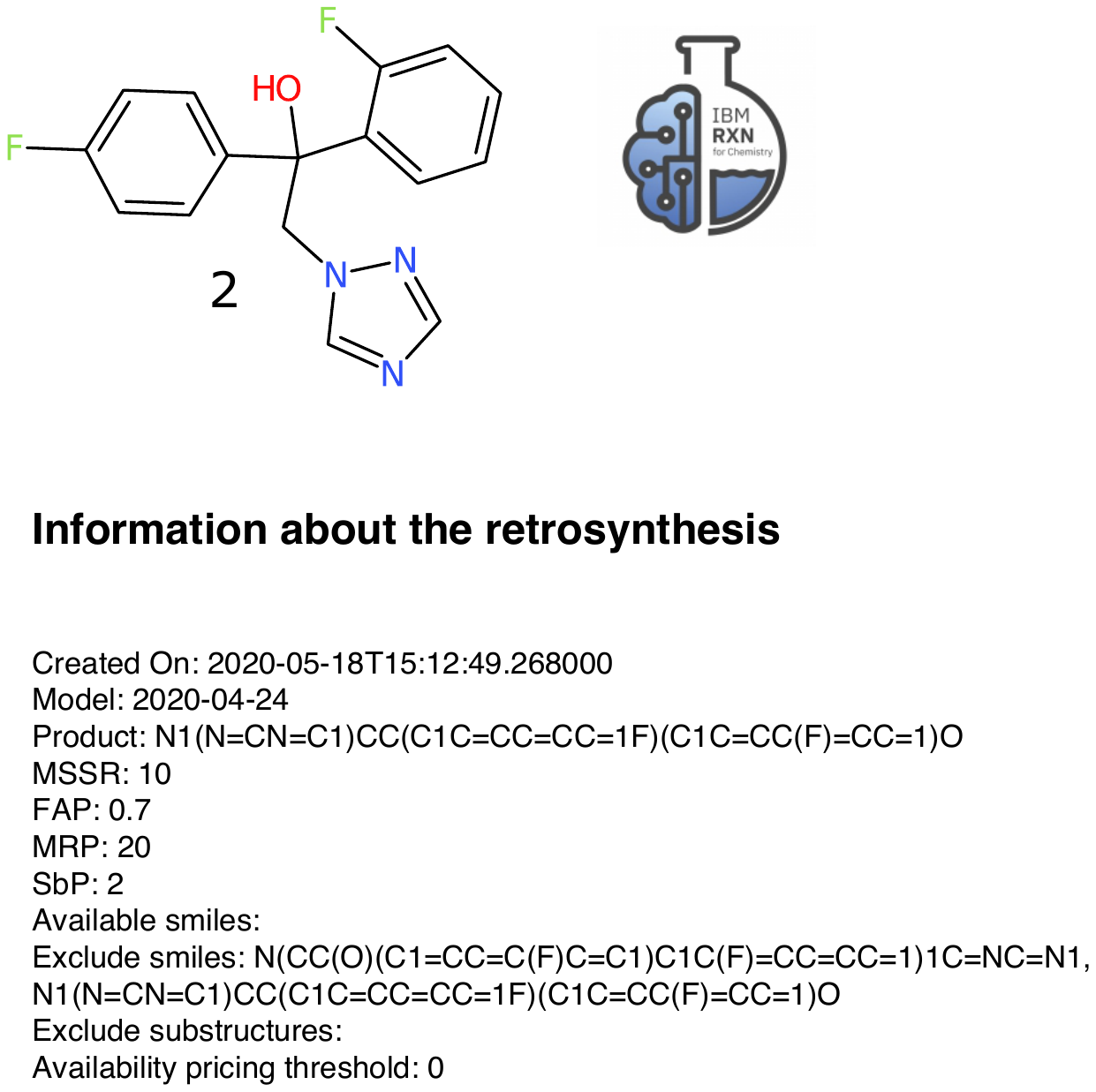}
\includepdf[pages=-]{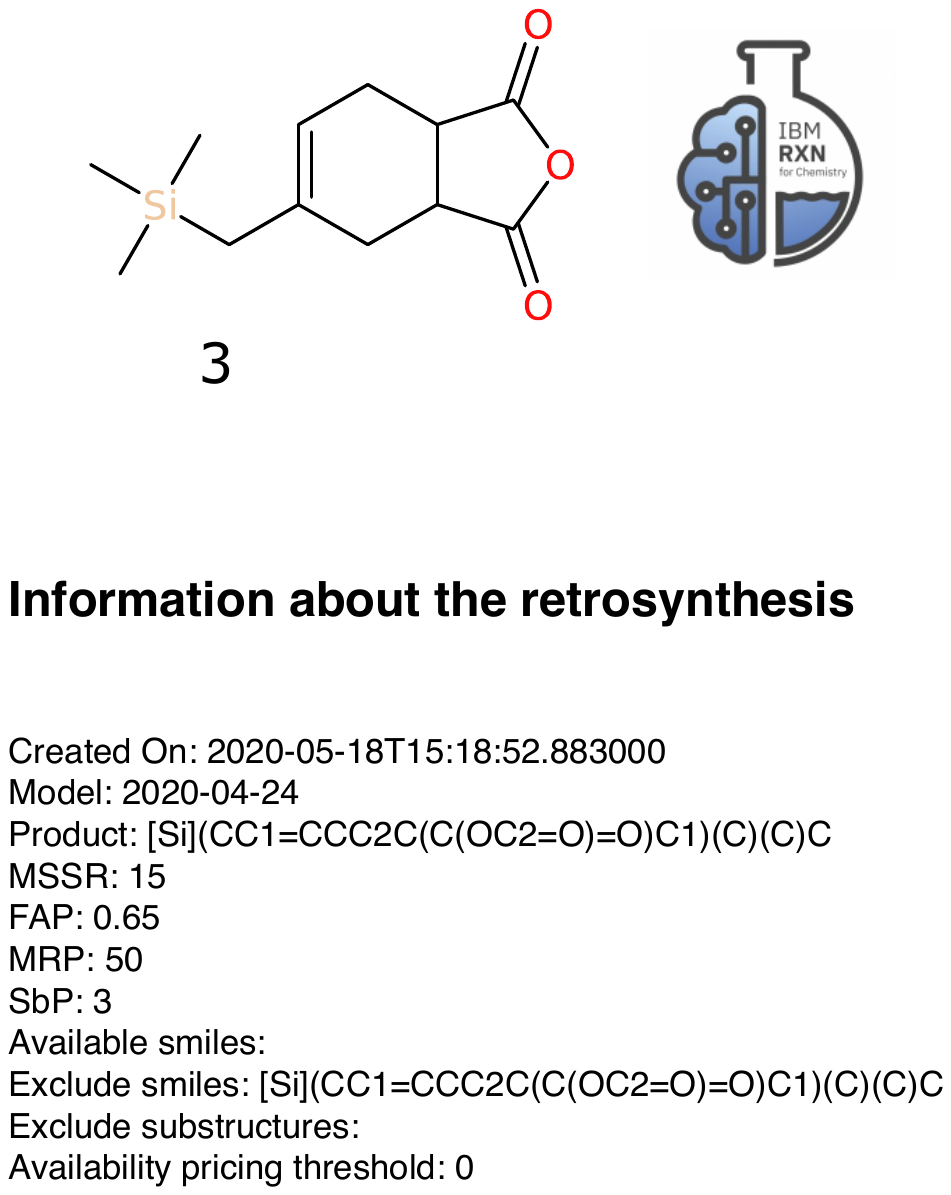}
\includepdf[pages=-]{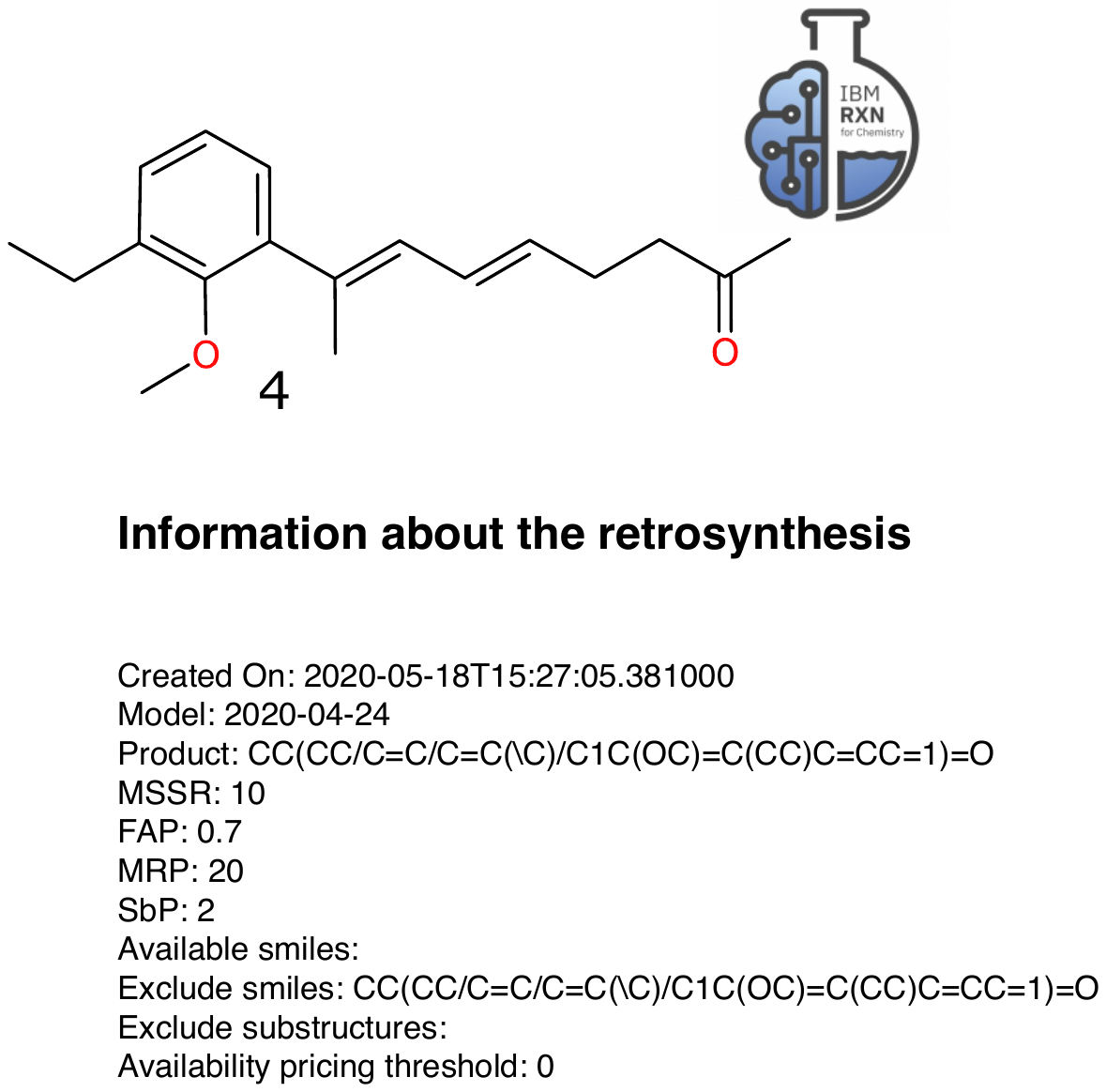}
\includepdf[pages=-]{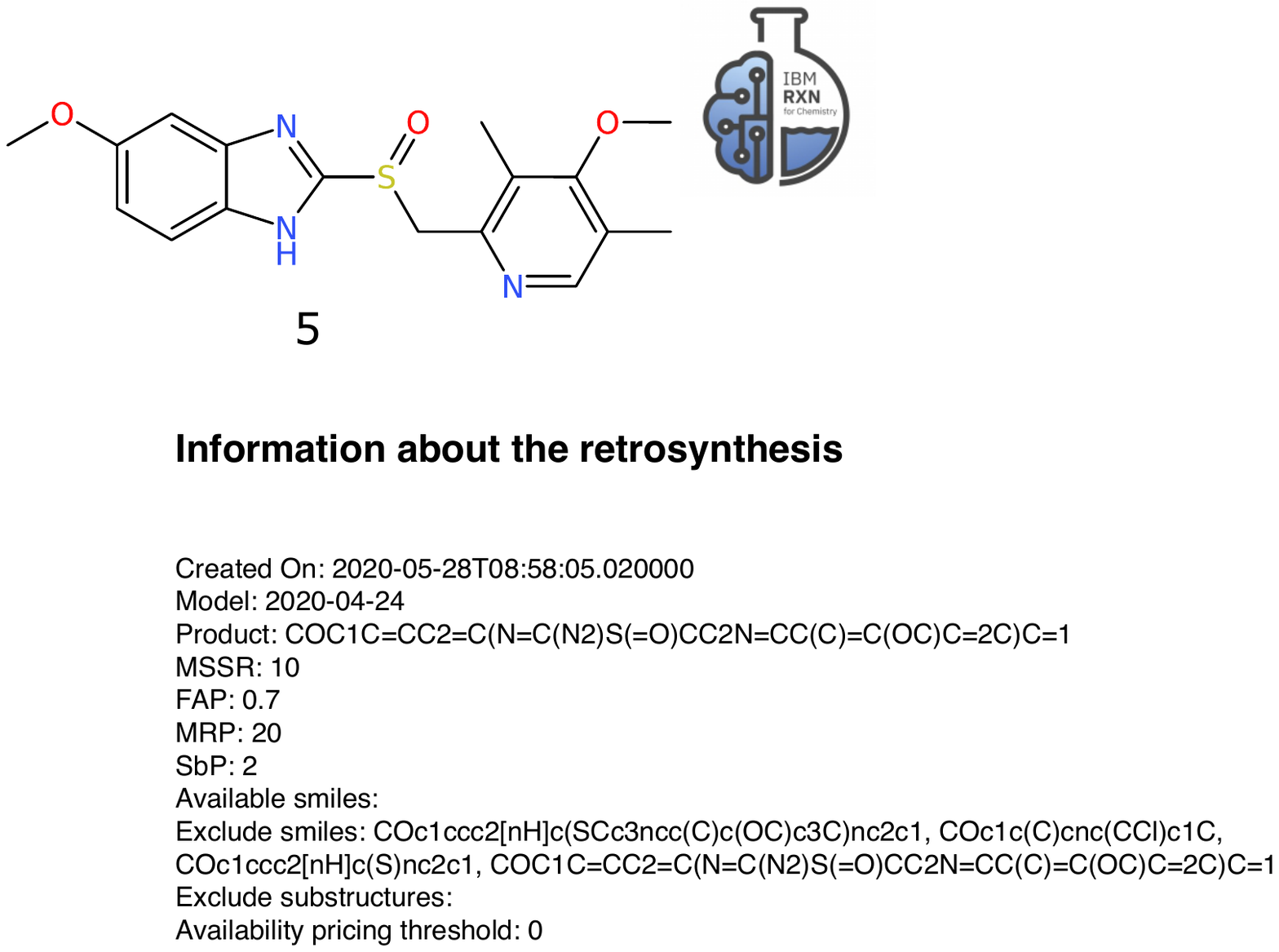}
\includepdf[pages=-]{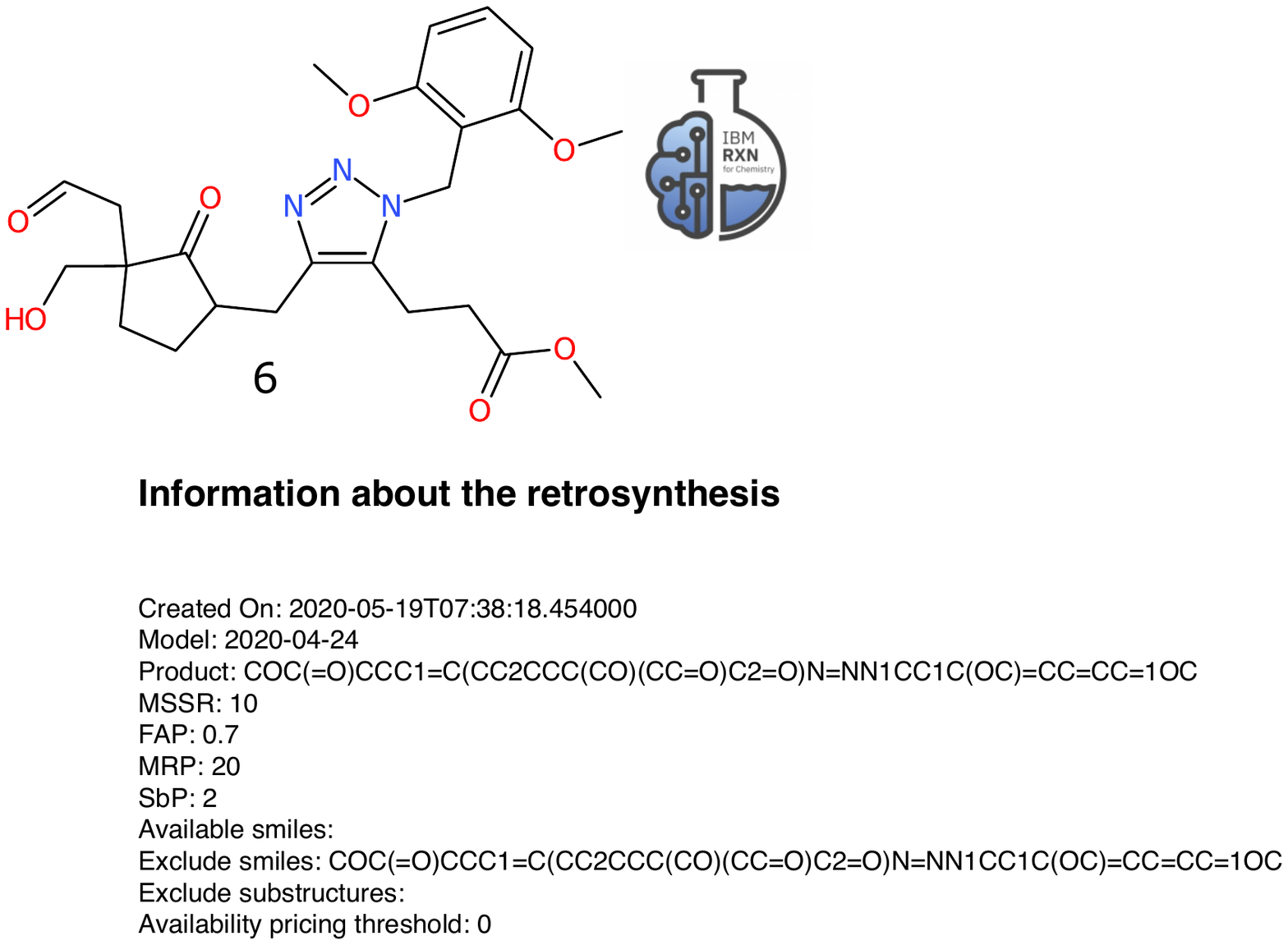}
\includepdf[pages=-]{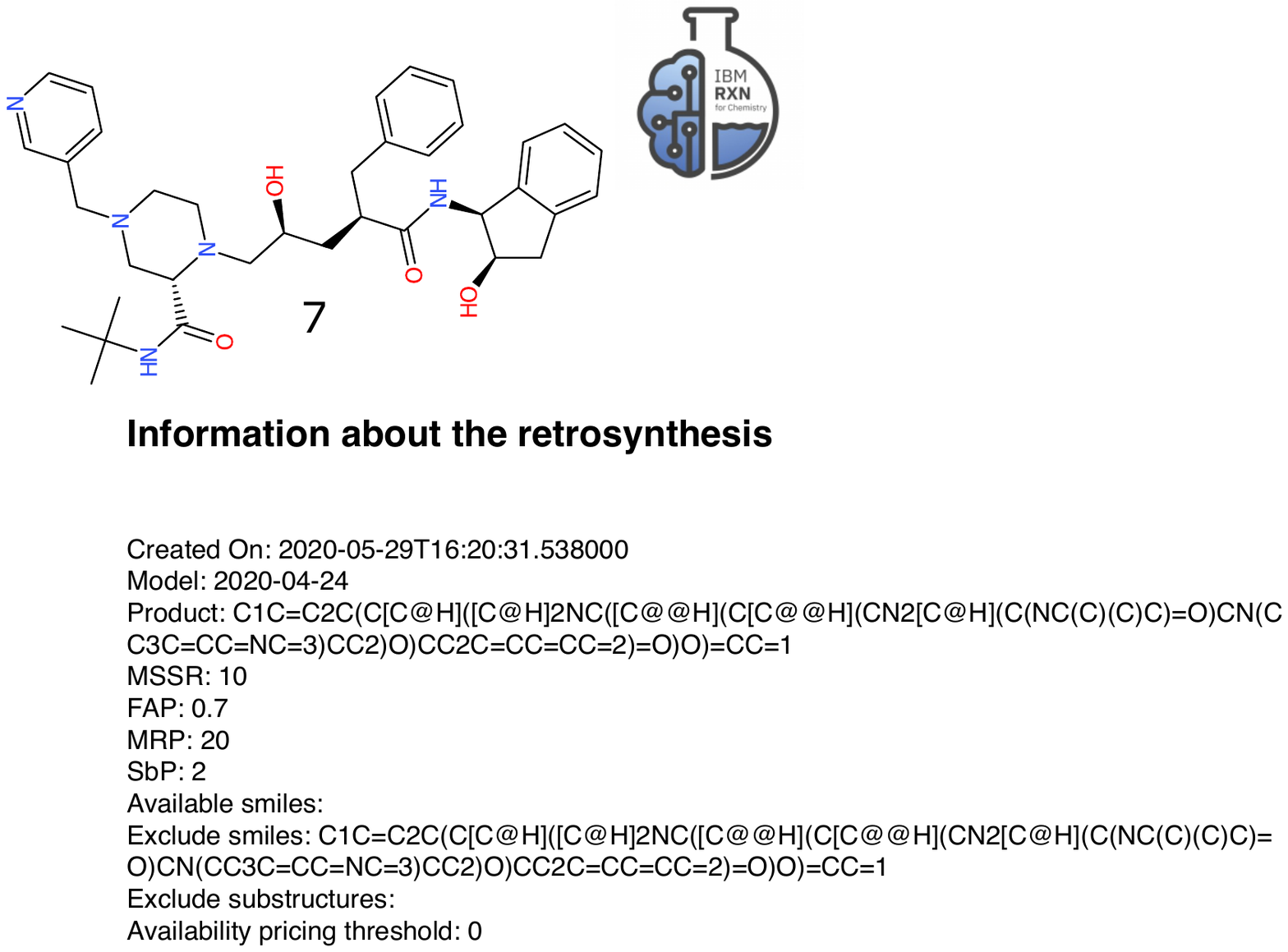}
\includepdf[pages=-]{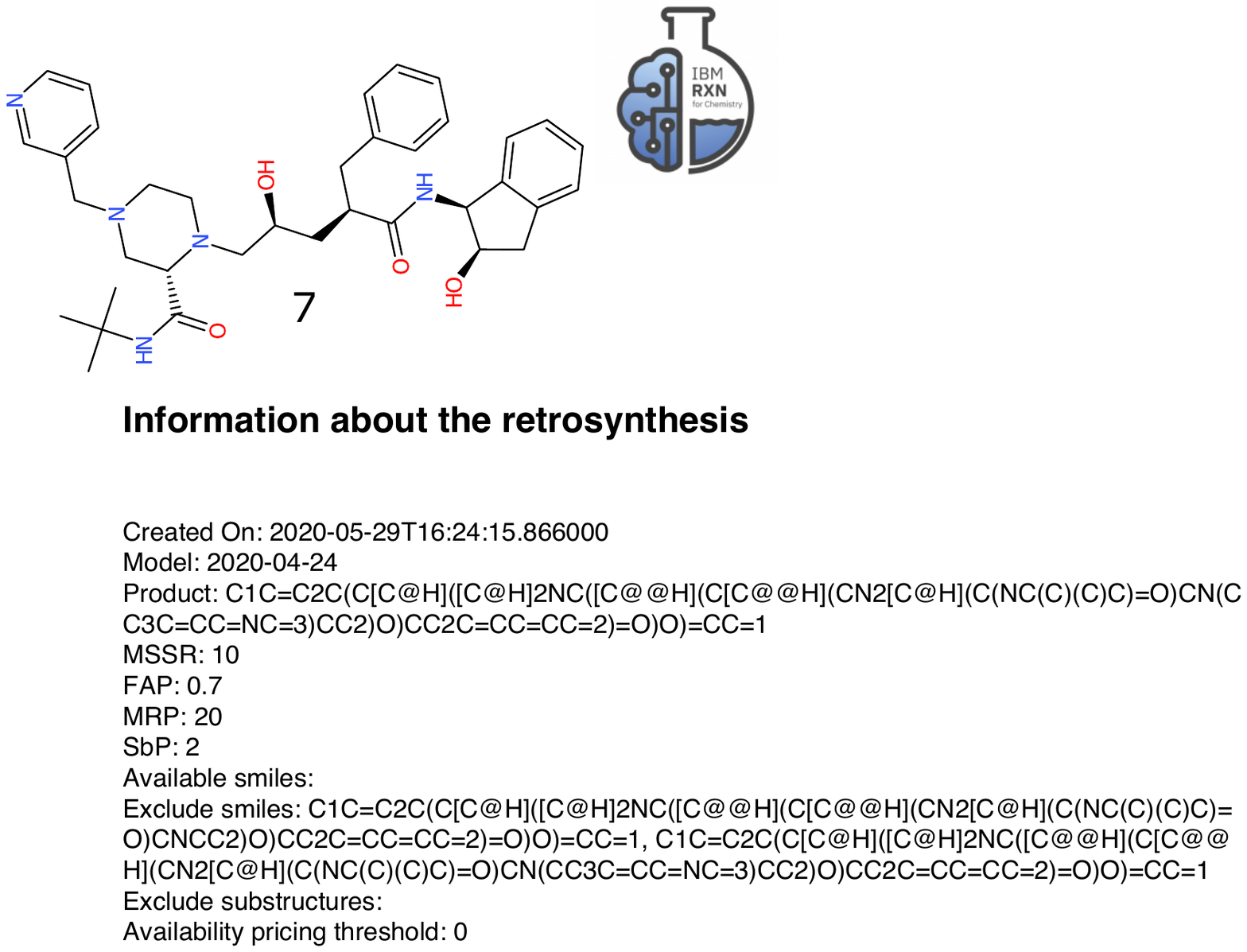}
\includepdf[pages=-]{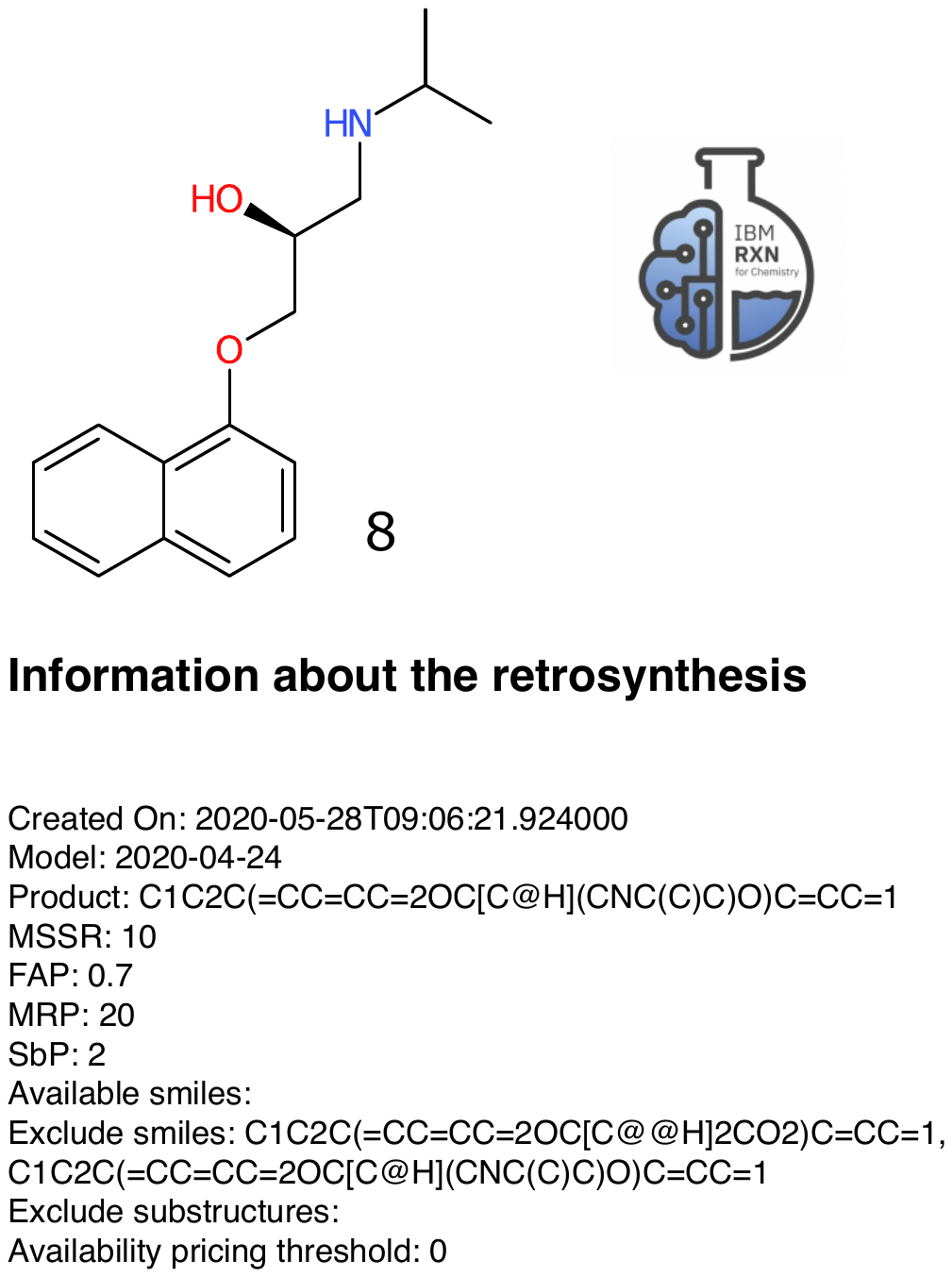}
\includepdf[pages=-]{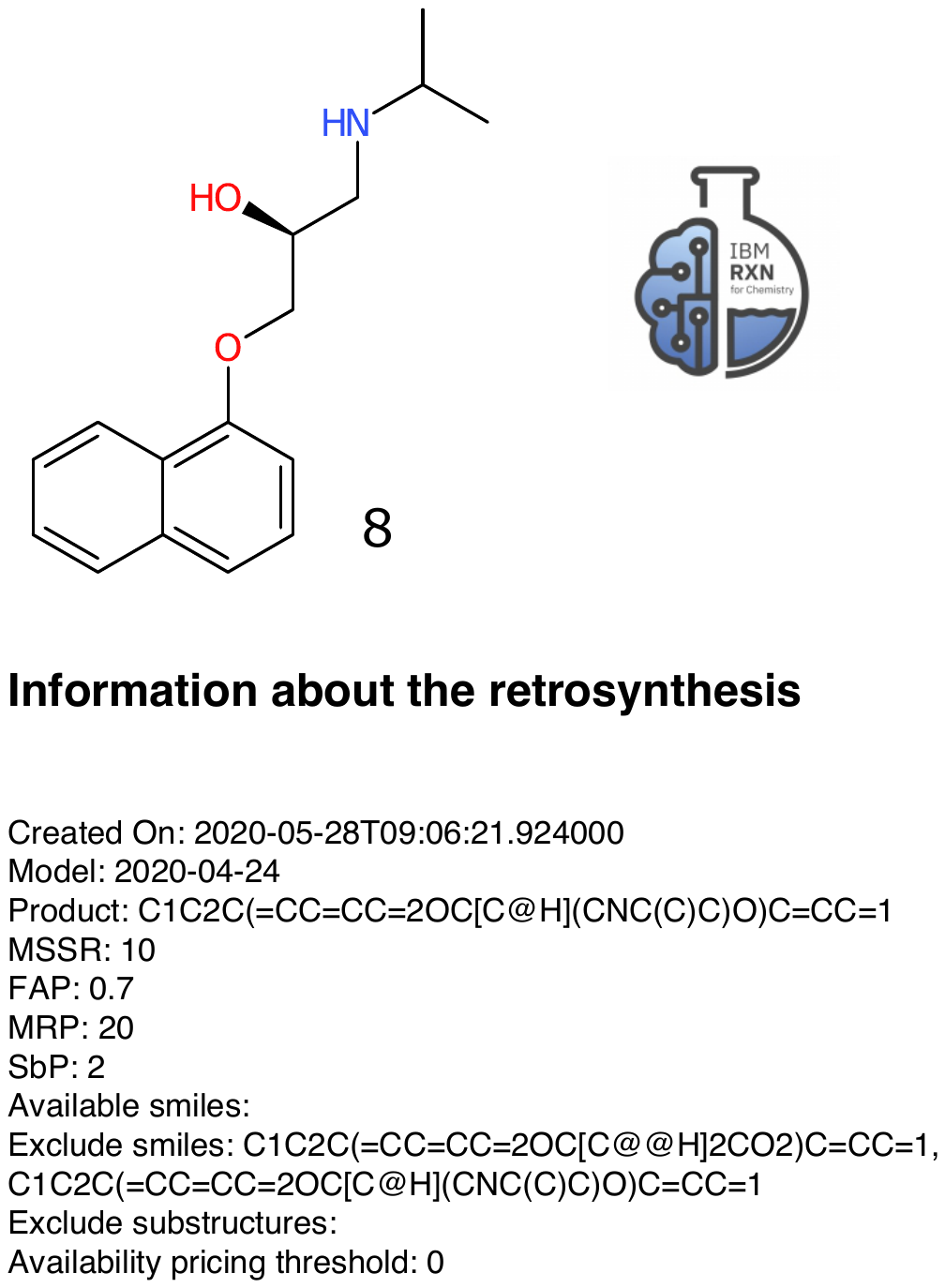}
\includepdf[pages=-]{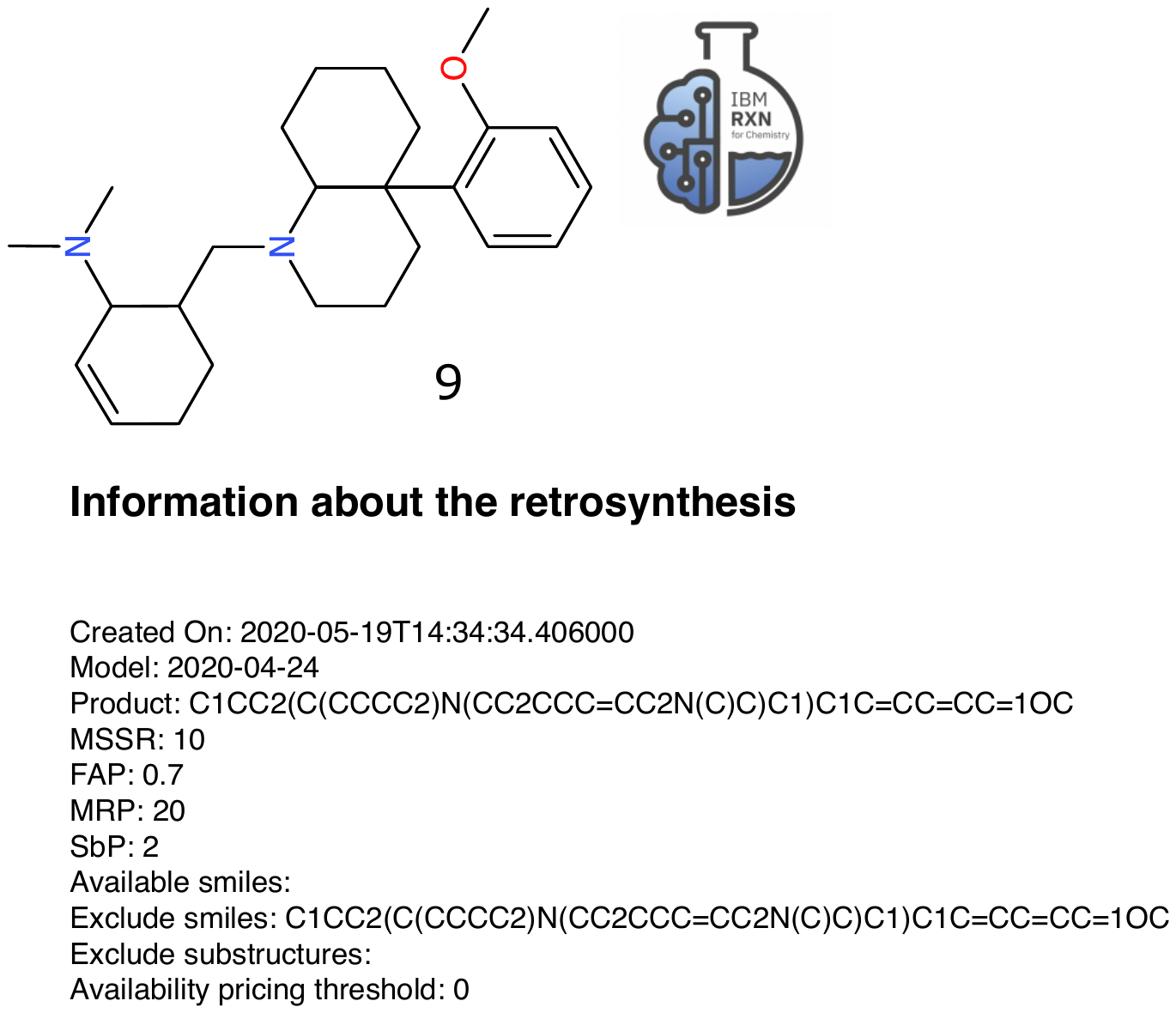}